\documentclass{article}

\usepackage{arxiv}
\renewcommand{\headeright}{}
\renewcommand{\undertitle}{}

\usepackage[utf8]{inputenc}
\usepackage[T1]{fontenc}
\usepackage{hyperref}
\usepackage{url}
\usepackage{booktabs}
\usepackage{amsmath,amssymb,amsfonts,mathtools}
\usepackage{amsthm}
\usepackage{microtype}
\usepackage{graphicx}
\usepackage{caption}
\usepackage[numbers,sort&compress]{natbib}
\usepackage{cleveref}
\usepackage{enumitem}
\usepackage{authblk}
\usepackage{multirow}
\usepackage{xcolor}
\usepackage{textcomp}
\usepackage{array}
\usepackage{longtable}
\usepackage{tabularx}
\usepackage{algorithm}
\usepackage{algorithmicx}
\usepackage{algpseudocode}
\usepackage{listings}

\graphicspath{{figures/}}

\newcommand{\sncpu}{\textsc{Symbolic Neural CPU}}
\newcommand{\dcpu}{\textsc{DCPU}}
\newcommand{\qat}{\textsc{QAT8}}
\newcommand{\ntm}{\textsc{NTM}}
\newcommand{\alu}{\textsc{ALU}}

\newcommand{\alif}{\textsc{ALIF}}
\newcommand{\gru}{\textsc{GRU}}
\newcommand{\riscv}{\textsc{RISC-V}}
\newcommand{\opset}{\ensuremath{\mathcal{O}}}
\newcommand{\regset}{\ensuremath{\mathcal{R}}}

\newcommand{\loss}{\ensuremath{\mathcal{L}}}
\newcommand{\R}{\ensuremath{\mathbb{R}}}

\newcommand{\onehot}{\ensuremath{\mathrm{onehot}}}
\newcommand{\softmax}{\ensuremath{\mathrm{softmax}}}
\newcommand{\argmax}{\ensuremath{\mathrm{arg\,max}}}
\newcommand{\argmin}{\ensuremath{\mathrm{arg\,min}}}
\newcommand{\CE}{\ensuremath{\mathrm{CE}}}
\newcommand{\MAE}{\ensuremath{\mathrm{MAE}}}
\newcommand{\ECE}{\ensuremath{\mathrm{ECE}}}
\newcommand{\Brier}{\ensuremath{\mathrm{Brier}}}
\newcommand{\clip}{\ensuremath{\mathrm{clip}}}
\newcommand{\round}{\ensuremath{\mathrm{round}}}
\newcommand{\sg}{\ensuremath{\mathrm{sg}}}

\title{A Symbolic Neural CPU for Quantization-Simulated Writeback and Interpretable Program Execution}
\date{}
\usepackage{authblk}
\usepackage{hyperref}

\author[1,2]{%
Jose Luis Lima de Jesus Silva%
\thanks{Corresponding author. Email
\href{mailto:jseluis.silva@gmail.com}
{\texttt{mailto:jseluis.silva@gmail.com}}}%
}

\affil[1]{Federal University of Bahia, Department of Geophysics, Salvador, BA 40170-115, Brazil}

\affil[2]{Grupo de Estudos e Aplica\c{c}\~ao de Intelig\^encia Artificial
em Geof\'isica (GAIA), Federal University of Bahia, Salvador,
BA 40170-115, Brazil}

\begin{document}
\maketitle

\begin{abstract} Neural networks can learn algorithmic input-output mappings, yet learned executors remain difficult to trust because their internal state transitions are usually hidden. We introduce a trace-supervised symbolic neural CPU, a factorized learned execution architecture that combines recurrent control, an explicit operation router over a fixed differentiable arithmetic-logic unit bank, destination-masked register writeback, complete trajectory supervision and matched fixed-point replay. The resulting execution process is auditable through the selected operation, source and destination registers, register trajectory, memory signals and writeback semantics. On the principal 16-wide benchmark, the non-quantized executor reproduces reference execution exactly, while the eight-bit quantization-simulated executor preserves the symbolic operation path through programs of 1000 instructions. Residual numerical drift disappears against a matched fixed-point replay, identifying a mismatch between continuous and low-precision reference semantics rather than execution failure. Architecture comparisons span recurrent, Transformer, temporal-convolution, temporal graph-inspired and state-space controllers. Ablations show that operation-gate supervision is necessary for an inspectable execution path, while hidden-opcode memory-pressure tasks expose the current limits of delayed state use and temporal binding. We further extend the interface with ValueMemory, hybrid adaptive leaky integrate-and-fire controllers, candidate-constrained symbolic control trained through behavior cloning and actor-critic reinforcement learning, and an RV32I base-integer semantic bridge. Together, these results establish a trace-verifiable framework for interpretable, low-precision and controllable neural execution. \end{abstract}

\section{Introduction}
Modern computers are not judged only by their final outputs. They are auditable transition systems where instructions select operations, operations act on addressable state, and the resulting trajectory can be inspected step by step. This separation between control, arithmetic, and memory is central to the stored-program view of computation, where variables can be manipulated by changing the addresses from which values are read and to which values are written \citep{Turing1936,vonNeumann1945,Graves2016DNC}. Modern neural networks have followed a different route. They learn continuous input-output maps from data and are usually evaluated by task-level accuracy, even when the task itself requires an algorithmic procedure. The difficulty is therefore not simply whether a neural model can approximate the input-output behavior of a program, but whether it can expose an execution process sufficiently faithful to be audited as computation.

This issue has shaped a long line of work at the boundary between learning, memory and reasoning. Recurrent neural networks (RNNs) introduced trainable hidden states for sequence processing, and long short-term memory (LSTM) networks improved the ability of recurrent models to retain information over extended time intervals \citep{Hochreiter1997LSTM,Graves2013SpeechRNN,Graves2013GeneratingSequences}. The sequence-to-sequence model, attention mechanisms, and pointer-like architectures further showed that neural networks can learn flexible forms of alignment, retrieval, and structured sequential manipulation \citep{Sutskever2014Seq2Seq,Bahdanau2015Attention,Vinyals2015PointerNetworks}. These new developments made neural systems increasingly capable of acting over temporally extended inputs, but the stored information generally remained embedded in hidden activations or learned parameters rather than represented as an explicitly inspectable machine state.

Memory-augmented neural networks made this computer analogy more direct. Neural Turing machines (NTMs) and differentiable neural computers (DNCs) coupled a neural controller to an external memory matrix that can be read from and written to through differentiable addressing \citep{Graves2014NTM,Graves2016DNC}. The DNC added content lookup, temporal linkage, and dynamic allocation, enabling learned systems to answer synthetic reasoning questions, traverse graphs, infer shortest paths, and carry out planning-like block manipulation \citep{Graves2016DNC}. Related memory-network and end-to-end memory-network architectures also demonstrated the value of separating a learned controller from an addressable store for reasoning over facts and sequences \citep{Weston2014MemoryNetworks,Sukhbaatar2015EndToEndMemory}. More recently, Transformer attention has been interpreted as a restricted, stateless form of DNC-like memory access, in which keys and values form a write-once memory and multi-head attention implements parallel content-based reads \citep{Vaswani2017Attention,Tang2026StatelessDNC}. These results establish external or attention-mediated memory as a central mechanism in neural computation. They do not, however, provide a register-level audit of which symbolic operation was selected, which state variable was updated, and whether the intermediate execution path matches a reference machine.

Neural algorithmic reasoning addresses a complementary question: whether neural networks can imitate the execution of classical algorithms. The CLRS Algorithmic Reasoning Benchmark, a suite of algorithmic tasks inspired by the textbook tradition of Cormen, Leiserson, Rivest, and Stein, evaluates models on sorting, searching, graph, string, and dynamic-programming problems, often using supervision over intermediate algorithm states rather than final answers alone \citep{Cormen2009CLRS,Velickovic2022CLRS,Li2025OpenBookNAR}. Open-book neural algorithmic reasoning extends this setting by allowing a model to consult training instances during inference, turning examples into an auxiliary memory for reasoning \citep{Li2025OpenBookNAR}. Recent work on Transformer-NAR hybrids, tropical attention, primal-dual neural reasoners, looped Transformer simulations, and knapsack-style intermediate supervision further shows that architectural bias and intermediate-state supervision can improve neural reasoning on graph, combinatorial, and dynamic-programming tasks \citep{Zhang2024TransNAR,TropicalAttention2025,PrimalDualNAR2025,HypergraphLoopedTransformers2025,KNARsack2025}. This literature is similar to trace-supervised execution because it treats intermediate computation as an object of learning. Yet its intermediate states are usually graph- or task-specific algorithm states. They are not machine states composed of registers, destination masks, operation gates, and low-precision writebacks.

A second related thread studies neural programs and neurosymbolic execution. In neural-program learning, a neural model is composed of a symbolic program, logic program, or black-box application programming interface (API), and learning requires propagating supervision through, around, or across that symbolic component \citep{SolkoBreslin2024NeuralPrograms}. Such methods are powerful because the program can enforce exact structure while the neural component handles perception, uncertainty, or incomplete inputs. The setting studied here is different. Rather than attaching a neural model to an external executor, the aim is to learn the executor itself while keeping it auditable. In this setting, the operation distribution, hard operation choice, register trajectory, and low-precision arithmetic are not hidden implementation details because they are the objects being evaluated. Related work on code simulation and differentiable program synthesis similarly treats stepwise execution as a demanding reasoning problem, but typically evaluates code-like behavior, latent program search, or learned symbolic primitives rather than register-level operation paths with matched fixed-point replay \citep{CodeSimulation2025,NeurallyInterpretedLanguages2026}.

The same demand for state-level evaluation appears at larger scales in memory-augmented agents based on large language models (LLMs). Persistent LLM agents increasingly rely on external memory banks, but recent work shows that memory is not a passive store. Therefore, the construction, retrieval, and utilization form a coupled cycle, and downstream errors should feed back into memory repair \citep{Lin2026MemMA}. The continual learning studies of memory-augmented agents similarly show that external memory does not remove interference. It actually shifts the bottleneck from parameter overwriting to memory representation, retrieval, and context competition \citep{Hu2026ContinualMemory}. These observations support a general principle that memory-augmented computation should not be evaluated solely by the correctness of the final response. This means we should also consider whether the stored, retrieved, and updated state supports the intended computation. We study this principle at the level of controlled register-machine traces, where the relevant state is a register file and optional external memory rather than a conversational memory bank.

The device-facing execution imposes an additional constraint. Quantization-aware training (QAT), mixed-precision search, and hardware-aware compression show that neural models often need to be evaluated under explicit budgets for memory, bit-width, latency, and computational cost before deployment on constrained hardware \citep{Jacob2018Quantization,Krishnamoorthi2018Quantizing,VanBaelen2025CGMQ}. In addition, energy-aware inference studies similarly emphasized that accuracy alone is insufficient when computation is constrained by power and latency \citep{Isenkul2025EnergyAwareVideo}. The analog in-memory training goes further by investigating which learning algorithms remain viable when computation is mapped onto non-ideal memory devices \citep{Rasch2024AnalogInMemory}. Therefore, these hardware-oriented works motivate low-precision and cost-aware evaluation, but they usually optimize neural inference workloads. By contrast, they do not test whether a learned symbolic executor preserves a stepwise machine trace when arithmetic is projected onto a fixed-point grid with quantization-simulated writeback, rather than with full integer-only inference. Recent work on full-integer deployment and rescaling-aware training makes this distinction especially important, because low-precision arithmetic can introduce deployment costs and behaviors that are not visible in floating-point evaluation alone \citep{RescalingAwareTraining2025}.

The hardware context is also changing, as evidenced by neuromorphic many-core processors such as Loihi, which demonstrated programmable, event-driven substrates with on-chip learning, while deterministic neuromorphic systems such as NeuroScale use local, event-driven synchronization to preserve reproducible spike timing without global barriers \citep{Davies2021Loihi,Li2025NeuroScale}. At a more conceptual level, recent work on neural computers frames learned runtimes as an emerging machine form in which computation, memory, and I/O are unified in model state, while emphasizing that stable execution and symbolic control remain open problems \citep{Zhuge2026NeuralComputers}. At the processor interface, \riscv{}-based SNN accelerators such as FeNN show that programmable instruction-set-compatible platforms are being adapted for spiking neural workloads \citep{Aizaz2025FeNN}. Together with analog in-memory training and full-integer deployment work, these systems point toward real neural-computing modules in which low-precision arithmetic, memory state and programmable control will need to be verified at the level of state transitions rather than only benchmark accuracy \citep{Rasch2024AnalogInMemory,RescalingAwareTraining2025}. However, they do not supply an audit trail for learned symbolic execution.

This gap motivates the central direction of the present work. If learned executors are to become controllers, co-processors or device-facing runtimes, their operation choices, register writes, memory effects and low-precision replay semantics must be visible before hardware deployment. We therefore develop an inspectable execution interface, where a simulation-stage symbolic neural executor whose operation gate, destination mask, register trajectory and matched fixed-point replay can be compared with a reference machine. The goal is to make future systems inspectable at the transition level. Therefore, the current model is not presented as a physical processor
implementation. The primary experiments investigated symbolic execution and matched fixed-point replay under controlled conditions. As a standardized-ISA bridge, we further tested whether the same design principles can be attached to an RV32I base-integer semantic substrate. This RV32I experiment was a semantic bridge and learned-control diagnostic, not official \riscv{} compliance, privileged execution, ABI/runtime support or hardware execution.

What remains missing is an audit layer between neural algorithmic behaviour and deployable execution, where a controlled setting in which a learned system must expose the selected operation, the updated register, the complete intermediate
trajectory and the effect of low-precision arithmetic. We address this gap by developing an inspectable execution interface. Therefore, we instantiated the interface with a trace-supervised symbolic neural executor. The model receives register-machine instructions and learns to update a normalized register state through a bank of named continuous arithmetic and logic operations. Its transition is trained as a differentiable approximation to a symbolic state machine,
\begin{equation}
    \mathcal{T}: \mathcal{S}\times\mathcal{I}\rightarrow\mathcal{S},
    \label{eq:symbolic-transition}
\end{equation}
but it is evaluated at the level at which a computer would be audited, considering a final state, full register trace, operation-gate agreement, deterministic hard execution, matched fixed-point replay behaviour, and dimensionless proxy energy and delay.

We have constructed this setting by generating controlled instruction sequences over continuous register vectors in $[0,1]$ (profile widths $8$, $16$ or $32$) with exact reference traces, including the initial register file, the per-step register trajectory, the final state, operation labels, destination masks, and proxy energy and delay traces. A neural executor was then trained to imitate the reference transition using a factorized architecture. Therefore, we have an architecture with an instruction encoder, a recurrent controller, an optional differentiable memory read/write pathway, a named operation router, a continuous arithmetic-logic unit (ALU) bank, and a masked destination-register writeback. In our case, the ALU bank denotes the set of differentiable arithmetic and logical operators that the model selects at each instruction step. During training, the application of stochastic operation selection through the Gumbel-Softmax relaxation provides a differentiable path, while evaluation uses deterministic hard operation choices to test whether the learned executor follows an inspectable symbolic path rather than only matching the endpoint \citep{Jang2017GumbelSoftmax,Maddison2017Concrete}.

This controlled register-machine setting exposes the stepwise machine state that is usually hidden in neural sequence prediction. Each sample contains the instruction stream, initial register file, operation labels, destination masks, complete register trajectory, and final state. The executor is trained and evaluated against these objects rather than against a scalar endpoint alone. A single experiment can therefore investigate separate questions of whether the model chooses the correct named operation, preserves non-destination registers, follows the reference trajectory, remains stable under deterministic hard execution, and agrees with a matched low-precision replay.

In this study, we have imposed three constraints on the learned executor: (i) supervision that includes the complete reference trajectory, therefore the success requires trace faithfulness rather than endpoint imitation alone, (ii) operation selection which is exposed through named gates and evaluated in both soft Gumbel-Softmax training and hard deterministic execution, so the learned path can be inspected rather than hidden inside a dense recurrent state, and (iii) quantization-simulated writeback which is scored against both a continuous reference and a matched fixed-point replay, separating low-precision reference drift from execution error under the model's own writeback semantics. On the principal 16-wide register-machine profile, these constraints enabled the evaluation of matched replay, length extrapolation, simulated input perturbations, calibration, equivalent lower-cost programs, architecture ablations, hidden-opcode memory pressure, spiking-controller variants, candidate-constrained closed-loop control, and an RV32I base-integer semantic bridge. The result is a controlled register-machine substrate, and not a general differentiable neural computer, neural algorithmic reasoner, or program synthesizer, for testing whether learned execution preserves the operation path, intermediate machine trajectory, and low-precision reference semantics behind a final answer. The scope limitations are consolidated in the Discussion.

\section{Methods}
\label{sec:methods}

\subsection{Execution interface}
\label{sec:methods_interface}

All experiments use the same controlled register-machine interface, shown
schematically in \Cref{fig:architecture}. This interface follows the
register-machine and small-step operational view of computation, in which
execution is represented as a sequence of state transitions over an explicit
machine state \cite{ShepherdsonSturgis1963,Plotkin2004SOS}. The figure gives
the architectural view of the executor, and shows the state,
instruction, operation, writeback and replay objects used throughout the
experiments.

A program has length $T$ and is executed over discrete steps
$t\in\{1,\ldots,T\}$. The machine state at step $t$ is a register file
\begin{equation}
    r_t \in [0,1]^{R\times W},
    \label{eq:interface-state}
\end{equation}
where $R$ is the number of registers and $W$ is the number of scalar lanes in
each register. The vector $r_{t,j}\in[0,1]^W$ denotes register $j$ at step $t$.
The width $W$ is a continuous register-vector width. It is not an integer bit
width and should not be read as a 16-bit or 32-bit instruction-set architecture.

Each instruction is a tuple
\begin{equation}
    \xi_t = (o_t,i_t,j_t,d_t),
    \label{eq:interface-instruction}
\end{equation}
where $o_t\in\mathcal{O}$ is the named operation, $i_t$ and $j_t$ are the two
source-register indices, and $d_t$ is the destination-register index. The source
and destination indices satisfy
\[
    i_t,j_t,d_t\in\{0,\ldots,R-1\}.
\]
The main operation set is
\begin{equation}
    \mathcal{O}
    =
    \{\mathrm{AND},\mathrm{OR},\mathrm{XOR},\mathrm{ADD},
    \mathrm{SUB},\mathrm{SHL},\mathrm{SHR},\mathrm{MUL}\}.
    \label{eq:interface-operation-set}
\end{equation}
The operation mnemonics denote logical AND, logical OR, exclusive OR (XOR),
addition (ADD), subtraction (SUB), shift left (SHL), shift right (SHR) and
multiplication (MUL). These names label fixed continuous operations inspired by
Central Processing Unit (CPU) instruction categories \cite{PattersonHennessyRISCVD}. They do not claim
bit-accurate integer arithmetic in the present implementation.

The reference interpreter reads the two source operands
\begin{equation}
    u_t = r_{t-1,i_t},
    \qquad
    v_t = r_{t-1,j_t},
    \label{eq:interface-operands}
\end{equation}
applies the operation $A_{o_t}$ from the fixed continuous arithmetic logic unit
(ALU) bank, and writes the result only to the destination register. For each
register $q$, the reference transition is
\begin{equation}
r_{t,q} =
\begin{cases}
A_{o_t}(u_t,v_t), & q=d_t,\\
r_{t-1,q}, & q\neq d_t .
\end{cases}
\label{eq:interface-reference-update}
\end{equation}
This write rule is the locality constraint of the interface. One instruction may
change the destination register, while all non-destination registers are
preserved.

The instruction vector used by the learned executor is
\begin{equation}
    x_t =
    \left[
    \mathrm{onehot}(o_t),
    \frac{i_t}{D_R},
    \frac{j_t}{D_R},
    \frac{d_t}{D_R}
    \right],
    \qquad
    D_R=\max(1,R-1).
    \label{eq:interface-encoding}
\end{equation}
Here, $\mathrm{onehot}(o_t)$ is the one-hot encoding of the operation label, and
$D_R$ is the guarded denominator used to normalize register indices. For the main
operation set, this gives an $|\mathcal{O}|+3=11$ dimensional instruction
encoding. The executor receives $x_t$, the previous predicted register file
$\hat r_{t-1}$ and, when enabled, a differentiable memory read vector. It then
predicts operation logits
\begin{equation}
    \ell_t =
    f_{\mathrm{router}}
    \left(
    [\hat r_{t-1,i_t},\hat r_{t-1,j_t},a_t,\mathrm{onehot}(o_t)]
    \right),
    \qquad
    \ell_t\in\mathbb{R}^{|\mathcal{O}|},
    \label{eq:interface-router-logits}
\end{equation}
where $a_t$ is an auxiliary controller projection and
$f_{\mathrm{router}}$ is the operation router. The corresponding soft operation
distribution is
\begin{equation}
    \pi_t = \mathrm{softmax}(\ell_t/\tau),
    \label{eq:interface-soft-gate}
\end{equation}
where $\tau>0$ is the gate temperature. Deterministic evaluation uses the hard
operation
\begin{equation}
    \hat o_t = \arg\max_{k\in\mathcal{O}} \ell_{t,k}.
    \label{eq:interface-hard-gate}
\end{equation}
The index $k$ ranges over the named operations in $\mathcal{O}$. This factorized pathway is related to earlier neural execution and neural
programming models, including recurrent program execution, differentiable
operation selection, supervised neural interpreters, neural random-access memory
and neural algorithm learners
\cite{ZarembaSutskever2014LearningExecute, Neelakantan2015NeuralProgrammer,
ReedDeFreitas2016NPI,Kurach2016NRAM,KaiserSutskever2016NeuralGPU}. The present
interface differs by making the operation gate, destination mask, register
trajectory and matched fixed-point replay explicit evaluation objects.

For each operation $k\in\mathcal{O}$, the executor computes a candidate ALU
output
\begin{equation}
    y_{t,k}=A_k(\hat r_{t-1,i_t},\hat r_{t-1,j_t}).
    \label{eq:interface-candidate-alu}
\end{equation}
Soft execution uses the mixture
\begin{equation}
    \hat y_t^{\mathrm{soft}}
    =
    \sum_{k\in\mathcal{O}}\pi_{t,k}y_{t,k},
    \label{eq:interface-soft-write}
\end{equation}
whereas hard deterministic execution uses
\begin{equation}
    \hat y_t^{\mathrm{hard}}
    =
    y_{t,\hat o_t}.
    \label{eq:interface-hard-write}
\end{equation}
The selected writeback value is denoted by $\hat y_t$. The predicted register
file is updated by the destination mask
\begin{equation}
\hat r_{t,q} =
\begin{cases}
\hat y_t, & q=d_t,\\
\hat r_{t-1,q}, & q\neq d_t .
\end{cases}
\label{eq:interface-model-writeback}
\end{equation}
If a padding mask marks step $t$ as invalid, the implementation leaves the
predicted state unchanged and excludes the step from masked trace and gate
losses. The padding mask is the binary batching mask that separates real program
steps from padded positions.

The interface is both a model pathway and a measurement surface. At each valid
step it exposes the encoded instruction, source operands, destination register,
operation logits, operation probabilities, hard operation choice, predicted
writeback value, predicted register state and reference target. Evaluation
compares these objects with the reference machine using final-state error,
full-trace error, gate agreement, deterministic hard execution and matched
fixed-point replay. Endpoint accuracy alone is not used as the definition of
successful execution.

In the eight-bit quantization-simulated writeback setting (QAT8), the selected
writeback value is projected onto a uniform eight-bit fixed-point grid before it
is written to the destination register. This simulated writeback projection is
related to the broader literature on low-precision and quantized neural
computation
\cite{Courbariaux2015BinaryConnect,Hubara2018QNN,Esser2020LSQ}, while the
matched replay below defines the specific reference semantics used in this
study. For a tensor $x\in[0,1]$ and bit depth
$B$, the projection is
\begin{equation}
    Q_B(x)
    =
    \frac{
    \mathrm{clip}
    \left(
    \mathrm{round}\left(x(2^B-1)\right),
    0,
    2^B-1
    \right)
    }{2^B-1}.
    \label{eq:interface-quantizer}
\end{equation}
Here, $\mathrm{round}$ denotes elementwise nearest-grid rounding, and
$\mathrm{clip}$ limits each component to the interval
$[0,2^B-1]$. QAT8 uses $B=8$. The quantized model writeback is therefore
$Q_8(\hat y_t)$ in the standard QAT8 executor.

The matched fixed-point replay applies the same writeback projection to the
reference interpreter. Its transition is
\begin{equation}
r^{(B)}_{t,q} =
\begin{cases}
Q_B\!\left(A_{o_t}(r^{(B)}_{t-1,i_t},r^{(B)}_{t-1,j_t})\right),
& q=d_t,\\
r^{(B)}_{t-1,q}, & q\neq d_t .
\end{cases}
\label{eq:interface-matched-replay}
\end{equation}
The state $r^{(B)}_t$ denotes the reference trajectory replayed at bit depth
$B$. This replay is an external scoring reference computed after model
execution. It does not consult the model outputs, it is not a runtime fallback,
and it is not full integer-only inference. It tests whether the learned QAT8
executor agrees with the reference machine under the same writeback precision.

Optional memory variants keep the same interface. The gated recurrent unit with
Neural Turing Machine memory (GRU-NTM) executor uses a slot-addressed memory
read vector. In addition, the ValueMemory variant uses a value-addressed memory pathway for
hidden-opcode memory-pressure tasks. At most one memory variant is active in a
trained instance. The exposed inspection objects remain the same, since they are the
operation path, destination writeback, register trajectory and replay semantics.

\subsection{Dataset generation and training}

The shared execution interface in \cref{sec:experimental_programme} is generated from synthetic register-machine programs. Each program is sampled by choosing a task family, length, opcode sequence, source-register indices and destination-register indices under a fixed CPU profile. The principal 16-wide profile uses training lengths 10-60. The original benchmark bundle evaluates held-out lengths 30, 60, 100 and 200, and the long-horizon analysis evaluates the trained instances on the extended grid $L\in\{20,30,40,60,80,100,140,200,260,320,400,500,600,700,800,900,1000\}$. Additionally, the 32-wide extrapolation analyses use the same generator on wider continuous register vectors. Each sample stores instruction features, initial registers, final reference registers, the reference register trace, operation labels, proxy energy and delay traces, and a valid-step mask.
Energy-choice datasets add alternate and minimum-energy traces, alternate instruction streams, total proxy costs and minimum-energy path labels.

The main medium 16-wide experiments use 3000 training programs, 500 validation programs, batch size 16, learning rate $10^{-3}$, 30 epochs, hidden size 64, 32 memory slots and memory dimension of 32. Gumbel gate training uses $\tau$ annealed from 2.0 to 0.5, and the deterministic evaluation uses hard argmax gates. The standard \qat{} runs use 8-bit quantization simulation of the selected \alu{} output before destination writeback, optionally with a 5-epoch warmup. The ValueMemory-\qat{} variants quantize the \alu{} output before the later value-memory blend rather than re-quantizing the final blended destination value.

\subsection{Metrics}
\label{sec:methods_metrics}

Let $m_{n,t}\in\{0,1\}$ be the padding mask and let
\begin{equation}
    N_m=\sum_{n,t}m_{n,t}
\end{equation}
be the number of active program steps in an evaluated batch or pooled benchmark family. For each sequence $n$, $T_n$ denotes its last active timestep. Final MAE is
\begin{equation}
    \MAE_F =
    \frac{1}{N_B R W}\sum_{n=1}^{N_B}\sum_{j=1}^{R}\sum_{k=1}^{W}
    |\hat{r}_{T_n,j,k}^{(n)}-r_{T_n,j,k}^{(n)}|.
\end{equation}
Trace MAE is
\begin{equation}
    \MAE_T =
    \frac{1}{N_m R W}\sum_{n,t}m_{n,t}\sum_{j=1}^{R}\sum_{k=1}^{W}
    |\hat{r}_{t,j,k}^{(n)}-r_{t,j,k}^{(n)}|.
\end{equation}
Gate agreement is
\begin{equation}
    A_G =
    \frac{1}{N_m}\sum_{n,t}m_{n,t}
    \mathbf{1}[\argmax_{o\in\opset}\pi_{t,o}^{(n)}=o_t^{(n)}].
\end{equation}
Tolerance faithfulness is
\begin{equation}
    F_{\epsilon} =
    \frac{1}{N_B}\sum_{n=1}^{N_B}
    \mathbf{1}\left[
    \|\hat{r}_{T_n}^{(n)}-r_{T_n}^{(n)}\|_{\infty}\leq \epsilon
    \right].
\end{equation}
Exact low-precision faithfulness is reported as equality after both predicted and reference register components are projected onto the same scalar quantization grid. The grid resolution is tied to the register-vector profile width $W$. The diagnostic is scalar-component agreement on that grid, not equality of packed integer machine words.
Calibration is computed after flattening active gate predictions. Expected calibration error is
\begin{equation}
    \ECE =
    \sum_{m=1}^{M}\frac{|C_m|}{N_m}
    \left|\mathrm{acc}(C_m)-\mathrm{conf}(C_m)\right|,
\end{equation}
where $C_m$ is confidence bin $m$ over active timesteps. The Brier score is
\begin{equation}
    \Brier =
	    \frac{1}{N_m}\sum_{n,t}m_{n,t}\sum_{o\in\opset}
	    \left(\pi_{t,o}^{(n)}-\mathbf{1}[o=o_t^{(n)}]\right)^2.
\end{equation}
Here $N_m=\sum_{n,t}m_{n,t}$ is the number of active instruction steps. The implemented score sums squared error over the operation dimension rather than dividing by $|\opset|$.
In benchmark reporting, calibration is interpreted for \emph{soft-mode} probabilities. Deterministic hard execution is summarized by argmax gate agreement, and the one-hot hard gates are not treated as evidence of probability calibration.

\subsection{Statistical aggregation}
\label{sec:methods_statistics}

For benchmark tables, each row aggregates over all examples and batches specified by the corresponding benchmark configuration. Where multi-seed sweeps are available, results are summarized as mean and standard deviation:
\begin{equation}
    \bar{x}=\frac{1}{S}\sum_{s=1}^{S}x_s,
    \qquad
    \sigma_x=\sqrt{\frac{1}{S-1}\sum_{s=1}^{S}(x_s-\bar{x})^2}.
\end{equation}
For paired comparisons, such as baseline versus candidate control under the same task set, the paired difference is
\begin{equation}
    \Delta_s = x_s^{\mathrm{candidate}}-x_s^{\mathrm{baseline}}.
\end{equation}
When the sample count is sufficient, a paired $t$ statistic is
\begin{equation}
    t =
    \frac{\bar{\Delta}}
    {s_{\Delta}/\sqrt{S}},
\end{equation}
where $s_{\Delta}$ is the standard deviation of paired differences. These experiments emphasize effect sizes, task-level breakdowns and confidence intervals over formal significance claims when long-horizon control evaluations have limited seed counts. The figures therefore separate stable benchmark effects from regimes where independent replication remains limited and where additional seeds would be required for stronger inferential claims.

For the symbolic-execution benchmark of \cref{fig:main_benchmark}, distribution-free uncertainty and significance are computed from the existing benchmark outputs. Each metric is summarized over the pooled per-length, per-batch evaluation samples of a family, and $95\%$ confidence intervals are obtained by a percentile bootstrap of the mean ($10{,}000$ resamples, fixed seed). Paired metric comparisons within a family (e.g.\ trace versus final \MAE{}) use the Wilcoxon signed-rank test, and two-regime comparisons (e.g.\ the 32-wide extrapolation profile at $L\le320$ versus $L\ge600$) use the Mann--Whitney $U$ test. These statistics are quoted inline where the corresponding claims are made.

\subsection{Diagnostic protocols and extensions}
\label{sec:methods_diagnostics_extensions}

This subsection defines the diagnostic protocols used outside the primary
visible-opcode execution benchmark. These protocols keep the same execution
interface when possible, but change the evaluation setting. They include the
RV32I semantic bridge, architecture baselines, hidden-opcode memory pressure,
spiking-controller variants and rollout-level diagnostic panels.

\begin{figure*}[t]
    \centering
    \includegraphics[width=\textwidth]{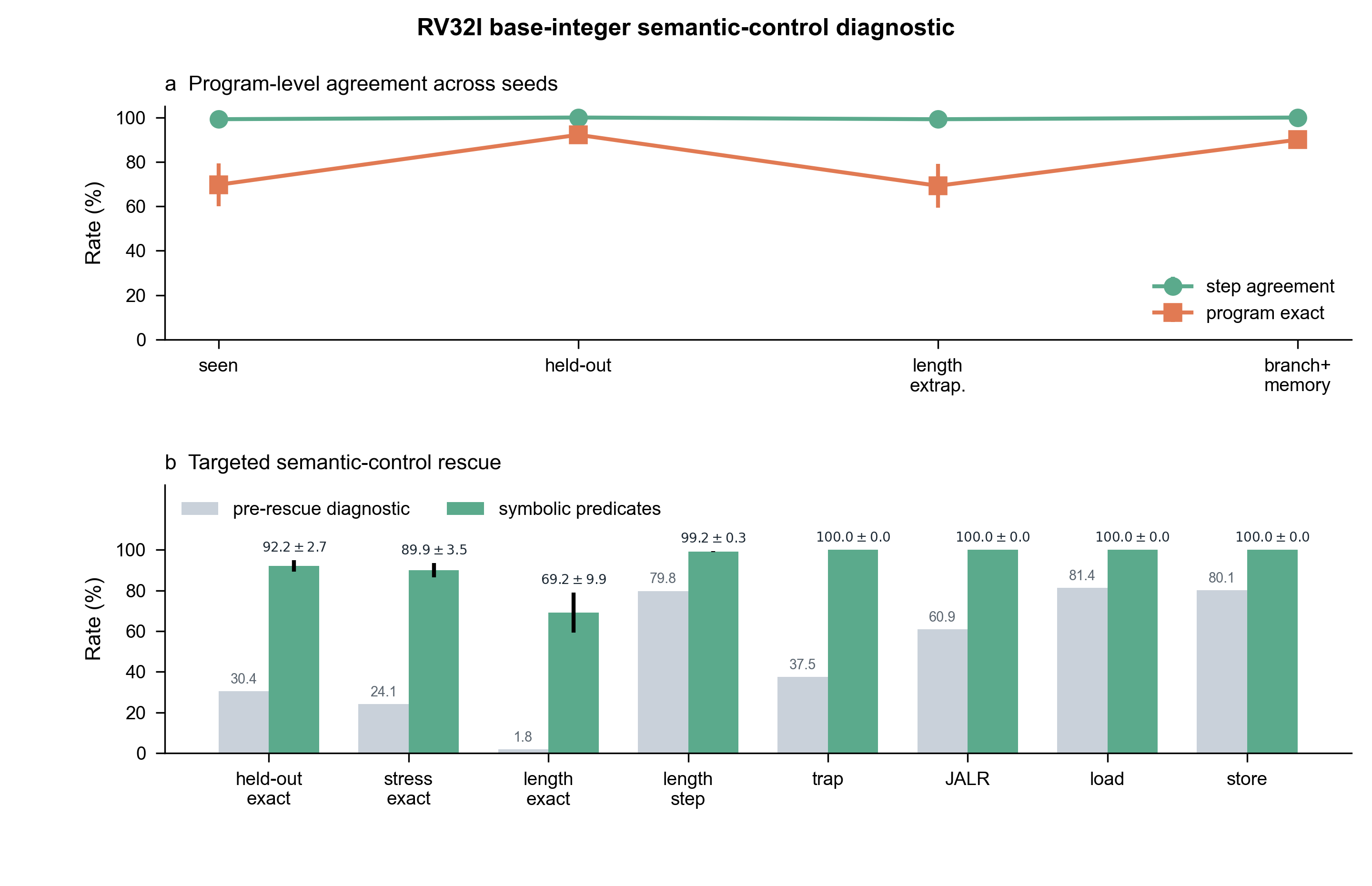}
    \caption{\textbf{RV32I base-integer semantic-control diagnostic.}
    RV32I decode, deterministic audit/harness, learned-control heads and architecture-backed executor definitions are given in the methods section.
    \textbf{(a)} Program-level agreement for the symbolic-predicate controller repeated over seeds 42, 43 and 44. Step agreement is the joint PC/memory/writeback semantic-control accuracy over teacher-trace steps. Program exact requires every semantic-control decision in a sequence to be correct. Error bars show s.d. across seeds.
    \textbf{(b)} Targeted semantic-control rescue. Grey bars show the pre-rescue RV32I learned-control diagnostic values used to identify the weak modes; green bars show the symbolic-predicate multi-seed mean with s.d. error bars. Numeric labels give the pre-rescue percentage for grey bars and mean $\pm$ s.d. across seeds for green bars. Because trap, JALR, load, store and memory-trap mode accuracies are uniformly $100.0\pm0.0\%$ after rescue, they are reported in the Results text rather than repeated as a separate panel. The panel evaluates a base-integer semantic-control bridge and does not constitute official \riscv{} compliance or arbitrary \riscv{} binary execution by the main trained executor.}
    \label{fig:rv32i_bridge}
\end{figure*}

\noindent\textbf{RV32I semantic bridge.}
The RV32I semantic bridge used in \cref{fig:rv32i_bridge} represents each
instruction as an unsigned 32-bit word. The decoder extracts opcode, rd, rs1,
rs2, funct3, funct7 and I/S/B/U/J immediate fields using RV32I bit positions and
sign extension. Here, rd is the destination register and rs1 and rs2 are
source-register fields. Legal base-integer encodings are mapped to instruction
names and instruction families. Unsupported encodings enter an illegal-instruction
trap state.

The machine state is represented by a 32-entry unsigned 32-bit register file, a
byte-addressed memory array, an integer program counter and a string-valued trap
field. Register x0 is reset to zero after every step, enforcing architectural
zero-register invariance. The step function checks for a pre-existing trap and
program-counter alignment before applying instruction semantics. LUI and AUIPC
write upper-immediate values. JAL and JALR write the link register and update the
program counter, with JALR clearing the low target bit. Branches compare signed
or unsigned operands according to funct3 and update the program counter when
taken. Loads and stores compute effective addresses from rs1 plus the relevant
immediate and apply width-specific alignment and bounds checks. OP-IMM and OP
instructions call symbolic \alu{} helpers for addition, subtraction, comparisons,
bitwise operations and shifts. FENCE is treated as a single-hart no-op. ECALL and
EBREAK enter explicit trap states.

\noindent\textbf{RV32I audit and harness.}
The deterministic audit enumerates the implemented RV32I base-integer inventory
and combines direct unit tests with randomized property tests. Direct tests cover
x0 invariance, arithmetic and comparisons, load and store widths, branch targets,
JAL and JALR link behaviour, LUI and AUIPC, alignment traps, FENCE, ECALL,
EBREAK and illegal-instruction handling. Randomized properties test immediate
sign extension, signed and unsigned comparisons, branch target selection, load
and store round trips, sign extension, alignment traps, memory out-of-bounds
traps and random program-level invariants. The full audit used 8,192 generated
programs, 16,384 property cases, lengths 16 to 384, a maximum of 512 execution
steps and 4,096 bytes of memory.

The architectural harness runs raw instruction-word programs through the
deterministic bridge and records a signature-memory region. It initializes
registers and memory deterministically, supports external machine-code programs
and structured test specifications, and can check expected signature prefixes,
trap states and register values. The harness is designed to match the shape of future
architectural compliance testing.

\noindent\textbf{RV32I learned-control diagnostic.}
The learned-control experiment separates deterministic RV32I replay from learned
control prediction. Instruction-level models predict operation name and register
fields from either raw instruction bits or decoded bitfield features.
Semantic-control models predict program-counter mode, memory mode and writeback
mode from instruction and state features. Program-counter modes are fall-through,
branch taken, branch not taken, jump, jump-register and trap. Memory modes are
none, load, store and trap. Writeback modes are none, write rd, x0 suppressed and
trap.

The rescue model reported in \cref{fig:rv32i_bridge} concatenates decoded
bitfield features with symbolic RV32I predicates. These include one-hot
instruction name and family, normalized immediate values, equality and signed or
unsigned comparison predicates, branch-taken predicates, branch, JAL and JALR
target-alignment predicates, load and store effective-address predicates,
memory-width indicators, signed-load indicators, out-of-bounds and misalignment
indicators, program-counter alignment, illegal and trap indicators and rd==0.
The model is a compact multilayer perceptron with separate heads for
program-counter mode, memory mode and writeback mode. The cross-entropy losses are
applied to each control head, with optional class weighting for rare modes.

For the multi-seed diagnostic in \cref{fig:rv32i_bridge}, the symbolic-predicate
semantic-control model was repeated with seeds 42, 43 and 44. Each run used
500,000 training samples, 100,000 validation samples, 4,096 evaluation programs,
96 epochs, batch size 1,024, hidden width 320, learning rate $10^{-3}$, program
lengths 16 to 256, a maximum of 384 execution steps, 4,096 bytes of memory and
semantic class-weight power 0.5. Program-level summaries report mean and s.d.
across seeds. Step agreement is the joint semantic-control accuracy over
program-counter, memory and writeback heads. Program-exact sequence accuracy
requires every program-counter, memory and writeback decision in the evaluated
sequence to be correct. Mode-level summaries stratify predictions by target head
and semantic mode. The targeted rescue panel reports the prespecified weak-mode
metrics, which means the held-out program exact, branch and memory challenge program exact,
length-extrapolation exact, length-extrapolation step agreement, trap,
jump-register, load and store.

\noindent\textbf{Architecture-backed RV32I executor.}
The architecture-backed RV32I executor is a separate derivative of the symbolic
neural CPU design principles rather than the main trained executor. RV32I decode
and predicate features are supplied to a learned micro-operation lowerer, passed
through a \gru{} controller, read through a scalar value-memory lane and routed
through a gated \alu{} operation bank. Output heads predict program-counter mode,
memory mode, writeback mode, auxiliary \alu{} operation, register indices and
\alu{} value. This executor mirrors the article's core ingredients, namely
recurrent control, value-addressed memory and learned \alu{} routing, while
adapting them to RV32I teacher traces generated by the deterministic bridge.

\noindent\textbf{Architecture baselines.}
All comparison architecture variants use the same dataset and benchmark protocol. They are not parameter-matched or memory-interface-matched. Some
models have different parameter counts due to controller architecture. The
comparison is therefore used as an architectural diagnostic, with the no-gate-loss
ablation providing the capacity-preserving test of gate supervision. The
scratchpad baseline predicts traces without explicit gates. Gate metrics are
therefore not computed for that baseline rather than imputed. The no-gate-loss
model uses the main architecture with $\lambda_G=0$, testing whether gate
supervision matters independently of capacity.

\noindent\textbf{Memory-pressure benchmark.}
Memory-pressure tasks hide or reduce opcode cues and require delayed state use.
Splits include seen tasks and seen lengths, seen tasks and held-out lengths,
held-out tasks and seen lengths, and held-out tasks and held-out lengths. The
benchmark changes the inference problem. Instead of reading a visible opcode from
the instruction vector, the model must infer operation identity, address use and
delayed writeback structure from register-state transitions. Degradation on
held-out memory-pressure tasks therefore measures latent program induction under
masked-opcode conditions, not visible-opcode program execution.

\noindent\textbf{Spiking variants.}
Spiking variants replace or augment the controller hidden state with leaky
integrate-and-fire or adaptive leaky integrate-and-fire dynamics. One symbolic
program step is expanded into $K$ micro-steps. The downstream \alu{} and register
pathways remain non-spiking. The design keeps spikes in temporal controller
memory while preserving deterministic non-spiking arithmetic and writeback.

\noindent\textbf{Rollout-level symbolic diagnostics.}
\Cref{fig:selected_stack} reports two trained open-loop rollouts and one
closed-loop evaluation summary. Column~\textbf{A} uses the medium 16-wide \qat{}
trained instance. Column~\textbf{B} uses the memory-pressure \qat{} ValueMemory
trained instance with opcode masking as in the memory-pressure data generator.
Column~\textbf{C} uses the reference factorized multilayer-perceptron candidate
control evaluation log together with the paired candidate-control summary
metrics.

For the open-loop panels, the model receives $x_{1:T}$, $r_0$ and mask
$m_{1:T}$ for a program of length $T$. At each step the \dcpu{} returns gate
logits $\ell_t$, gate probabilities, register trace $r^{\mathrm{pred}}_t$ and
memory addressing weights. The addressing weights correspond to Neural Turing
Machine slots or value-memory slots depending on the architecture.
Operation-gate match $m_t$ and router entropy $H_t$ follow the same definitions
as in \cref{sec:methods_snn_signals} with $\tau=0.5$, hard gates and \qat{}
writeback when enabled. Per-step trace MAE uses
\texttt{trace\_mae\_by\_step}. Rollout exemplars are selected from fixed
candidate seeds by maximizing gate agreement and minimizing mean trace MAE. For
the memory-pressure rollout, structured memory addressing is also required.

Each panel displays the non-flat execution variable for the selected trace. If
$m_t\equiv 1$, the gate and router row reports $\log_{10}(H_t+\varepsilon)$ with
the gate-match summary. If the reference opcode is constant, the opcode strip is
represented by an in-panel label. If operation probabilities or Neural Turing
Machine addressing are uniform, operand routing or per-register trace MAE is
shown instead. Varying reference opcodes use a change-point strip and vertical
guides at $o_t^{\mathrm{ref}}\neq o_{t-1}^{\mathrm{ref}}$ transitions.

For the closed-loop panel, episode-level evaluation records contain final MAE,
success label, return, task family and the beam-selected action sequence. The
beam uses width 4, depth 2 and up to 512 candidates per step. Success is
$\mathrm{MAE}^{\mathrm{final}}\leq 10^{-2}$. The joint scatter uses a
log-binned hexbin for non-success episodes plus a point overlay for successes.
Marginal histograms show densities split by success label, and the legend
reports counts. The per-task bar chart uses the most frequent tasks in the
evaluation log and encodes success rate with a left-to-right colormap plus
numeric percent labels. The paired-lift panel reads the candidate-control
summary table and plots four paired comparisons: reference and learned-\qat{}
backends, each for factorized multilayer perceptron and factorized \alif{}
spiking neural network. Line style and marker distinguish multilayer perceptron
versus spiking neural network. Text annotations report the absolute success-rate
lift in percentage points.

\subsection{Rollout-level spiking signal plots (\texorpdfstring{\cref{fig:snn_signals}}{Supplementary spiking signal figure})}
\label{sec:methods_snn_signals}

\Cref{fig:snn_signals} analyzes the trained spiking ValueMemory instance used for the spiking rollout panels. Memory-pressure reference programs are built and executed by the reference CPU with optional opcode masking. Specifically, when masking is enabled, the operation one-hot prefix of each instruction vector is set to zero while register operands remain visible.

\paragraph{Open-loop forward evaluation (panels \textbf{a--c}).}
For a program of length $T$, the model receives instruction tensor $x_{1:T}$, initial registers $r_0$ and padding mask $m_{1:T}$. At each step the spiking ValueMemory \dcpu{} returns gate logits $\ell_t$, gate probabilities (hard or soft), predicted registers $r^{\mathrm{pred}}_t$, micro-step-averaged spikes $\bar{z}_t\in[0,1]^H$ and membrane voltages $u_t$. Operation-gate match is
\begin{equation}
    m_t = \mathbf{1}\!\left[\arg\max_{o\in\opset}\ell_{t,o}=o_t^{\mathrm{ref}}\right].
\end{equation}
The router entropy uses temperature-scaled softmax probabilities $p_{t,o}=\softmax(\ell_t/\tau)_o$ with default $\tau=0.5$:
\begin{equation}
    H_t = -\sum_{o\in\opset} p_{t,o}\log\!\left(\max(p_{t,o},10^{-12})\right),
\end{equation}
displayed as $\log_{10}(H_t+\varepsilon)$ with $\varepsilon=10^{-4}$. Per-step trace error follows:
\begin{equation}
    \mathrm{MAE}_t=
    \mathrm{mean}_{r,w}\left|r^{\mathrm{pred}}_{t,r,w}-r^{\mathrm{ref}}_{t,r,w}\right|
\end{equation}
over active padded steps. Furthermore, the hidden units are reordered by descending mean spike rate $\mathbb{E}_t[\bar{z}_{t,i}]$ for display. As for the spike heatmaps, we have used values in $[0,1]$ with nearest-neighbour interpolation. The membrane heatmaps use a perceptually uniform sequential colormap with display limits set by the 2nd and 98th percentiles of the shown matrix. The Opcode-change vertical guides are drawn at steps where $o_t^{\mathrm{ref}}\neq o_{t-1}^{\mathrm{ref}}$, and trace reference lines use the stated MAE threshold for the corresponding panel.

\paragraph{Three-dimensional landscape (panel \textbf{c}).}
Panel \textbf{c} plots the membrane matrix $u_{t,i}$ as a surface over $(t,i)$ with the same unit ordering as panel \textbf{a}. Spike events with $\bar{z}_{t,i}>0.5$ are rendered as markers slightly above the surface. The subtitle reports mean spike rate $\rho_{\mathrm{spike}}=(TH)^{-1}\sum_{t,i}\bar{z}_{t,i}$ over the displayed rollout.

\paragraph{Closed-loop episode scatter (panel \textbf{d}).}
Panel \textbf{d} uses the episode-level evaluation log from the reference factorized-\alif{} candidate fine-tuned instance. Each row is one episode of the control environment. Final MAE is
\begin{equation}
    \mathrm{MAE}^{\mathrm{final}}=\mathrm{mean}_{r,w}\left|R_{r,w}-R^{\star}_{r,w}\right|,
\end{equation}
and success is $\mathbf{1}[\mathrm{MAE}^{\mathrm{final}}\leq\epsilon]$ with default $\epsilon=10^{-2}$. Episode mean policy spike rate is the arithmetic mean of per-step values recorded during the episode. Failures are shown with a log-scaled hexbin density, and successes are overlaid as green points with small horizontal jitter near zero MAE. The marginal histograms show normalized densities of final MAE and spike rate split by success label. Evaluation uses candidate action masking ($\leq 512$ actions) and beam search with width~4 and depth~2, matching the candidate-control protocol described above.

\subsection{Closed-loop control objectives}
\label{sec:methods_closed_loop_control}

Closed-loop policies are trained by behaviour cloning followed by actor-critic reinforcement learning. We use A2C and PPO-style variants \citep{Mnih2016A3C,Schulman2017PPO,SuttonBarto2018}. The transition backend is either the reference CPU transition or the learned \qat{} Neural CPU one-step transition. Candidate action sets and beam-style evaluation are search constraints on top of the same transition semantics. Because this protocol optimizes target-reaching actions rather than replaying a supplied reference program, it is treated as a controllability extension to the trace-supervised executor rather than as the primary execution benchmark.

\paragraph{State, action and reward.}
The controller observation is the implemented vector:
\begin{equation}
    s_t =
    \left[
    \mathrm{vec}(R_t),
    \mathrm{vec}(R^\star),
    \mathrm{vec}(R^\star-R_t),
    \frac{H-t}{H},
    \MAE(R_t,R^\star)
    \right],
\end{equation}
where $R_t\in[0,1]^{R\times W}$ is the current register file, $R^\star$ is the target register file and $H$ is the control horizon. A symbolic action is $a_t=(o_t,i_t,j_t,d_t)$ and is flattened exactly as
\begin{equation}
    \mathrm{idx}(a_t)
    =
    (((o_t R+i_t)R+j_t)R+d_t).
\end{equation}
The environment reward matches the implementation:
\begin{equation}
    r_t =
    10\left(\MAE_t-\MAE_{t+1}\right)
    -c_{\mathrm{step}}
    -w_E E(o_t)
    +
    \mathbf{1}_{\mathrm{done}}
    \left[
    \mathbf{1}_{\mathrm{succ}}b_{\mathrm{succ}}
    -
    (1-\mathbf{1}_{\mathrm{succ}})\MAE_{t+1}
    \right],
\end{equation}
with default $c_{\mathrm{step}}=0.01$, $b_{\mathrm{succ}}=1$, $w_E=10^{-3}$ and success threshold $\MAE_{t+1}\leq10^{-2}$. Thus the controller is rewarded for immediate target-error reduction, penalized for step count and proxy operation cost, and given a terminal success bonus or terminal residual-error penalty.

\paragraph{Behaviour cloning.}
Oracle examples are collected from the task generator's symbolic programs. The flat behaviour-cloning term is
\begin{equation}
    \loss_{\mathrm{flat}}(\theta)
    =
    -\frac{1}{N}\sum_{n=1}^{N}
    \log \pi_{\theta}^{\mathcal{C}}(a_n^{\star}\mid s_n),
\end{equation}
where $\pi_{\theta}^{\mathcal{C}}$ denotes the candidate-masked categorical distribution when candidate masking is enabled. For factorized policies, the implemented auxiliary loss decomposes the same action into operation, source-register and destination-register labels:
\begin{equation}
    \loss_{\mathrm{fac}}
    =
    \frac{1}{4}\left[
    \CE(\ell^{\mathrm{op}},o^\star)
    +\CE(\ell^{a},i^\star)
    +\CE(\ell^{b},j^\star)
    +\CE(\ell^{\mathrm{dst}},d^\star)
    \right].
\end{equation}
Spiking policies add a rate regularizer $(\rho-\rho^\star)^2$, where $\rho$ is the policy's last recorded mean spike rate and $\rho^\star$ is the configured target spike rate. The behaviour-cloning objective is therefore
\begin{equation}
    \loss_{\mathrm{BC}}
    =
    \loss_{\mathrm{flat}}
    +\lambda_{\mathrm{fac}}\loss_{\mathrm{fac}}
    +\lambda_{\mathrm{spike}}(\rho-\rho^\star)^2.
\end{equation}

\paragraph{Actor-critic objectives.}
For an episode reward sequence, discounted returns are
\begin{equation}
    G_t=\sum_{k=t}^{T}\gamma^{k-t}r_k,
\end{equation}
and A2C advantages $\hat{A}_t=G_t-V_{\phi}(s_t)$ without normalizing them. The implemented A2C loss is
\begin{equation}
    \loss_{\mathrm{A2C}}
    =
    -\frac{1}{T}\sum_t
    \log\pi_{\theta}^{\mathcal{C}}(a_t\mid s_t)\,\mathrm{stopgrad}(\hat{A}_t)
    +c_v\frac{1}{T}\sum_t(V_{\phi}(s_t)-G_t)^2
    -c_h\frac{1}{T}\sum_t H(\pi_{\theta}^{\mathcal{C}}(\cdot\mid s_t))
    +\lambda_{\mathrm{spike}}(\rho-\rho^\star)^2.
\end{equation}
For PPO, the collected trajectory log probabilities define the behaviour policy. Advantages are normalized when the trajectory has more than one step, with a
small numerical stabilizer. We define the PPO likelihood ratio between the
current candidate-masked policy and the candidate-masked behaviour policy that
produced the logged trajectory as
\begin{equation}
    \eta_t^{\mathrm{PPO}}(\theta)
    =
    \frac{\pi_{\theta}^{\mathcal{C}}(a_t\mid s_t)}
    {\pi_{\mathrm{beh}}^{\mathcal{C}}(a_t\mid s_t)}
    =
    \exp\left[
    \log\pi_{\theta}^{\mathcal{C}}(a_t\mid s_t)
    -
    \log\pi_{\mathrm{beh}}^{\mathcal{C}}(a_t\mid s_t)
    \right].
    \label{eq:ppo_likelihood_ratio}
\end{equation}
Here, $\pi_{\theta}^{\mathcal{C}}$ is the current candidate-masked policy,
$\pi_{\mathrm{beh}}^{\mathcal{C}}$ is the candidate-masked behaviour policy
recorded during rollout collection, $a_t$ is the logged action and $s_t$ is the
controller state. Values greater than one mean that the current policy assigns
higher probability to the logged action than the behaviour policy did, whereas
values below one mean that it assigns lower probability.
the implemented clipped actor term is
\begin{equation}
    \loss_{\mathrm{PPO,actor}}
    =
    -\frac{1}{T}\sum_t
    \min\left(
    \eta_t^{\mathrm{PPO}}(\theta)\hat A_t,\,
    \mathrm{clip}\!\left(
    \eta_t^{\mathrm{PPO}}(\theta),
    1-\epsilon_{\mathrm{clip}},
    1+\epsilon_{\mathrm{clip}}
    \right)\hat A_t
    \right),
    \label{eq:ppo_actor}
\end{equation}
with the same value, entropy and spike-rate terms as above.

\paragraph{Factorized policy heads.}
The strongest closed-loop runs use factorized action heads. For a symbolic action $a=(o,i,j,d)$, corresponding to operation, first source register, second source register and destination register, the factorized head scores the flat action by additive component logits,
\begin{equation}
    L_{\theta}(o,i,j,d\mid s)
    =
    \ell_{\theta}^{\mathrm{op}}(o\mid s)
    +
    \ell_{\theta}^{a}(i\mid s)
    +
    \ell_{\theta}^{b}(j\mid s)
    +
    \ell_{\theta}^{\mathrm{dst}}(d\mid s).
\end{equation}
These joint logits are flattened back to the original action vocabulary before softmax, candidate masking and beam evaluation. This formalism covers the multilayer-perceptron, spiking-\alif{} and graph-based factorized policies used in the closed-loop experiments.

\paragraph{Candidate control.}
Candidate control changes the action distribution but not the transition semantics. The candidate generator first selects high-error destination registers, always inserts simple symbolic primitives, and then ranks one-step actions by
\begin{equation}
    S_{\mathrm{cand}}(a)
    =
    \MAE_{\mathrm{after}}(a)
    -0.15\max(0,\MAE_{\mathrm{before}}-\MAE_{\mathrm{after}}(a))
    +0.005\,\mathrm{idx}(o).
\end{equation}
The best actions up to the configured candidate budget are retained. Invalid actions receive logit of $-10^9$ before softmax. During evaluation, a beam of width $B_{\mathrm{beam}}$ maintains the highest-scoring partial action sequences under the cumulative score
\begin{equation}
    \sum_{\tau=t}^{t+D-1}
    \left[
    \log\pi_{\theta}^{\mathcal{C}}(a_\tau\mid s_\tau)
    -10\,\MAE(F(s_\tau,a_\tau),R^\star)
    -w_E E(a_\tau)
    +\mathbf{1}[\MAE(F(s_\tau,a_\tau),R^\star)\leq\epsilon]
    \right],
\end{equation}
where $D$ is the configured beam depth, i.e.\ exactly $D$ score terms starting at $\tau=t$ (one initial-action term plus $D-1$ expansion terms). This is a constrained search procedure over the learned policy, not a new oracle.

\subsection{Mathematical and algorithmic details}

All symbols below refer to tensors or operations in the reported simulator, with assumptions restricted to the simulation and replay protocols evaluated here.

\subsubsection{Symbolic CPU state}

Let the register width be $W\in\{8,16,32\}$, with the main experiments using $W=16$. In this simulator, width is the length of a normalized continuous register vector, not an integer bit-vector. The register file contains $R$ registers,
\begin{equation}
    r_t = (r_{t,1},\ldots,r_{t,R}) \in [0,1]^{R\times W}.
\end{equation}
The notation ``8-bit'', ``16-bit'' and ``32-bit'' follows the profile names used during experimentation, but the reference transition uses continuous vectors in $[0,1]$ and continuous operation proxies. Quantization experiments then project \alu{} writeback values onto an 8-bit grid.

We have used the following operation set:
\begin{equation}
    \opset = \{\mathrm{AND},\mathrm{OR},\mathrm{XOR},\mathrm{ADD},\mathrm{SUB},\mathrm{SHL},\mathrm{SHR},\mathrm{MUL}\},
\end{equation}
with extended task families using combinations and memory-mediated patterns. Each instruction is a tuple
\begin{equation}
    \xi_t = (o_t, i_t, j_t, d_t),
\end{equation}
where $o_t\in\opset$ is the opcode, $i_t,j_t,d_t\in\{0,\ldots,R-1\}$ are zero-based source and destination register indices. Dataset context such as padding masks, task labels, energy-choice policies and alternative traces is stored in separate arrays, not inside the encoded instruction vector.

The deterministic reference-interpreter transition used to generate the training and evaluation trajectories is $r_t = \mathcal{T}(r_{t-1},\xi_t)$, where the reference interpreter itself does not use neural memory. For each instruction, $r_{t,d_t} = \mathrm{ALU}_{o_t}(r_{t-1,i_t}, r_{t-1,j_t})$ and $r_{t,j}=r_{t-1,j}$ for $j\neq d_t$. The implemented continuous operation bank is given by:
\begin{align}
    A_{\mathrm{AND}}(u,v) &= \clip(u\odot v,0,1),\\
    A_{\mathrm{OR}}(u,v) &= \max(u,v),\\
    A_{\mathrm{XOR}}(u,v) &= |u-v|,\\
    A_{\mathrm{ADD}}(u,v) &= \clip(u+v,0,1),\\
    A_{\mathrm{SUB}}(u,v) &= \clip(u-v,0,1),\\
    A_{\mathrm{SHL}}(u,v) &= \clip(2u,0,1),\\
    A_{\mathrm{SHR}}(u,v) &= 0.5u,\\
    A_{\mathrm{MUL}}(u,v) &= \clip(u\odot v,0,1).
\end{align}
These names are symbolic labels for differentiable CPU-like operations. They are not claims of exact integer bitwise execution in the present implementation.
The training dataset contains the full reference trace
\begin{equation}
    \left\{X_{\mathrm{instr}}^{(n)}, r_0^{(n)}, r_{1:T}^{(n)}, r_T^{(n)}, o_{1:T}^{(n)}, e_{1:T}^{(n)}, \delta_{1:T}^{(n)}, m_{1:T}^{(n)}\right\}_{n=1}^{N},
\end{equation}
where $X_{\mathrm{instr}}$ is the encoded instruction sequence, $e_t$ and $\delta_t$ are simulation proxy energy and delay traces, and $m_t$ is the padding mask.

\subsubsection{Instruction encoding}

Each instruction is encoded as the concatenation of an operation one-hot vector and normalized source/destination register indices. Let $D_R=\max(1,R-1)$ denote the guarded denominator used for register-index normalization:
\begin{equation}
    x_t =
    \left[
    \onehot(o_t);
    \frac{i_t}{D_R};
    \frac{j_t}{D_R};
    \frac{d_t}{D_R}
    \right]\in\R^{|\opset|+3}.
\end{equation}
The model then applies a learned linear encoder,
\begin{equation}
    \phi_t=\mathrm{ReLU}(W_{\mathrm{enc}}x_t+b_{\mathrm{enc}}).
\end{equation}
Decoding rounds and clips the final three normalized entries:
\begin{equation}
    (\hat{i}_t,\hat{j}_t,\hat{d}_t)
    =
    \clip\left(\round(D_R x_{t,-3:}),0,R-1\right).
\end{equation}
When padding is present, a binary mask $m_t\in\{0,1\}$ determines whether the step updates the register state and contributes to masked trace/gate losses. The implementation masks the state update and loss terms rather than redefining the logits themselves:
\begin{equation}
    \hat{r}_t = m_t\,\hat{r}_t^{\mathrm{step}} + (1-m_t)\,\hat{r}_{t-1},
    \qquad
    \loss_G \propto \sum_t m_t\,\CE(\ell_t,o_t).
\end{equation}
Here, $\hat{r}_t^{\mathrm{step}}$ is the register state that would be produced
by executing the instruction at step $t$ before applying the padding mask.
When $m_t=1$, the step contains a real instruction, so the candidate update is
kept and the gate loss contributes to training. When $m_t=0$, the position is
only a batching placeholder used to align programs of different lengths. In
that case, the previous register state is copied forward and the gate target at
that position is ignored.

The logits $\ell_t$ are not redefined for padded steps. They may still be
computed by the unrolled network, but the mask prevents the resulting candidate
update from changing the register state and prevents that step from contributing
to the supervised losses. Padding is therefore a batching convention, not an
additional symbolic instruction and not a learned no-op. In batched notation,
the same rule is applied using $m_{n,t}$, and losses are normalized over the
active-step count
\[
N_m=\sum_{n,t}m_{n,t}.
\]
The effective final state for sequence $n$ is taken at its last active step,
\[
T_n=\max\{t\mid m_{n,t}=1\}.
\]

\subsubsection{Controller recurrence}

The reference model uses a standard \gru{} controller \citep{Cho2014GRU}. Let $q_t$ be the \ntm{} read vector and $\phi_t$ the encoded instruction. The recurrent input is
\begin{equation}
    g_t = [\phi_t; q_t].
    \label{eq:gru-input}
\end{equation}
The previous register values do not enter the GRU input directly in the implemented primary model. They enter the \alu{} router through the selected operands. The hidden state evolves according to the update-gate convention, in which the update gate $z_t$ weights the \emph{previous} hidden state rather than the candidate:
\begin{align}
    z_t &= \sigma(W_z g_t + U_z h_{t-1}+b_z),\\
    q_t^{\mathrm{gru}} &= \sigma(W_q g_t + U_q h_{t-1}+b_q),\\
    \tilde{h}_t &= \tanh(W_h g_t + U_h(q_t^{\mathrm{gru}}\odot h_{t-1})+b_h),\\
    h_t &= (1-z_t)\odot \tilde{h}_t+z_t\odot h_{t-1}.
\end{align}
The architecture comparison substitutes this controller with transformer, temporal convolution, graph, temporal graph and state-space alternatives \citep{Vaswani2017Attention,Bai2018TCN,Kipf2017GCN,Battaglia2018GraphNetworks,Gu2022S4}.

\subsubsection{Operation gates}

The gated \alu{} router emits logits over the operation set from the
concatenated source operands, auxiliary controller projection and opcode hint:
\begin{equation}
    u_t=\hat{r}_{t-1,i_t},
    \qquad
    v_t=\hat{r}_{t-1,j_t},
\end{equation}
where $u_t,v_t\in[0,1]^W$ are operand vectors read from the current predicted
register state, and $i_t,j_t$ are the corresponding source-register indices.
The router is
\begin{equation}
    \ell_t =
    f_{\mathrm{router}}\!\left(
    [u_t;v_t;\mathrm{aux}_t;\onehot(o_t)]
    \right),
    \qquad
    \ell_t\in\R^{|\opset|}.
\end{equation}
Here,
\begin{equation}
    \mathrm{aux}_t
    =
    \tanh(W_{\mathrm{aux}}h_t+b_{\mathrm{aux}}),
\end{equation}
and $f_{\mathrm{router}}$ is a two-layer multilayer perceptron with ReLU
activation.

The opcode one-hot vector $\onehot(o_t)$ is available to the router in the
visible-opcode execution regime. Gate agreement in this setting therefore
measures whether the model preserves and executes the supplied symbolic
instruction through the named operation path. It does not measure recovery of
an unobserved opcode. Operation inference without this cue is evaluated
separately in the hidden-opcode memory-pressure benchmark.

Soft execution uses the temperature-scaled gate distribution
\begin{equation}
    \pi_t^{\mathrm{soft}}
    =
    \softmax(\ell_t/\tau),
\end{equation}
where $\tau>0$ is the gate temperature. During stochastic training, the
Gumbel--Softmax relaxation is used
\citep{Jang2017GumbelSoftmax,Maddison2017Concrete}:
\begin{equation}
    \pi_{t,k}^{\mathrm{gumbel}}
    =
    \frac{
        \exp\!\left((\ell_{t,k}+g_{t,k})/\tau\right)
    }{
        \sum_j
        \exp\!\left((\ell_{t,j}+g_{t,j})/\tau\right)
    },
\end{equation}
with
\begin{equation}
    g_{t,k}
    =
    -\log[-\log q_{t,k}],
    \qquad
    q_{t,k}\sim\mathrm{Uniform}(0,1).
\end{equation}
The auxiliary variable $q_{t,k}$ is used here to avoid confusion with the
operand vector $u_t$.

Hard deterministic evaluation uses
\begin{equation}
    \hat{o}_t
    =
    \argmax_k \ell_{t,k},
    \qquad
    \pi_{t,k}^{\mathrm{hard}}
    =
    \mathbf{1}[k=\hat{o}_t].
\end{equation}
This separation provides a stochastic differentiable routing mechanism during
training and a deterministic, auditable symbolic operation path during
evaluation.

\paragraph{Straight-through hardening.}
For diagnostic runs using deterministic straight-through hardening, the gate
can be written as
\begin{equation}
    \pi_t^{\mathrm{ST}}
    =
    \sg\!\left(
        \pi_t^{\mathrm{hard}}-\pi_t^{\mathrm{soft}}
    \right)
    +
    \pi_t^{\mathrm{soft}},
\end{equation}
where $\sg(\cdot)$ stops gradients. The forward pass uses the discrete hard
gate, while the backward pass follows the soft relaxation. Deterministic
hardening and stochastic Gumbel training serve different purposes rather than
representing competing scientific claims. Gumbel--Softmax supplies a
stochastic differentiable routing mechanism during optimization, while
deterministic gates provide reproducible execution audits. Hard execution is
therefore treated as the interpretable evaluation and deployment-facing mode.

\paragraph{Gate margin.}
Let
\begin{equation}
    k_t^{(1)}
    =
    \argmax_k \pi_{t,k}^{\mathrm{soft}}
\end{equation}
denote the most probable operation, and let
\begin{equation}
    k_t^{(2)}
    =
    \argmax_{j\neq k_t^{(1)}}
    \pi_{t,j}^{\mathrm{soft}}
\end{equation}
denote the second-most probable operation. The gate confidence margin is
\begin{equation}
    \mu_t
    =
    \pi_{t,k_t^{(1)}}^{\mathrm{soft}}
    -
    \pi_{t,k_t^{(2)}}^{\mathrm{soft}}.
\end{equation}
A large margin indicates that the selected symbolic operation is well
separated from the nearest alternative. The margin is therefore a diagnostic
of routing confidence and sensitivity to perturbations of the router logits or
gate probabilities. In the present QAT8 implementation, quantization is applied
to the selected ALU value or writeback pathway rather than to the router
logits. Gate margin should consequently not be interpreted as a direct bound
on writeback-quantization error.

\subsubsection{Mixture ALU}

For each possible operation $k\in\opset$, the model computes a candidate register result using the same continuous operation bank as the reference interpreter:
\begin{equation}
    y_{t,k} =
    A_k\!\left(\hat{r}_{t-1,i_t},\hat{r}_{t-1,j_t}\right),
\end{equation}
where $A_k$ is not a learned arithmetic head in the primary executor. This term is a fixed differentiable operation. Soft execution writes
\begin{equation}
    \hat{y}_t^{\mathrm{soft}} = \sum_{k\in\opset}\pi_{t,k}y_{t,k}.
\end{equation}
Hard execution writes
\begin{equation}
    \hat{y}_t^{\mathrm{hard}} = y_{t,\hat{o}_t}.
\end{equation}
The destination update is masked:
\begin{equation}
    \hat{r}_{t,j} =
    (1-w_{t,j})\hat{r}_{t-1,j} + w_{t,j}\hat{y}_t,
    \qquad
    w_{t,j}=\mathbf{1}[j=d_t].
\end{equation}
For padding steps, $\hat{r}_t=\hat{r}_{t-1}$.

\paragraph{Locality of the write rule.}
The write equation enforces a CPU-like locality constraint: one instruction updates the destination register and leaves non-destination registers unchanged. Without this constraint, a black-box sequence model can distribute a small amount of error correction across all registers. Such behaviour may improve final loss but destroys the auditability of the transition. The write mask therefore acts as an inductive bias and as a verification surface. Trace faithfulness can be decomposed into destination error and preservation error:
\begin{align}
    \MAE_{\mathrm{dst}} &=
    \frac{1}{N_m W}\sum_{n,t,k}m_{n,t}
    |\hat{r}_{t,d_{n,t},k}^{(n)}-r_{t,d_{n,t},k}^{(n)}|,\\
    \MAE_{\mathrm{preserve}} &=
    \frac{1}{N_m (R-1) W}\sum_{n,t,j\neq d_{n,t},k}m_{n,t}
    |\hat{r}_{t,j,k}^{(n)}-r_{t,j,k}^{(n)}|.
\end{align}
Here $N_m=\sum_{n,t}m_{n,t}$ is the active-step count, and $d_{n,t}$ is the zero-based destination register index decoded for sample $n$ at step $t$.
A faithful executor should make both quantities small, and a final-state imitator can hide preservation errors until they accumulate.

\subsubsection{Implemented differentiable ALU bank}

The \alu{} bank is shared by the reference interpreter and the model. The model does not learn independent operation-specific arithmetic heads for the main executor. Instead, it learns the router that chooses among fixed continuous operations:
\begin{equation}
    Y_t =
    \left[
    u_t\odot v_t,\ \max(u_t,v_t),\ |u_t-v_t|,\ u_t+v_t,\ u_t-v_t,\ 2u_t,\ 0.5u_t,\ u_t\odot v_t
    \right],
\end{equation}
followed by clipping to $[0,1]$. The operation labels \texttt{AND}, \texttt{OR}, \texttt{XOR}, \texttt{ADD}, \texttt{SUB}, \texttt{SHL}, \texttt{SHR} and \texttt{MUL} name these continuous CPU-like operations. This point is essential for interpretability: the present study tests faithful execution of the differentiable reference interpreter, not exact emulation of integer bitwise arithmetic.
The current continuous proxies also make \texttt{AND} and \texttt{MUL} identical at the value-transition level, because both are implemented as elementwise multiplication. They remain distinct supervised symbolic labels in gate space, but not distinct arithmetic maps in this version of the reference interpreter.

\subsubsection{Differentiable memory}

Memory-augmented variants use an \ntm-style memory \citep{Graves2014NTM,Graves2016DNC}. Let $M_t\in\R^{S\times D}$ contain $S$ slots of dimension $D$. A content key $k_t$ and strength $\beta_t$ define content weights
\begin{equation}
    c_{t,i} =
    \frac{\exp(\beta_t\,\cos(k_t,M_{t-1,i}))}
    {\sum_{j=1}^{S}\exp(\beta_t\,\cos(k_t,M_{t-1,j}))}.
\end{equation}
The memory-addressing rule then applies a learned three-way circular shift mixture over left, stay and right positions:
\begin{equation}
    \alpha_t =
    s_{t,-1}\,\mathrm{roll}_{+1}(c_t)
    +s_{t,0}\,c_t
    +s_{t,+1}\,\mathrm{roll}_{-1}(c_t),
    \qquad
    s_t=\softmax(W_s h_t+b_s).
\end{equation}
The read vector is
\begin{equation}
    q_t = \sum_{i=1}^{S}\alpha_{t,i}M_{t-1,i}.
\end{equation}
Write operations use erase and add vectors $e_t,v_t^{\mathrm{add}}\in\R^D$:
\begin{equation}
    M_{t,i}=M_{t-1,i}\odot(1-\omega_{t,i}e_t)+\omega_{t,i}v_t^{\mathrm{add}},
\end{equation}
where $\omega_t$ is a learned write distribution. No-memory ablations set $q_t=0$ and skip the memory update, while value-memory variants expose value-addressable pathways for hidden-opcode memory-pressure tasks.

\subsubsection{Quantization-simulated execution}

The \qat{} mode applies quantization simulation to the selected \alu{} output. In the standard models this tensor is the value written to the destination register, so it is also the writeback tensor. In value-memory variants, the same round-and-clamp quantization simulator is applied before the later value-memory blend, which is clamped but not re-quantized. For a tensor $x\in[0,1]$ and $B$ quantization bits, the projection is
\begin{equation}
    Q_B(x) =
    \frac{
    \clip\left(\round(x(2^B-1)),0,2^B-1\right)
    }{2^B-1}.
\end{equation}
For \qat{}, $B=8$. The simulator uses this direct round-and-clamp projection. Therefore, it does not implement signed zero-points, per-channel weight quantization or a custom straight-through estimator for the quantizer. Consequently, once quantization simulation is active, gradients through rounded \alu{} values are not replaced by a custom surrogate. The term \qat{} in this manuscript therefore denotes training/evaluation with quantized forward \alu{} simulation and optional warmup, not a complete integer-only deployment recipe. QAT warmup delays quantization simulation for the first $E_w$ epochs, allowing the continuous model to learn stable gate and trace structure before low-precision \alu{} simulation is introduced.

The key evaluation distinction is between continuous-reference scoring and matched fixed-point replay. Continuous-reference scoring compares the \qat{} model against the full-precision reference trace. Matched replay quantizes the reference semantics to the same fixed-point representation used by the quantization-simulated writeback. A low-precision model may show continuous drift yet be exact under the replay semantics. This distinction is essential for defensible device-facing claims.

\paragraph{Matched fixed-point replay criterion.}
Let $Q_B$ be the fixed-point projection induced by the writeback quantizer and let $\mathcal{T}$ be the continuous reference transition. The matched replay replays the same instruction stream, leaves initial registers at input precision and quantizes each executed destination update:
\begin{equation}
    \mathcal{T}_{B}(r,x)
    =
    \mathrm{Write}_{d}\!\left(r,Q_B(A_o(r_i,r_j))\right).
\end{equation}
The \qat{} model transition is $F_{\theta,B}$. We say that a trajectory is replay faithful if
\begin{equation}
    F_{\theta,B}^{(t)}(r_0,x_{1:t})
    =
    \mathcal{T}_{B}^{(t)}(r_0,x_{1:t})
    \quad
    \forall t\leq T.
\end{equation}
Continuous-reference MAE instead compares $F_{\theta,B}^{(t)}$ with $\mathcal{T}^{(t)}$. The two targets differ by the quantization residual
\begin{equation}
    \epsilon_t =
    \mathcal{T}^{(t)}(r_0,x_{1:t})
    -
    \mathcal{T}_{B}^{(t)}(r_0,x_{1:t}).
\end{equation}
By the triangle inequality,
\begin{equation}
    \|F_{\theta,B}^{(t)}-\mathcal{T}^{(t)}\|_1
    \leq
    \|F_{\theta,B}^{(t)}-\mathcal{T}_{B}^{(t)}\|_1
    +
    \|\epsilon_t\|_1.
\end{equation}
Thus non-zero continuous MAE is expected even when the first term is zero. The manuscript therefore reports both continuous-reference drift and matched fixed-point replay faithfulness.

\paragraph{Quantization error bound.}
For a uniform scalar quantizer with step size $s$, each scalar projection satisfies
\begin{equation}
    |Q(x)-x|\leq \frac{s}{2}
\end{equation}
away from saturation. If a register vector has $D$ scalar components, the per-register $\ell_1$ projection error is bounded by $Ds/2$. Saturation violates this simple bound, which is why the implementation records the deployment profile and uses matched-reference scoring rather than relying only on an analytic quantization bound.

\subsubsection{Training objective}

The full loss is
\begin{equation}
    \loss =
    \lambda_F\loss_F+
    \lambda_T\loss_T+
    \lambda_G\loss_G+
    \lambda_H\loss_H+
    \lambda_S\loss_S+
    \lambda_E\loss_E+
    \lambda_D\loss_D+
    \lambda_Z\loss_Z.
\end{equation}
Let $m_{n,t}\in\{0,1\}$ be the padding mask and let
\begin{equation}
    N_m=\sum_{n,t}m_{n,t}.
\end{equation}
Training uses global masked means over active timesteps and tensor elements, not a per-sequence average. The final-state loss is
\begin{equation}
    \loss_F =
    \frac{1}{N_B R W}\sum_{n=1}^{N_B}\left\|\hat{r}_{T_n}^{(n)}-r_{T_n}^{(n)}\right\|_1,
\end{equation}
where $N_B$ is the mini-batch size. The trace loss is
\begin{equation}
    \loss_T =
    \frac{
    \sum_{n,t}m_{n,t}
    \left\|\hat{r}_{t}^{(n)}-r_t^{(n)}\right\|_1
    }{N_m R W}.
\end{equation}
Gate supervision is cross-entropy over the active flattened timesteps:
\begin{equation}
    \loss_G =
    -\frac{1}{N_m}
    \sum_{n,t}m_{n,t}\log \pi_{t,o_t}^{(n)}.
\end{equation}
Entropy regularization is
\begin{equation}
    \loss_H =
    -\frac{1}{N_m}
    \sum_{n,t}m_{n,t}
    \sum_{k\in\opset}\pi_{t,k}^{(n)}\log \pi_{t,k}^{(n)}.
\end{equation}
Because the total loss adds $+\lambda_H\loss_H$ with $\lambda_H>0$ in the reported runs, this term penalizes high-entropy gates and encourages sharper operation choices. It is not an entropy-bonus exploration term.
Gate-smoothness regularization is the active-pair mean of the mean absolute operation-wise difference:
\begin{equation}
    \loss_S =
    \frac{
    \sum_{n,t\geq 2}m_{n,t}m_{n,t-1}
    \frac{1}{|\opset|}
    \sum_{k\in\opset}
    \left|\pi_{t,k}^{(n)}-\pi_{t-1,k}^{(n)}\right|
    }{
    \sum_{n,t\geq 2}m_{n,t}m_{n,t-1}
    }.
\end{equation}
The denominator is clamped to a positive value so that length-one batches return a zero smoothness penalty rather than a division error.
Energy and delay proxy terms are optional:
\begin{equation}
    \loss_E =
    \frac{\sum_{n,t}m_{n,t}\sum_k \pi_{n,t,k}E_k}{\sum_{n,t}m_{n,t}},
    \qquad
    \loss_D =
    \frac{\sum_{n,t}m_{n,t}\sum_k \pi_{n,t,k}D_k}{\sum_{n,t}m_{n,t}}.
\end{equation}
For spiking variants, spike-rate regularization is
\begin{equation}
    \loss_Z =
    \left|
    \frac{1}{N_m H}\sum_{n,t,i}m_{n,t}\bar{z}_{n,t,i}-\rho_{\mathrm{target}}
    \right|.
\end{equation}
Here $\bar{z}_{n,t,i}$ is the micro-step-averaged spike activity returned for hidden unit $i$ at program step $t$, $H$ is the number of spiking hidden units, and $N_m=\sum_{n,t}m_{n,t}$. This absolute deviation defines the spike-rate regularizer, which means that reported spike rates are therefore step-level averages over internal micro-steps, not raw micro-step event counts.
Unless otherwise stated, the main 16-wide runs use
\begin{equation}
    \lambda_F=1,\quad \lambda_T=0.5,\quad \lambda_G=1,\quad
    \lambda_H=10^{-3},\quad \lambda_S=10^{-3}.
\end{equation}

\subsubsection{Training and evaluation protocol}

The training procedure can be written without relying on implementation-specific names:
\begin{enumerate}[leftmargin=1.4em]
    \item Sample a mini-batch of programs, initial registers, reference traces and masks.
    \item Encode each instruction into $\phi_t$ and initialize the model state with $r_0$ and memory $M_0$.
    \item For each valid timestep, compute memory read $q_t$, recurrent hidden state $h_t$, gate logits $\ell_t$, gate distribution $\pi_t$ and candidate \alu{} outputs $y_{t,k}$.
    \item Apply soft, Gumbel or hard gate selection according to the training/evaluation mode.
    \item Write the selected output to the destination register and update differentiable memory if enabled.
    \item Accumulate final, trace, gate, entropy, smoothness, energy, delay and spike penalties.
    \item Backpropagate through the unrolled computation graph. For standard \qat{} runs, quantization simulation of the selected \alu{} writeback is active after the warmup schedule. Therefore, for ValueMemory-\qat{} runs, the quantized \alu{} value is blended with the value-memory read before the final clamp.
    \item At evaluation time, disable stochastic sampling, select hard deterministic gates and report final, trace, gate, calibration, robustness and proxy-cost metrics.
\end{enumerate}
Every valid timestep therefore has both a register-state target and a symbolic operation label.

\subsubsection{Benchmark taxonomy and comparative architecture interface}

The benchmark suite is grouped into four task families. The first family
contains direct symbolic execution tasks such as random \alu{} programs, add
chains, sub chains and mixed arithmetic. The second family contains
algorithmic held-out tasks such as parity, prefix sum, running maximum,
reverse, small sort and function composition. The third family contains
memory-pressure tasks such as delayed copy, associative lookup, pointer
chasing and stack-style reverse. The fourth family contains
equivalent-program energy-choice tasks such as redundant noop elimination,
identity-op skipping, low-power shift-add, memory-copy minimum-energy paths
and micro-op fusion.

The sequence of task families increases mechanistic pressure. Exact results on
the first family establish the executor under direct visible-opcode symbolic
execution. The later families test algorithmic generalization, delayed memory
use and selection among equivalent symbolic paths with different proxy costs.
Degradation on these later families identifies where stronger memory and
control mechanisms are required.

These task families are evaluated using a controlled set of architecture
variants rather than a collection of unrelated baselines. The architectures
share the same forward input/output interface:
\begin{equation}
    \mathrm{model}_{\theta}
    \!\left(
        X_{\mathrm{instr}},
        X_{\mathrm{regs0}},
        m
    \right)
    \mapsto
    \left\{
        \hat{r}_T,
        \hat{r}_{1:T},
        \ell_{1:T},
        \pi_{1:T},
        \alpha^{\mathrm{read}}_{1:T},
        \alpha^{\mathrm{write}}_{1:T},
        z_{1:T}^{\mathrm{optional}}
    \right\}.
\end{equation}

The model output contains the final register state, the complete register
trace, gate logits, gate probabilities, memory read weights, memory write
weights and optional spike traces. Energy and delay are not predicted tensors
returned by the model. They are computed post hoc from operation labels, gate
probabilities or spike traces using the proxy tables defined below.

The shared output structure allows final-state, trace, gate and memory-access
metrics to be computed consistently whenever the corresponding quantities are
exposed by an architecture. Every model is asked to solve the same symbolic
execution problem under the applicable task-family protocol. Models without
explicit operation gates leave gate-path metrics unreported rather than being
assigned an artificial gate interpretation.

\paragraph{\dcpu-\gru-\ntm.}
The primary model combines instruction embeddings, a recurrent controller, differentiable memory, operation gates and masked register writes. Its transition can be summarized as
\begin{align}
    q_t &= \mathrm{Read}_{\theta}(M_{t-1},h_{t-1}),\\
    h_t &= \mathrm{GRU}_{\theta}([\phi_t;q_t],h_{t-1}),\\
    \mathrm{aux}_t &= \tanh(W_{\mathrm{aux}}h_t+b_{\mathrm{aux}}),\\
    \pi_t &= \mathrm{Gate}_{\theta}\!\left([r_{t-1,i_t};r_{t-1,j_t};\mathrm{aux}_t;\onehot(o_t)]\right),\\
    r_t &= \mathrm{WriteReg}(r_{t-1},\pi_t,h_t,x_t),\\
    M_t &= \mathrm{WriteMem}_{\theta}(M_{t-1},h_t).
\end{align}
The memory write in the primary \ntm{} path is therefore controller-state driven, since it does not directly consume the post-\alu{} register value.

\paragraph{No-gate-loss ablation.}
The no-gate-loss ablation keeps the same architecture but sets $\lambda_G=0$. It tests whether the explicit symbolic path matters. If final loss stays low but gate accuracy and calibration collapse, the model has learned a less interpretable latent execution path.

\paragraph{No-memory ablation.}
The no-memory model sets $q_t=0$ and $M_t=M_{t-1}$ for all $t$. Its transition is
\begin{equation}
    h_t = \mathrm{GRU}_{\theta}([\phi_t;0],h_{t-1}).
\end{equation}
This ablation removes the explicit retrieved-value channel. On visible, local register tasks the recurrent state and instruction embedding can still carry enough context, but on delayed memory tasks the controller hidden state must now represent operation context, address history and retrieved value content simultaneously. Degradation in this setting therefore identifies loss of addressable value retrieval, not a generic capacity difference.

\paragraph{ValueMemory variant.}
The ValueMemory variant exposes an execution-coupled value lane. Read and write addresses are linear projections of controller state, and the read value can blend directly into \alu{} operands and writeback:
\begin{equation}
    \alpha_t=\softmax(W_r h_{t-1}),\qquad
    \nu_t=\sum_i \alpha_{t,i}V_{t-1,i}.
\end{equation}
Operand blending is
\begin{equation}
    [\lambda_u,\lambda_v]=\sigma(W_{\lambda}[h_t;\nu_t]),
    \qquad
    u_t^{\mathrm{eff}}=u_t+\lambda_u(\nu_t-u_t),
    \qquad
    v_t^{\mathrm{eff}}=v_t+\lambda_v(\nu_t-v_t).
\end{equation}
After \alu{} execution, writeback blending and memory update are
\begin{equation}
    \hat{y}_t \leftarrow \clip(\hat{y}_t+\lambda_y(\nu_t-\hat{y}_t),0,1),
    \qquad
    V_t=V_{t-1}(1-\omega_t)+\omega_t\hat{y}_t,
\end{equation}
where $\omega_t=\softmax(W_w h_t)$ and $\lambda_y=\sigma(W_y[h_t;\nu_t])$. This variant is motivated by hidden-opcode memory-pressure tasks, where delayed state can help execution.
When quantization simulation is enabled for ValueMemory variants, the quantizer is applied before the value-memory writeback blend. The subsequent blend is clamped but not re-quantized. ValueMemory-\qat{} results should therefore be read as quantized-\alu{}/value-memory robustness evaluations rather than strict final-destination-write quantization.

\paragraph{Transformer-\ntm.}
The transformer controller replaces the recurrent update with causal self-attention, producing a context vector that is then fused with the \ntm{} read and the previous controller state through a shared projection:
\begin{align}
    H &= \mathrm{Transformer}_{\theta}(\Phi + P),\\
    q_t &= \mathrm{Read}_{\theta}(M_{t-1},h_{t-1}),\\
    h_t &= \tanh\!\left(W_{\mathrm{step}}[H_t;\,q_t;\,h_{t-1}]\right),
\end{align}
where $\Phi$ is the sequence of instruction/state embeddings and $P$ positional encoding. The attention output $H_t$ is therefore an intermediate context, not the controller state used for gating and \ntm{} writes; $h_t$ plays that role, matching how the \gru-\ntm{} primary model separates a named mechanism from the shared \ntm{}-coupled recurrence. This construction can model long dependencies through attention but is less hardware-friendly than the recurrent model because attention cost scales with sequence length.

\paragraph{Transformer scratchpad.}
The scratchpad baseline receives the same instruction and initial-register information but predicts the trace without explicit operation gates:
\begin{equation}
    \hat{r}_{1:T}=\mathrm{Decoder}_{\theta}(\mathrm{Transformer}_{\theta}(X_{\mathrm{instr}},r_0)).
\end{equation}
It is included as an opaque trace-prediction baseline. It can report register errors, but because it has no gate trace it cannot support operation-path metrics.

\paragraph{TCN-\ntm.}
The temporal-convolution controller uses dilated causal convolutions to produce a context vector, again fused with the \ntm{} read and previous controller state through the shared projection:
\begin{align}
    c_t^{(\ell+1)} &=
    \sigma\left(
    \sum_{j=0}^{K-1}
    W_{\ell,j}c_{t-d_{\ell}j}^{(\ell)}
    +b_{\ell}
    \right),\\
    q_t &= \mathrm{Read}_{\theta}(M_{t-1},h_{t-1}),\\
    h_t &= \tanh\!\left(W_{\mathrm{step}}[c_t;\,q_t;\,h_{t-1}]\right),
\end{align}
where $c_t$ denotes the final-layer TCN output at step $t$. The dilation schedule increases receptive field while keeping the model more parallel than a \gru{} and more local than a transformer; as with Transformer-\ntm{}, $c_t$ is an intermediate context and $h_t$ is the controller state used for gating and \ntm{} writes.

\paragraph{Graph-\dcpu.}
Despite its name, the implemented \texttt{GraphDCPUBase} controller is not a message-passing graph neural network: it has no adjacency structure, no edge features and no neighbourhood aggregation. At each step it embeds a fixed set of six feature blocks: (i) the instruction, (ii) the mean-pooled register file, (iii) the $\mathrm{src}_a$ addressed-register embedding, (iv) the $\mathrm{src}_b$ addressed-register embedding, (v) the $\mathrm{dst}$ addressed-register embedding, and (vi) the pooled \ntm{} read vector. In addition, it fuses them through one shared two-layer \texttt{ReLU} MLP:
\begin{equation}
    n_t = \tanh\!\left(
    \mathrm{MLP}_{\theta}\!\left(
    \left[\phi_t;\ \bar{v}_t;\ v_{i_t}^t;\ v_{j_t}^t;\ v_{d_t}^t;\ q_t\right]
    \right)
    \right),
\end{equation}
where $v_r^t=\mathrm{ReLU}(W_{\mathrm{reg}}r_{t-1,r}+e_r)$ is a per-register embedding with a learned register-identity term $e_r$, $\bar{v}_t$ is its mean over registers, and $q_t$ is the \ntm{} read vector. The non-temporal variant uses $h_t=n_t$ directly. This fixed six-block fusion, not a relational graph over nodes and edges, is what the architecture name refers to in the current implementation.

\paragraph{TemporalGraph-\dcpu.}
TemporalGraph-\dcpu{} adds a single \texttt{GRUCell} across timesteps on top of the same fused feature vector:
\begin{equation}
    h_t=\mathrm{GRUCell}_{\theta}(n_t,h_{t-1}).
\end{equation}
It is included as a comparison architecture controller variant to test whether an explicit register/operand feature fusion followed by lightweight recurrence is a useful alternative to the primary \gru{} controller's direct instruction/memory input, not as a genuine graph-structured message-passing model.

\paragraph{SSM-\dcpu.}
The state-space variant replaces recurrent gating with a lightweight diagonal update, whose output is likewise fused with the \ntm{} read and previous controller state:
\begin{align}
    s_t&=\delta\odot s_{t-1}+(1-\delta)\odot \tanh(W_{\mathrm{in}}\phi_t),\\
    c_t&=\tanh(W_{\mathrm{out}}\mathrm{LayerNorm}(s_t)),\\
    q_t &= \mathrm{Read}_{\theta}(M_{t-1},h_{t-1}),\\
    h_t &= \tanh\!\left(W_{\mathrm{step}}[c_t;\,q_t;\,h_{t-1}]\right),
\end{align}
where $\delta=\sigma(\eta)$ is a learned per-channel decay and $c_t$ is the diagonal-update output before fusion. It is included to test efficient long-sequence control as a future alternative to attention.

\subsubsection{Spiking controller derivation}

The spiking models keep arithmetic and register writes non-spiking and use spikes only for controller memory. This is a crucial design decision. The continuous \alu{} bank is already the symbolic execution surface, therefore forcing the arithmetic itself to spike would make the model harder to train and less auditable. The spiking controller receives a direct-current encoding of the instruction and memory/value read:
\begin{equation}
    d_t=W_I[\phi_t;q_t]+b_I,
    \qquad k=1,\ldots,K,
\end{equation}
where $K$ is the number of micro-steps per program step. For LIF neurons,
\begin{align}
    c_{t,k+1} &= \alpha_c c_{t,k}+d_t+W_{\mathrm{rec}}z_{t,k},\\
    u_{t,k+1}^{\mathrm{raw}} &= \alpha_u u_{t,k}+c_{t,k+1},\\
    z_{t,k+1} &= \mathrm{surrogate\_spike}(u_{t,k+1}^{\mathrm{raw}}-\theta),\\
    u_{t,k+1} &= u_{t,k+1}^{\mathrm{raw}}(1-z_{t,k+1}).
\end{align}
For adaptive LIF neurons,
\begin{align}
    \theta_{t,k} &= \theta_0+\beta a_{t,k},\\
    a_{t,k+1} &= \gamma a_{t,k}+z_{t,k+1}.
\end{align}
The spike train is summarized as
\begin{equation}
    \bar{z}_t = \frac{1}{K}\sum_{k=1}^{K}z_{t,k},
\end{equation}
and passed to the same symbolic readout heads used by the non-spiking controller. The spike operator emits a hard threshold in the forward pass and uses a fast-sigmoid surrogate gradient in the backward pass:
\begin{equation}
    z=\mathrm{surrogate\_spike}(u-\theta).
\end{equation}
The model returns the spike train averaged over the $K$ micro-steps of each program step, rather than raw micro-step event tensors. The training metric therefore reports an averaged spike rate and a simple spike-energy proxy:
\begin{equation}
    \bar{z}_{n,t,i}=\frac{1}{K}\sum_{k=1}^{K}z_{n,t,k,i},
    \qquad
    \rho_{\mathrm{spike}} =
    \frac{1}{N_{\mathrm{active}}H}
    \sum_{n,t,i} m_{n,t} \bar{z}_{n,t,i},
    \qquad
    E_{\mathrm{spike,proxy}}=\rho_{\mathrm{spike}}H,
\end{equation}
where $H$ is the spiking hidden size. Reported spike counts are sums of averaged spike traces over active program steps. These are not raw micro-step event counts. The spike-energy proxy is not a hardware event-energy model: it does not include synaptic fan-out, routing, memory access, membrane-update cost, technology node, leakage or measured event energy. Our work reports these as proxy quantities only, and future neuromorphic deployment must replace them with measured event energy.

\subsubsection{Energy and delay proxies}
\label{sec:methods_energy_proxies}

Energy and delay values in this manuscript are dimensionless simulation proxies, not physical measurements. For a program $x_{1:T}$, proxy energy is
\begin{equation}
    E(x_{1:T})=\sum_{t=1}^{T} c(o_t) + c_{\mathrm{mem}}(t),
\end{equation}
where $c(o_t)$ is the operation-dependent cost. In sequence-energy scoring, fixed read, write and memory-attention costs are added for each active step. In the optional training loss, only the gate-probability-weighted operation table is used. The spike-rate proxy $\rho_{\mathrm{spike}}$ and spike-energy proxy $E_{\mathrm{spike,proxy}}$ introduced below are reported as separate spiking-controller diagnostics. The implemented sequence-energy scoring used for benchmark and energy-choice scoring never adds a spike-derived term to $E(x_{1:T})$. The delay equation is expressed as:
\begin{equation}
    \Delta(x_{1:T})=\sum_{t=1}^{T} d(o_t).
\end{equation}
The sequence delay is the sum of operation delays only. The fixed sequence-energy memory terms are architecture-agnostic proxy charges and are not conditioned on whether a particular model variant disables memory or whether a baseline is scratchpad-style. The purpose is to test whether the learned executor can expose meaningful cost traces and choose lower-cost equivalent symbolic paths. The manuscript therefore avoids claiming measured power, physical energy or physical delay. Where a voltage parameter appears in the simulator, it acts only as a $V^2$ multiplier on synthetic operation costs, therefore no capacitance, frequency, switching-probability, leakage, thermal or circuit model is implemented.

\subsubsection{Equivalent-program energy-choice tasks}
\label{sec:methods_energy_choice}

The energy-choice dataset extends each sample with canonical and alternative valid traces:
\begin{equation}
    (x^{\mathrm{can}},x^{\mathrm{alt}},x^{\min},r_0,r_T,
    r_{1:T}^{\mathrm{can}}, r_{1:T'}^{\mathrm{alt}}, r_{1:T''}^{\min},
    E^{\mathrm{can}},E^{\mathrm{alt}},E^{\min}).
\end{equation}
The final state is shared:
\begin{equation}
    \mathcal{T}(r_0,x^{\mathrm{can}}) =
    \mathcal{T}(r_0,x^{\mathrm{alt}}) =
    \mathcal{T}(r_0,x^{\min}) = r_T.
\end{equation}
A model trained with target policy $\pi_{\mathrm{target}}=\mathrm{min\_energy}$ receives $x^{\min}$ and is scored by final correctness, canonical trace distance, alternative trace distance, minimum-energy trace distance and any-valid trace distance:
\begin{equation}
    d_{\mathrm{any}}(\hat{r}_{1:T}) =
    \min\left\{
    d(\hat{r}_{1:T},r_{1:T}^{\mathrm{can}}),
    d(\hat{r}_{1:T},r_{1:T'}^{\mathrm{alt}}),
    d(\hat{r}_{1:T},r_{1:T''}^{\min})
    \right\}.
\end{equation}
The energy gaps are
\begin{equation}
    \Delta E_{\mathrm{can}} = E^{\mathrm{exec}}-E^{\mathrm{can}},
    \qquad
    \Delta E_{\min} = E^{\mathrm{exec}}-E^{\min}.
\end{equation}
Here $E^{\mathrm{exec}}$ is computed post hoc from the executed or selected symbolic path, and not a direct model output. This turns energy from a passive reporting channel into an execution constraint.

\section{Results}

\subsection{Register-machine execution interface}
\label{sec:dataset_protocols}
\label{sec:experimental_programme}

\begin{figure}[tbp]
    \centering
    \includegraphics[width=\textwidth,height=0.62\textheight,keepaspectratio]{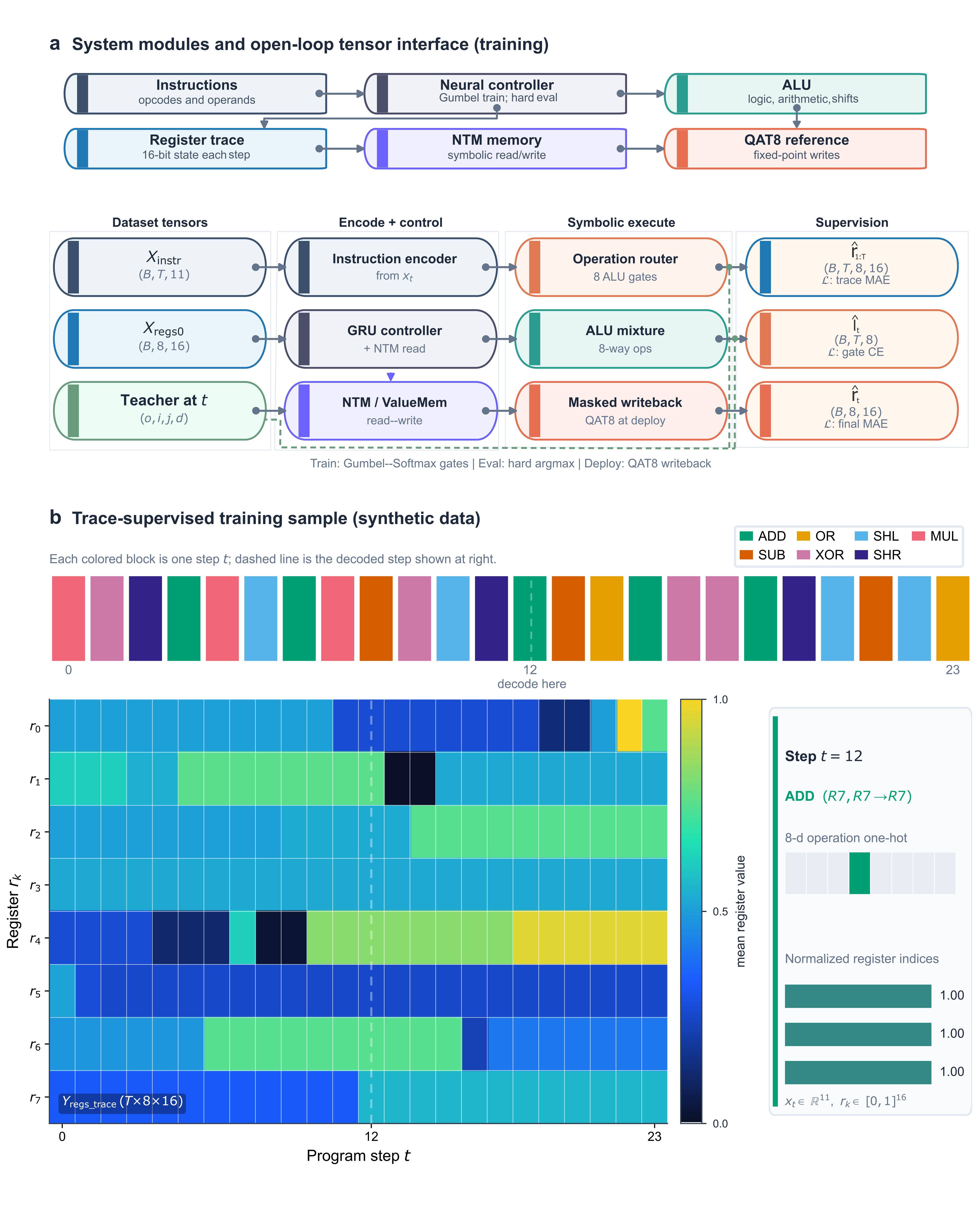}
    \caption[Training interface for trace-supervised symbolic execution.]
{\textbf{Training interface for trace-supervised symbolic execution.}
\textbf{(a)} Open-loop training interface with the module schematic above and
batched tensor unrolling below. Encoded instructions drive a GRU controller and
a gated fixed-operation ALU bank. Masked writeback produces a register trace
supervised against the reference interpreter, with an optional memory pathway
and optional QAT8 writeback.
\textbf{(b)} One reference-executed program showing the opcode strip, the
register-trajectory heatmap averaged over the $W{=}16$ scalar lanes of each
register, and the 11-dimensional instruction encoding at a selected step.}
    \label{fig:architecture}
\end{figure}

The \sncpu{} is constructed so that the quantities exposed by the model are the same quantities used to audit it. These quantities include the input
instruction, the selected operation gate, the source and destination
registers, the masked writeback, the complete register trace, optional memory
access and the low-precision replay reference. This differs from an unconstrained sequence model \(f_{\theta}(x_{1:T},r_0)\mapsto r_T\), which can match a final state without exposing whether the intermediate computation followed the intended program. \Cref{fig:architecture} therefore defines the execution interface used throughout the investigation.
The architecture is not only a trainable mapping, but the measurement surface on which
trace faithfulness, gate agreement, replay exactness and failure modes are evaluated.

The schematic should be read with the following scope conventions. Solid arrows indicate
runtime information flow, whereas dashed green paths denote reference targets (register trace, operation gates and final state) which are used for supervision and evaluation, and not for
extra runtime inputs. The memory block denotes \emph{alternative} optional variants such as the Neural Turing Machine (NTM, slot-addressed) or ValueMemory (value-addressed), both
sharing the same execution interface. At most one memory variant is active in a trained
instance, and the principal executor uses the \gru-\ntm{} configuration. \textbf{QAT8}
denotes eight-bit quantization-simulated writeback at the register boundary, where selected
register writes are projected onto an eight-bit fixed-point grid and evaluated against a
matched fixed-point replay, as shown in
\cref{fig:main_benchmark}d and Supplementary
Fig.~\ref{fig:device_ref}. This procedure does not constitute full
integer-only inference. The schematic label ``16-bit state each step'' refers to the principal
$16$-wide continuous register profile, with $W{=}16$ scalar lanes per register
in $[0,1]$. It does not refer to bit-accurate 16-bit integer arithmetic. State
variables, instruction encoding, gate notation and metrics are defined in
\cref{sec:methods} and Table~\ref{tab:notation}.

At each program step, the executor receives an encoded instruction \(x_t\), reads two source registers, predicts an operation distribution, computes a candidate update through a named ALU bank and writes only the destination register. The resulting transition is
\begin{equation}
    \hat{r}_t = F_{\theta}(\hat{r}_{t-1},x_t),
\end{equation}
where \(\hat{r}_t\) is the predicted register file. In practice, \(F_{\theta}\) is factorized into an instruction encoder, optional memory read, recurrent controller state, interpretable operation-gate distribution and masked destination-register write. This factorization is the core architectural contribution where final-state imitation, trace reconstruction and operation-path audit become separable measurements rather than one entangled endpoint score.

The operation choice is represented by a probability vector over named CPU-like operations,
\begin{equation}
    \hat{y}_t =
    \sum_{k\in\opset}\pi_{t,k}
    A_k(\hat{r}_{t-1,i_t},\hat{r}_{t-1,j_t}),
\end{equation}
where \(i_t\) and \(j_t\) are zero-based source-register indices, \(A_k\) is the implemented continuous operation bank and \(\pi_{t,k}\) is the learned gate probability. The operation bank is not a bit-accurate integer ISA. It is a differentiable, bounded execution substrate whose operation identities remain visible. Each transition therefore has a symbolic explanation, where which operation was selected, which operands were read and which register was written.

The low-precision behaviour is also made part of the interface rather than treated as a post-hoc compression step. When QAT8 is enabled, the selected writeback value is projected onto the same eight-bit simulated quantization grid used by the matched fixed-point replay,
\begin{equation}
    Q_B(x)=\frac{\clip(\round(x(2^B-1)),0,2^B-1)}{2^B-1}.
\end{equation}
Here \(B\) denotes the quantization bit depth only. The matched replay is an
independent fixed-point reimplementation of the reference interpreter. It is
executed on the same programs at bit depth \(B\) and never consults the model's
outputs. Agreement with this replay is therefore an external verification of
low-precision execution semantics rather than a self-consistency measure. The
implementation does not claim full integer-only inference or measured hardware
energy.

The same interface supports variants without changing the audit surface. The primary \gru-\ntm{} executor uses slot-addressed NTM memory. The ValueMemory replaces the latter with value-addressed memory for hidden-opcode pressure tests, and spiking \alif{} controllers replace the recurrent controller state with event-driven temporal state while leaving the symbolic ALU and writeback path non-spiking. The architecture ablations and scratchpad baselines consume the same instruction and register layout, which makes their failures interpretable. Therefore, a baseline can match endpoints yet fail to expose the named operation path.

Finally, equivalent-program tasks turn cost from a passive label into a symbolic choice. The benchmark constructs alternative programs that reach the same final state but have different dimensionless proxy costs, so the executor can be scored against a minimum-energy symbolic path,
\begin{equation}
    x^\star = \argmin_{x\in\mathcal{X}(r_0,r_T)} E(x)
    \quad\mathrm{subject\ to}\quad
    \mathcal{T}(r_0,x)=r_T.
\end{equation}
Thus canonical-path agreement is not always the appropriate metric. A lower-cost program can be correct when it follows a different valid trace. These distinctions organize the Results. The main evidence is trace-verified visible-opcode execution with matched replay, where the next evidence identifies mechanisms and limits through ablations, perturbations, cost choice and hidden-opcode memory pressure. The final sections extend the same interface to rollout diagnostics, spiking controllers, closed-loop control and RV32I base-integer semantics. \Cref{fig:main_benchmark} evaluates whether the architecture preserves long-horizon execution under hard gates and QAT8 writeback. \Cref{fig:stress_cost_limits} then changes the evaluation of the perturbation input condition, equivalent program cost, architecture class or hidden-opcode memory pressure to identify which component of the execution stack is challenged. Finally, \cref{fig:selected_stack,fig:snn_signals,fig:closed_loop_control} inspect the same execution interface at rollout level, where selected non-spiking executions, spiking-controller dynamics and closed-loop candidate-constrained instruction selection. The detailed tensor shapes, loss definitions, generation protocols and controller objectives are given in Methods.

\subsection{Length-stable \qat{} execution on visible-opcode programs}
\label{sec:main_benchmark}

\begin{figure}[!tbp]
    \centering
    \includegraphics[width=\textwidth,height=0.6\textheight,keepaspectratio]{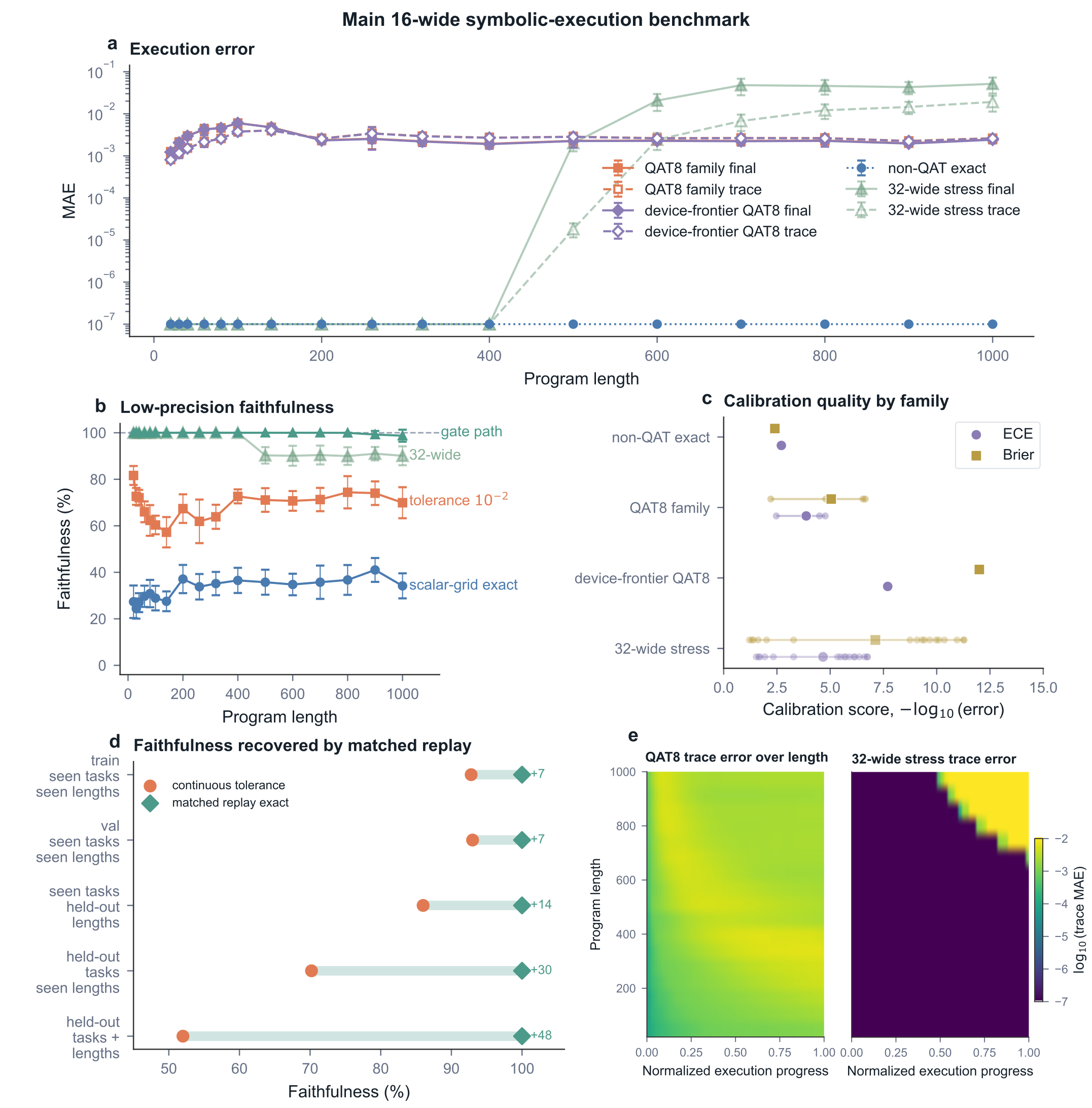}
    \caption[Symbolic-execution benchmark envelope on the 16-wide register-machine profile.]
    {\textbf{Symbolic-execution benchmark envelope on the 16-wide register-machine profile.}
    Panels show continuous-reference hard-execution error, QAT8 faithfulness, gate
    calibration, matched fixed-point replay recovery and within-program trace-error
    localization. The 32-wide extrapolation profile is overlaid in panels~(a--c);
    detailed metric, replay, uncertainty and proxy-cost conventions are given in the
    surrounding text.}
    \label{fig:main_benchmark}
\end{figure}

\Cref{fig:main_benchmark} establishes the main execution result before relaxing the assumptions of the benchmark. The experiment follows the principal 16-wide visible-opcode register profile ($W{=}16$) from short programs to $L{=}1000$ under deterministic hard execution, and separates five quantities that would otherwise be a conflated and continuous-reference register error, symbolic-path faithfulness, gate calibration, matched fixed-point replay and the location of error within a program. This separation matters because a small continuous drift, a wrong opcode path and a reference-semantics mismatch are different scientific outcomes. A low-precision executor should therefore not be judged only by its terminal state.

All trained families in \cref{fig:main_benchmark} are evaluated with hard argmax gates, evaluation temperature $\tau{=}0.5$ for the stored gate probabilities, continuous-reference scoring unless replay is stated explicitly, and 8 benchmark batches of 64 synthetic programs per length (512 programs per length point, evaluation seed~321). The length grid combines a core sweep, $L\in\{20,30,40,60,80,100,140,200,260,320,400,500\}$, with a long-horizon extension, $L\in\{600,700,800,900,1000\}$. Curves aggregate batch means. Error bars combine within-batch standard deviation and across-instance spread when a plotted family contains multiple trained instances. Exact metric definitions (MAE, faithfulness, \ECE{}, \Brier{}, proxy energy and proxy delay) are given in Methods.

The comparison has three primary families. The \textbf{non-QAT exact} \gru-\ntm{} executor omits quantization simulation and defines the continuous-reference ceiling. Hard final and trace errors are effectively zero and are drawn at the $10^{-7}$ display floor on log axes. The \textbf{QAT8 family} pools four eight-bit quantization-simulated writeback trained instances with the same architecture but different seed or warmup schedule. The warmup condition trains briefly in full precision before enabling 8-bit writeback simulation. The \textbf{long-program ValueMemory QAT8} profile is a \qat{} \gru-ValueMemory instance trained under a longer-program, memory-augmented regime to test deployment-facing extrapolation. This condition denotes a training and data regime rather than a hardware result. No device-specific timing, bandwidth or power model is applied.

A wider $W{=}32$ profile is overlaid in \cref{fig:main_benchmark}a to c only as a secondary extrapolation envelope. This trained instance saw programs no longer than 40 steps during training and was then evaluated far beyond that horizon. Its long-horizon limit is therefore not interpreted as a state-width capacity limit. Retraining the same architecture at a longer horizon removes the limit (Supplementary Fig.~\ref{fig:t400_mechanism} and Table~\ref{tab:t400_mechanism}). The complete 32-wide length-resolved analysis, including faithfulness, calibration and per-operation structure, is kept in Supplementary Figs.~\ref{fig:supp_32wide_longhorizon} and~\ref{fig:width_comparison} as our main analysis remains centred on the robust 16-wide profile.

The figure separates complementary failure modes, including length stability under hard execution, preservation of the named operation path, calibration of the soft gate probabilities, scoring against matched low-precision semantics, and the location of residual error within a program (\cref{fig:main_benchmark}a to e). The central result is that \qat{} decouples discrete logical correctness from continuous numerical matching. Operation paths stay effectively exact through $L{=}1000$ while continuous-reference register MAE remains in the $10^{-3}$ band.

We first assess performance and logical reliability. \Cref{fig:main_benchmark}a plots \emph{mean absolute error} (\MAE{}) between hard-executed register files and the \emph{continuous} reference on a log scale versus program length.
\textbf{Final MAE} compares terminal states $\hat{r}_T$ vs.\ $r_T^{\mathrm{ref}}$. \textbf{Trace MAE} averages over active timesteps and register lanes (masked $\ell_1$ norm as in training, \cref{sec:methods}). The solid lines are final error and dashed lines are trace error.
\textbf{non-QAT exact} hard execution is numerically exact against the continuous reference (final MAE effectively $0$ at all reported lengths, including the longest tested $L{=}1000$). Because the $y$-axis is logarithmic, these zero values are drawn at a $10^{-7}$ display floor, so the flat bottom line marks exactness rather than a small nonzero error.
The \textbf{QAT8 family} and \textbf{long-program ValueMemory QAT8} curves lie orders of magnitude above that floor yet remain bounded across the full horizon. Primary \qat{} hard final MAE is $2.01\times10^{-3}\pm3.8\times10^{-4}$ at $L{=}30$, peaks at $4.35\times10^{-3}\pm1.8\times10^{-3}$ at $L{=}100$, remains elevated at $3.83\times10^{-3}\pm7.8\times10^{-4}$ at $L{=}200$, and eases toward the range from $2\times10^{-3}$ to $2.5\times10^{-3}$ by longer horizons ($2.28\times10^{-3}\pm6.7\times10^{-4}$ at $L{=}500$, $2.46\times10^{-3}\pm4.4\times10^{-4}$ at $L{=}1000$, Table~\ref{tab:main_length}). Pooling the four 16-wide \qat{} trained instances over all benchmark lengths and batches, the aggregated family final \MAE{} is $2.87\times10^{-3}$ ($95\%$ bootstrap confidence interval $[2.76,2.99]\times10^{-3}$, $n{=}544$ per-length, per-batch samples).
Trace MAE follows a similar pattern ($1.14\times10^{-3}$ at $L{=}30$, $2.83\times10^{-3}$ at $L{=}100$, $4.54\times10^{-3}$ at $L{=}200$, $2.86\times10^{-3}$ at $L{=}500$, and $2.64\times10^{-3}$ at $L{=}1000$).
The long-program ValueMemory \qat{} profile tracks the primary within ${\sim}1\%$ at representative lengths through $L{=}1000$.
For inspectable symbolic execution, the critical threshold is not bit-exact floating-point agreement with a continuous reference but \emph{opcode routing and trace inspectability} under quantization-simulated writeback.
\Cref{fig:main_benchmark}b shows that although \qat{} exhibits higher continuous-reference MAE than non-\qat{} exact, this numerical noise does not corrupt logical routing.
\textbf{Gate path} (\textbf{gate agreement}, $A_G$ in \cref{sec:methods}) is the fraction of active steps where the hard-selected operation $\hat{o}_t=\arg\max_k \ell_{t,k}$ (the operation with the largest gate logit, \cref{eq:interface-hard-gate}) equals the reference opcode (the operation the reference interpreter executed at that step). For the aggregated family it remains \textbf{100.0\%} at every length through $L{=}700$ and stays at or above $\mathbf{98.7\%}$ through $L{=}1000$, with the primary and long-program ValueMemory trained instances individually holding $100.0\%$ at every length to $L{=}1000$ (the small aggregate dip at $L\geq800$ comes from two seed replicates, $n{=}32$ batch records per length across four \qat{} trained instances).

This measurement isolates the main execution claim. The neural controller selects the correct named operation at essentially every step even at long extrapolation lengths, when register values drift in continuous space.
\textbf{Tolerance $10^{-2}$} measures a softer register criterion (the fraction of programs with $\|\hat{r}_T-r_T^{\mathrm{ref}}\|_\infty\leq10^{-2}$ in normalized $[0,1]$ space) and stays in the range from $\mathbf{56\%}$ to $\mathbf{74\%}$ (dropping to $\mathbf{56.3\%}$ at $L{=}200$, $\mathbf{71.1\%}$ at $L{=}500$ and $\mathbf{69.9\%}$ at $L{=}1000$).
That tolerance dip co-varies with the MAE peak at lengths from $L{\approx}100$ to $L{=}200$ while gate agreement remains exact, explicitly decoupling continuous numerical looseness from opcode faithfulness.
\textbf{Scalar-grid exact} is stricter still. Each scalar lane must match the reference after projecting both sides onto the profile-dependent fixed-point grid (not packed machine-word equality). It stays in the range from $\mathbf{23\%}$ to $\mathbf{41\%}$ against the \emph{continuous} reference (the strictest lower bound before matched replay scoring).
Continuous-reference MAE therefore measures reference-semantics mismatch under 8-bit writeback, not gate collapse. Matched fixed-point replay yields 100\% scalar-grid faithfulness (\cref{fig:main_benchmark}d, Supplementary Fig.~\ref{fig:device_ref}, Table~\ref{tab:device_ref}).

Calibration provides a complementary view of gate trustworthiness. Deterministic hard execution is assessed as an opcode decision problem through gate agreement, whereas probability calibration is meaningful for the soft gate distribution before argmax selection.
\Cref{fig:main_benchmark}c therefore reports \textbf{soft-mode} diagnostics. Expected calibration error (\ECE{}) and Brier score (\Brier{}) between predicted gate probabilities and reference opcode labels \citep{Guo2017Calibration,Brier1950} are averaged over all length and batch records in each family and displayed as $-\log_{10}(\mathrm{error})$ (higher is better). These scores use the model's raw predicted gate probabilities directly. No post-hoc temperature scaling or Platt recalibration is applied before computing \ECE{} or \Brier{}, so they reflect the executor's native confidence rather than a recalibrated proxy.
These metrics expose residual probability mass when gates are evaluated in soft form during benchmarking, not a contradiction with exact hard routing.
Evaluated across the full range from $L{=}20$ to $L{=}1000$ (deep extrapolation included), soft errors are larger and more variable than within the core sweep. The \textbf{non-QAT exact} mean soft \ECE{} is $2.0\times10^{-3}$ ($-\log_{10}\ECE\approx2.7$) and its \Brier{} score is $4.0\times10^{-3}$ ($-\log_{10}\Brier\approx2.4$). The \textbf{QAT8 family} mean soft \ECE{} is $9.0\times10^{-4}$ ($-\log_{10}\ECE\approx3.0$) and its \Brier{} score is $1.5\times10^{-3}$ ($-\log_{10}\Brier\approx2.8$).
These soft errors are dominated by the long extrapolation lengths and, within the \qat{} family, by the warmup instance (soft \ECE{} $3.4\times10^{-3}$). The primary \qat{} instance alone remains an order of magnitude better (soft \ECE{} $3.2\times10^{-5}$, $-\log_{10}\ECE\approx4.5$).
These are \emph{soft}-mode residuals only. Hard argmax routing is summarized separately by gate agreement, so the soft drift reflects less-peaked gate probability mass at long horizons rather than necessarily erroneous opcode routing.
Visual scatter in \cref{fig:main_benchmark}c largely reflects within-family spread across the four \qat{} trained instances (the warmup instance being the least soft-calibrated) and the compression of errors spanning from $10^{-3}$ to $10^{-14}$ on a $-\log_{10}$ axis, not competing statistical verdicts.
\textbf{Long-program ValueMemory QAT8} shows the best soft calibration ($-\log_{10}\ECE\approx7.7$, $-\log_{10}\Brier\approx13.7$) despite MAE similar to primary \qat{}, indicating that calibration is sensitive to the training regime and initialization rather than to continuous-reference MAE alone.
Both \ECE{} and \Brier{} are included to avoid relying on a single calibration metric. Extended calibration curves appear in the supplementary execution diagnostics.

Proxy cost is not the axis that separates these families. A natural concern is whether proxy cost confounds the accuracy interpretation in \cref{fig:main_benchmark}a to c. It does not, because under the shared proxy formula the standard length benchmark assigns nearly identical per-step operation costs to the compared families.
The comparison uses dimensionless simulation proxies for operation-level energy and delay (\cref{sec:methods}), \emph{not} measured silicon power or cycle-accurate timing.
Per-family proxy totals are divided by program length $L$, then normalized to the mean \qat{} primary per-step values.
Across \textbf{non-QAT exact}, \textbf{QAT8 family}, and \textbf{long-program ValueMemory QAT8}, per-step energy and delay cluster within the range from $\mathbf{0.995}$ to $\mathbf{1.005}$ of the \qat{} reference, and are width-independent by construction because all families and profiles share the same proxy ISA formula. Trained-instance or width choice does not artificially inflate or deflate cost in the simulator.
Per-step normalization therefore rules out a proxy-cost artefact. The primary \qat{} proxy energy grows approximately linearly with length ($\approx8059$ at $L{=}200$, $\approx20150$ at $L{=}500$ and $\approx40297$ at $L{=}1000$), but the compared families are not separated by this axis (Table~\ref{tab:long_horizon} and supplementary proxy-cost diagnostics).
Proxy cost becomes scientifically relevant only when the program itself is allowed to change. Equivalent-program energy-choice tasks are therefore analysed later (\cref{fig:stress_cost_limits}b and Table~\ref{tab:energy_choice}), where lower-cost valid traces are competing execution paths rather than post-hoc labels attached to one path.

The analysis then moves from continuous scoring to replay semantics. \Cref{fig:main_benchmark} is an open-loop length benchmark scored first against the continuous reference transition. In that reference frame, the \qat{} family trades floating-point precision for length-stable symbolic execution. Final-state drift stays in the range from ${\sim}2\times10^{-3}$ to $2.5\times10^{-3}$ at long horizons, gate paths remain effectively exact, and the drift is an order of magnitude below the $10^{-2}$ final-state tolerance used for downstream closed-loop success (\cref{sec:closed_loop_control}).
The appropriate hardware-facing question is then whether this drift persists when the reference is changed to the same low-precision semantics as the model. The matched fixed-point replay in \cref{fig:main_benchmark}d and Supplementary Fig.~\ref{fig:device_ref} answers that question. On the task-split device-reference evaluations of the primary \qat{} executor (\cref{tab:device_ref}), final and trace MAE are zero and scalar-grid faithfulness is 100\% under the same eight-bit writeback semantics the executor was trained to satisfy.
The evidence remains simulation-level. It does not establish measured silicon energy,
cycle-accurate timing, full integer-only inference or a complete instruction-set emulator.
The supported claim is semantics-preserving quantization-simulated writeback with an
inspectable opcode trace.

\Cref{fig:main_benchmark} summarizes the symbolic-execution benchmark envelope on the
principal 16-wide register-machine profile (\cref{sec:main_benchmark}). Metric formulas
are defined in \cref{sec:methods_metrics}, and symbols are summarized in
Table~\ref{tab:notation}. The profile label denotes register \emph{width} in continuous
state variables, \emph{not} bit-accurate integer arithmetic. ``16-wide'' is the principal
$16$-state-variable profile and is unrelated to a 16-bit instruction-set architecture
(ISA).

A wider $32$-state-variable \emph{extrapolation} profile, shown in sage green and drawn
semi-transparently, is overlaid as a secondary envelope in \cref{fig:main_benchmark}a to c.
This profile comes from a trained instance exposed only to programs of at most 40
instructions during training, and is included to probe extrapolation far past its training
horizon. Its long-horizon limit is a training-horizon effect, not a state-width limit
(Supplementary Fig.~\ref{fig:t400_mechanism}). Its full length-resolved breakdown is
reported separately in Supplementary Figs.~\ref{fig:supp_32wide_longhorizon}
and~\ref{fig:width_comparison}.

The legend names three trained-instance families. \textbf{non-\qat{} exact}
denotes the non-quantized hard-execution continuous-reference ceiling.
\textbf{\qat{} family} denotes four eight-bit quantization-simulated writeback
instances that differ in seed or warmup. \textbf{long-program ValueMemory
\qat{}} denotes the extended value-memory long-program variant.

\Cref{fig:main_benchmark}a reports hard final and trace mean absolute error (\MAE{}) as
a function of program length, from $L{=}20$ to $L{=}1000$, where $L$ is the number of executed
instructions. Solid curves are final-state errors and dashed curves are trace errors,
both measured against the continuous-reference trajectory and plotted on a logarithmic
scale. Matched-replay recovery is summarized in \cref{fig:main_benchmark}d. Final
\MAE{} measures endpoint imitation, whereas trace \MAE{} measures whether the whole
trajectory is preserved. The non-\qat{} exact baseline is numerically exact at every
length, with final \MAE{} equal to $0$ and drawn at a $10^{-7}$ display floor only for
log-axis visibility. It establishes the continuous-reference ceiling against which the
\qat{} families are scored. The semi-transparent sage-green \emph{32-wide extrapolation}
curves, with final error solid and trace error dashed, trace the short-horizon-trained
extrapolation profile. It is numerically exact at the $10^{-7}$ floor through
$L{\approx}400$, then shows a sharp long-horizon divergence rising to between
${\sim}4\times10^{-2}$ and $5\times10^{-2}$ by $L{=}1000$, in contrast to the length-stable 16-wide
families. This overlaid $32$-state-variable profile is exact at short lengths and exhibits
a clear long-horizon extrapolation boundary. Its multi-seed length-resolved breakdown,
including faithfulness, calibration and per-operation structure, is detailed in
Supplementary Figs.~\ref{fig:supp_32wide_longhorizon} and~\ref{fig:width_comparison}.
Error bars combine benchmark-batch standard deviation and across-instance spread within
each family, not cross-seed uncertainty for single-instance families. Formal $95\%$
bootstrap confidence intervals and the significance tests quoted in the main text
(\cref{sec:methods}) are computed from the pooled per-length, per-batch samples.

\Cref{fig:main_benchmark}b reports faithfulness on the aggregated \qat{} family. The
\emph{gate path} is the fraction of active steps with the correct hard argmax opcode
relative to the reference interpreter. It is saturated through the core horizon and remains
near-saturated at the longest lengths. \emph{Tolerance $10^{-2}$} is the fraction of
programs whose final-state $\ell_\infty$ error is at most $10^{-2}$. \emph{Scalar-grid
exact} is agreement after projecting register components onto the fixed-point grid, and
should not be read as packed-word bit equality. All three quantities are measured against
the continuous reference interpreter. The semi-transparent sage-green \emph{32-wide
scalar-grid} curve overlays the same short-horizon-trained extrapolation profile. It is
exact, approximately $100\%$, through $L{\approx}400$, then declines to between
${\sim}88\%$ and $90\%$ at long horizons, consistent with Supplementary
Fig.~\ref{fig:width_comparison}c.

\Cref{fig:main_benchmark}c reports gate calibration over operation-gate predictions for
the three 16-wide families (non-\qat{} exact, \qat{}, and long-program ValueMemory
\qat{}) together with the overlaid 32-wide extrapolation profile. Soft-mode expected
calibration error (\ECE{}) and the \Brier{} score are distinct metrics, and both are
plotted as $-\log_{10}(\mathrm{error})$, so higher values are better. Hard argmax routing
is reported as gate agreement in \cref{fig:main_benchmark}b and is not interpreted as
probability calibration. Within each row, faint points and the connecting bar show the
score distribution, and the large marker shows its mean. For the three 16-wide rows, this
spread is taken \emph{across seeds/runs} at matched length aggregation. For the single-run
\emph{32-wide extrapolation} row, the spread is taken \emph{across program lengths}. Its
wide range therefore reflects calibration that is tight at short programs but degrades
steeply toward the long-horizon limit, rather than seed-to-seed variability.

\Cref{fig:main_benchmark}d reports faithfulness recovered by matched fixed-point replay
across five evaluation splits, combining seen versus held-out task families with seen
versus held-out program lengths as defined in \cref{sec:methods}. The panel compares
continuous-reference tolerance faithfulness, shown in orange, with exact matched-replay
faithfulness, shown as green diamonds, and annotations give the recovered percentage
points, from $+7$ to $+48$, with the largest recovery on held-out tasks \emph{and}
held-out lengths. Matched replay is an independent fixed-point reimplementation of the
reference instruction set executed at the model's 8-bit writeback precision. It is not the
model grading its own rounding, and it is an \emph{offline scoring/verification reference
computed after execution, not a runtime fallback}. The neural executor remains the object
being evaluated, and only its register writeback is quantization-simulated rather than
full integer-only inference. The full split-level breakdown is reported in Supplementary
Fig.~\ref{fig:device_ref} and Table~\ref{tab:device_ref}. \Cref{fig:main_benchmark}e localizes execution error within each program. It shows
per-step hard trace \MAE{} on a $\log_{10}$ scale, with a shared colour bar, as a heatmap
over normalized execution progress, step$/(L{-}1)$, on the horizontal axis and program
length on the vertical axis. The left heatmap shows the 16-wide \qat{} family, and the
right heatmap shows the 32-wide extrapolation profile. The 16-wide plateau is reached
within the first ${\sim}10\%$ of execution and is essentially length-independent. The
32-wide high-error boundary is diagonal because error onset occurs at a fixed absolute
step, ${\approx}494$, corresponding to normalized progress $494/L$. The full diagnostic
is shown in Supplementary Fig.~\ref{fig:trace_divergence}.

Per-step energy and delay proxy cost is omitted from the main benchmark panel because it
is width- and family-agnostic under the shared proxy formula. The cost-choice experiment
where proxy cost changes the executed path appears in \cref{fig:stress_cost_limits}b.
\begin{table*}[!t]
\centering
\caption{\textbf{Primary \qat{} length benchmark under hard deterministic execution and the continuous reference.}
Means $\pm$ benchmark-batch standard deviation over the benchmark batches at each length (4 batches for $L\leq200$ from the original benchmark bundle; 8 batches for $L\geq500$ from the long-horizon grid).
Full curves are shown in \cref{fig:main_benchmark}a,b.
At $L{=}500$, gate agreement remains 1.000 and tolerance faithfulness is 0.711 under the continuous reference.
At $L{=}1000$, gate agreement remains 1.000, tolerance faithfulness is 0.699 and final MAE is $2.46\times10^{-3}$.
Matched replay is exact, as shown in \cref{fig:main_benchmark}d and Supplementary Fig.~\ref{fig:device_ref}.
The long-horizon rows extend the same evaluation protocol to $L{=}500$ and $L{=}1000$.}
\label{tab:main_length}

\small
\setlength{\tabcolsep}{5pt}
\renewcommand{\arraystretch}{1.2}

\begin{tabular}{@{}cccccc@{}}
\toprule
Length &
\shortstack{Final\\MAE} &
\shortstack{Trace\\MAE} &
\shortstack{Tolerance\\faithfulness} &
\shortstack{Scalar-grid\\exact} &
\shortstack{Gate\\agreement} \\
\midrule

30 &
\shortstack{$2.01\times10^{-3}$\\$\pm\,3.8\times10^{-4}$} &
\shortstack{$1.14\times10^{-3}$\\$\pm\,1.5\times10^{-4}$} &
0.742 &
0.227 &
1.000 \\

100 &
\shortstack{$4.35\times10^{-3}$\\$\pm\,1.8\times10^{-3}$} &
\shortstack{$2.83\times10^{-3}$\\$\pm\,1.0\times10^{-3}$} &
0.621 &
0.305 &
1.000 \\

200 &
\shortstack{$3.83\times10^{-3}$\\$\pm\,7.8\times10^{-4}$} &
\shortstack{$4.54\times10^{-3}$\\$\pm\,9.5\times10^{-4}$} &
0.563 &
0.293 &
1.000 \\

500 &
\shortstack{$2.28\times10^{-3}$\\$\pm\,6.7\times10^{-4}$} &
\shortstack{$2.86\times10^{-3}$\\$\pm\,7.0\times10^{-4}$} &
0.711 &
0.357 &
1.000 \\

1000 &
\shortstack{$2.46\times10^{-3}$\\$\pm\,4.4\times10^{-4}$} &
\shortstack{$2.64\times10^{-3}$\\$\pm\,4.6\times10^{-4}$} &
0.699 &
0.342 &
1.000 \\

\bottomrule
\end{tabular}
\end{table*}

\subsection{Matched replay separates reference drift from execution error}

The continuous reference used in the length benchmark is the right ceiling for a
non-quantized executor, but it is not the only meaningful reference for a model
that is constrained to eight-bit writeback. In the implemented metrics,
``exact'' low-precision faithfulness is computed by projecting continuous
register-vector components onto the profile-dependent scalar grid. It is not
equality of packed machine-word bits.

The matched fixed-point replay provides the corresponding low-precision
reference. It is an independent reimplementation of the reference instruction
set, executed on the same programs at the model's 8-bit writeback precision. It
does not consult the model's outputs and is not a runtime fallback. The neural
executor remains the object being evaluated, while the replay defines the
correct eight-bit trajectory for that program.

Under this matched reference, every tested split has zero replay final MAE,
zero replay trace MAE, 100\% exact scalar-grid faithfulness and gate agreement
of 1.000. These saturated matched-replay quantities are stated in the text
rather than repeated in every row of Table~\ref{tab:device_ref}. The table
therefore reports the continuous-reference MAE, which is the quantity that
varies across the evaluation splits
(\cref{fig:main_benchmark}d and Supplementary
Fig.~\ref{fig:device_ref}).

The recovery is largest exactly where continuous scoring is most pessimistic.
Tolerance faithfulness that falls to $70\%$ on held-out tasks and $52\%$ on
held-out tasks \emph{and} lengths returns to $100\%$ under matched replay.
These changes correspond to gains of $+30$ and $+48$ percentage points
(\cref{fig:main_benchmark}d and Supplementary
Fig.~\ref{fig:device_ref}c). Final and trace MAE also collapse to the numerical
floor on every split, as shown in Supplementary
Fig.~\ref{fig:device_ref}b,d. The continuous-reference drift in
\cref{fig:main_benchmark}a is therefore a reference-semantics mismatch, not a
failure of the quantization-simulated replay.

\begin{table}[!t]
\centering
\caption{\textbf{Continuous-reference error across matched fixed-point \qat{}
replay splits.}
Matched replay gives zero final MAE, zero trace MAE, 100\% exact scalar-grid
faithfulness and gate agreement of 1.000 on every listed split.}
\label{tab:device_ref}

\small
\setlength{\tabcolsep}{5pt}
\renewcommand{\arraystretch}{1.15}

\begin{tabularx}{\columnwidth}{@{}Xr@{}}
\toprule
Evaluation split & Continuous MAE \\
\midrule
train seen/seen                         & 0.000516 \\
val seen/seen                           & 0.000499 \\
seen tasks / held-out lengths           & 0.000548 \\
held-out tasks / seen lengths           & 0.003397 \\
held-out tasks / held-out lengths       & 0.006027 \\
\bottomrule
\end{tabularx}
\end{table}

\subsection{Long-horizon quantization-simulated writeback extrapolation}

The replay result motivates a stricter extrapolation check. If the low-precision executor is following the intended symbolic path, increasing the number of executed instructions should not accumulate unbounded writeback error. In an extended 16-wide \qat{} value-memory benchmark at lengths 100, 200, 400 and 800, hard final \MAE{} remained low. The reported values were $1.20\times 10^{-4}$ at $L=100$, $6.38\times10^{-5}$ at $L=200$, $7.12\times10^{-5}$ at $L=400$ and $2.99\times10^{-5}$ at $L=800$, with hard gate agreement 1.0 at all lengths. The proxy energy increased approximately linearly with length, from 4028 at $L=100$ to 32245 at $L=800$, as expected for a per-instruction simulation cost. These results support length scalability of the implemented symbolic execution substrate, while still being limited to the synthetic program families used in this study.

\begin{table}[t]
\centering
\caption{\textbf{Extended long-horizon \qat{} value-memory profile.} Energy and delay are simulation proxies.}
\label{tab:long_horizon}
\begin{tabular}{rrrrr}
\toprule
Length & Hard final MAE & Hard trace MAE & Gate agreement & Mean proxy energy\\
\midrule
100 & $1.20\times10^{-4}$ & $1.31\times10^{-4}$ & 1.000 & 4028\\
200 & $6.38\times10^{-5}$ & $1.04\times10^{-4}$ & 1.000 & 8098\\
400 & $7.12\times10^{-5}$ & $1.22\times10^{-4}$ & 1.000 & 16136\\
800 & $2.99\times10^{-5}$ & $4.56\times10^{-5}$ & 1.000 & 32245\\
\bottomrule
\end{tabular}
\end{table}

The main experimental target remains the 16-wide profile with \qat{} writeback, because that setting provides the strongest low-precision evidence under the available evaluation budget. A separate 32-wide profile probes whether the same implementation can handle larger continuous state vectors before moving toward realistic processor semantics. Over the same short-length range used for the 16-wide profile, the 32-wide extrapolation trained instance has hard final \MAE{} 0.0 at lengths 40, 80 and 120, with gate agreement 1.0. This supports state-width scalability for the implemented continuous symbolic reference interpreter, but it should not be read as emulation of a full 32-bit instruction-set architecture. The remaining transition to processor realism requires typed immediates, branches, memory alignment, exceptions and system-level state.

\begin{table}[t]
\centering
\caption{\textbf{Supplementary 32-wide extrapolation profile: short-length range.} The result supports state-width scalability under controlled symbolic tasks.}
\label{tab:thirty_two_bit}
\begin{tabular}{rrrr}
\toprule
Length & Hard final MAE & Hard trace MAE & Gate agreement\\
\midrule
40  & 0.0 & 0.0 & 1.000\\
80  & 0.0 & 0.0 & 1.000\\
120 & 0.0 & 0.0 & 1.000\\
\bottomrule
\end{tabular}
\end{table}

The same 32-wide non-\qat{} trained instance reveals its limit only when the short-length grid is extended to long horizons (three evaluation seeds with eight benchmark batches each at $L\in\{400,\dots,1000\}$, Supplementary Fig.~\ref{fig:supp_32wide_longhorizon}). Execution is numerically exact through $L{=}400$. The first deviation appears at $L{=}500$ and is confined to the endpoint (final \MAE{} $2.0\times10^{-3}$ with trace \MAE{} still at the ${\sim}2\times10^{-5}$ floor). Beyond this onset the error grows rather than recovering. Final \MAE{} rises roughly an order of magnitude into $L{=}600$ ($2.1\times10^{-2}$) and then saturates between $4\times10^{-2}$ and $5\times10^{-2}$ through $L{=}1000$, while trace \MAE{} climbs monotonically ($2.4\times10^{-3}$ at $L{=}600$ to $1.9\times10^{-2}$ at $L{=}1000$) and gate-path agreement drifts down from $1.0$ to $0.968$. Scalar-grid-exactness nevertheless holds near $90\%$ across the entire range from $500$ to $1000$. The model does not smear all values uniformly but instead loses a minority (${\sim}10\%$) of scalars that drift increasingly far from grid. The separation between the exact short-horizon regime and the diverged long-horizon regime is statistically unambiguous (final \MAE{} $0$ at $L\le320$ versus $4.2\times10^{-2}$ at $L\ge600$, Mann--Whitney $U$ test, $p{\approx}5\times10^{-57}$, $n{=}160$ versus $120$ pooled benchmark samples). Thus the wider-state profile exhibits \emph{progressive, bounded degradation} past its training horizon. This is a soft extrapolation limit rather than a hard collapse, which reinforces the use of the length-robust 16-wide \qat{} profile as the primary low-precision benchmark. A controlled training ablation (below) shows that this limit is set by the training horizon of this particular trained instance (programs of at most 40 steps) rather than by the wider state itself, and is removed by training at a modestly longer horizon.

Placing every family on a matched evaluation horizon from $L{=}20$ to $L{=}1000$ makes the comparison symmetric (Supplementary Fig.~\ref{fig:width_comparison}). The breakdown is not a function of \emph{evaluation} length per se (every profile is exact through ${\sim}500$ steps), but appears only for the short-horizon-trained 32-wide instance at long evaluation lengths, while the controlled ablation below shows that the governing factor is the \emph{training} horizon rather than state width. At $W{=}16$ the non-\qat{} baseline is bit-exact at every length (final and trace \MAE{} $0$, $100\%$ scalar-grid-exact, $100\%$ tolerance), and the \qat{} families hold final \MAE{} in the $10^{-3}$ band with effectively exact operation paths through $L{=}1000$. Doubling the continuous state to $W{=}32$ keeps execution exact only through $L{=}400$ and then introduces the endpoint error, scalar-grid loss and gate drift documented above. A qualification is that even the $W{=}16$ non-\qat{} instance shows a small operation-path dip at $L\geq900$ (to ${\sim}96\%$) while its outputs remain bit-exact, consistent with occasional routing through an output-equivalent opcode at extreme lengths rather than an execution error. The short-horizon 32-wide onset coincides with a fixed \emph{absolute} step (${\approx}494$, Supplementary Fig.~\ref{fig:trace_divergence}c) rather than a fixed fraction of the program, which is the signature of a learned-horizon boundary rather than a width-capacity bound. This reading is confirmed directly by the controlled training ablation that follows.

To investigate why the 32-wide trained instance deviates at long horizons and whether the wider state is fundamentally limited, we retrained the identical 32-wide architecture (non-\qat{}, fixed seed) while changing one factor at a time and benchmarked each variant across the same sweep from $L{=}20$ to $L{=}1000$ (Supplementary Fig.~\ref{fig:t400_mechanism} and Table~\ref{tab:t400_mechanism}). Every variant is bit-exact through at least $L{=}520$, but the long-horizon behaviour separates cleanly by \emph{training horizon}. Training on programs up to 80 or 160 instructions removes the deviation entirely. Both are bit-exact (final and trace \MAE{} exactly $0$) through $L{=}1000$. By contrast, neither memory capacity nor training duration accounts for it. Varying the memory slots while keeping the maximum training length at 40 instructions is non-monotonic and follows no capacity law (halving to $128$ slots remains bit-exact through $L{=}1000$, while doubling to $512$ gives only a mild late drift, final \MAE{} $9.2\times10^{-4}$ at $L{=}1000$), and \emph{tripling} the training epochs to $45$ at the same maximum training length does not remove the deviation and in fact \emph{worsens} the far-horizon error ($7.0\times10^{-2}$ at $L{=}1000$, the largest of any variant), consistent with overfitting to the short trained length. The variability among the variants trained up to 40 instructions (some bit-exact, some developing long-horizon drift) is itself consistent with a marginal, seed-sensitive learned-horizon boundary at the short horizon that a longer training horizon removes decisively. We therefore conclude that the wider-state long-horizon breakdown is a \emph{learned training-horizon limit} of the short-horizon trained instance, not a capacity bound of the $W{=}32$ state. The $32$-wide executor is not fundamentally length-limited, and a modestly longer training horizon makes it bit-exact across the full sweep, matching the $16$-wide profile. The ${\approx}494$-step onset reported above is therefore specific to the instance trained with maximum length 40 rather than a fixed architectural constant.

\subsection{The control path is preserved while error localizes in the datapath}
\label{sec:exec_diagnostics}

The supplementary execution diagnostics explain why the aggregate curves in \cref{fig:main_benchmark} separate only under particular conditions. Across soft and hard decoding, gate confidence, per-operation calibration and per-step trace divergence (Supplementary Figs.~\ref{fig:soft_hard_gap} to \ref{fig:trace_divergence}), the same mechanism appears. The discrete \emph{control} path (which operation is executed at each step) is preserved across decoding modes, lengths and width profiles, whereas residual error localizes in the continuous \emph{datapath}. The 32-wide extrapolation series combines a two-seed low-length grid ($L\le320$) with a three-seed long-horizon sweep ($L\ge400$), without additional training.

A neural executor can be evaluated with stochastic (soft) or deterministic (hard) gates, and the critical question is whether the two modes disagree on \emph{what} is executed. Supplementary Fig.~\ref{fig:soft_hard_gap}b shows that soft-vs-hard gate-sequence agreement stays at $\geq99.9\%$ across the range from $L{=}20$ to $L{=}1000$ for all five profiles, so both modes traverse the same opcode sequence. The final states nevertheless drift apart (Supplementary Fig.~\ref{fig:soft_hard_gap}a). The soft-vs-hard final \MAE{} sits at the $10^{-4}$ level for the 16-wide non-\qat{} and \qat{} instances through $L{\approx}600$ and grows to $4.2\times10^{-4}$ and $2.8\times10^{-3}$ respectively by $L{=}1000$, while the 32-wide extrapolation instance jumps from $1.7\times10^{-4}$ at $L{=}400$ to $5.5\times10^{-2}$ at $L{=}700$. The long-program ValueMemory \qat{} instance is the limiting case. Its soft and hard executions are numerically identical at essentially every length (soft-vs-hard final \MAE{} ${\sim}10^{-16}$, at the floating-point floor, e.g.\ at $L{=}1000$), with a minority of individual benchmark batches (e.g.\ near the range from $L{=}80$ to $L{=}260$) showing a small non-floor residual up to ${\sim}2\times10^{-5}$. The two panels together establish that the soft and hard discrepancy is a continuous-datapath effect, never a difference in opcode selection.

Additionally, Supplementary Fig.~\ref{fig:gate_confidence} shows why the datapath drift appears only at long horizons. Gate entropy (Supplementary Fig.~\ref{fig:gate_confidence}a) and the top-two gate margin (Supplementary Fig.~\ref{fig:gate_confidence}b) are flat, with near-zero entropy and unit margin, through $L{\approx}400$ for all profiles, after which probability mass begins to spread. Counter-intuitively, the bit-exact non-\qat{} run is the \emph{least} soft-confident at extreme lengths (entropy $0.040$, margin $0.964$ at $L{=}1000$), the \qat{} family is intermediate (entropy $0.014$, margin $0.988$), and the long-program ValueMemory run is essentially unaffected (entropy ${\sim}2\times10^{-6}$, margin $1.000$ at every length). Read against Supplementary Fig.~\ref{fig:soft_hard_gap}b, this isolates the soft-mode calibration loss of \cref{fig:main_benchmark}c as a spreading of gate \emph{probabilities}, not a change in the \emph{argmax}. The most numerically exact model spreads the most probability mass while still selecting the correct opcode at essentially every step.

The Supplementary Fig.~\ref{fig:per_op_calibration} localizes the error to particular opcodes. Summarized across the eight operations, operation accuracy (under soft-gate decoding) holds at $100\%$ through $L{\approx}500$ for the 16-wide profiles and declines only mildly thereafter (Supplementary Fig.~\ref{fig:per_op_calibration}a), while per-operation calibration falls from $-\log_{10}\ECE\approx7$ to ${\approx}4$ for every family except the long-program ValueMemory track, which holds ${\approx}8$ throughout (Supplementary Fig.~\ref{fig:per_op_calibration}b). The per-opcode heatmaps make the structure explicit. For the 16-wide \qat{} family only \texttt{ADD} loses accuracy ($95.7\%$ at $L{=}1000$, while the other seven opcodes remain $100\%$), and \texttt{ADD} is also the worst-calibrated row (Supplementary Fig.~\ref{fig:per_op_calibration}c,d). For the 32-wide extrapolation profile the loss concentrates in \texttt{MUL} ($64.0\%$) and \texttt{SHR} ($66.5\%$) at $L{=}1000$, with \texttt{ADD} mildly affected ($97.4\%$) and the remaining five opcodes exact (Supplementary Fig.~\ref{fig:per_op_calibration}e,f). The carry- and shift-heavy arithmetic is therefore the first to leave the fixed-point grid. The degradation is operation-specific rather than a uniform collapse. Crucially, calibration degrades \emph{ahead of} accuracy, and reading the two heatmap rows against each other exposes this ordering. At $L{=}1000$ most opcodes still classify exactly yet shed up to ${\sim}3$ orders of magnitude of calibration score. In the 16-wide \qat{} family every opcode except \texttt{ADD} holds $100\%$ accuracy while \texttt{AND} and \texttt{SHL} fall to values from $-\log_{10}\ECE\approx2.6$ to $-\log_{10}\ECE\approx2.7$ (from values between ${\approx}4$ and ${\approx}6$ at $L{=}100$, Supplementary Fig.~\ref{fig:per_op_calibration}c,d), and in the 32-wide extrapolation profile \texttt{OR} remains $100\%$ accurate but drops to $-\log_{10}\ECE\approx2.3$ while \texttt{SHL}, \texttt{SUB} and \texttt{XOR} stay both exact and well-calibrated ($-\log_{10}\ECE\gtrsim4.6$, Supplementary Fig.~\ref{fig:per_op_calibration}e,f). Only the carry/shift-heavy \texttt{ADD}, \texttt{MUL} and \texttt{SHR} (where the probability spread grows large enough to flip the hard argmax) lose accuracy as well. This ordering follows directly from the two metrics measuring different quantities. Operation accuracy depends only on whether the correct opcode remains the argmax, whereas \ECE{} penalises the full confidence-versus-correctness gap, so the long-horizon spreading of gate probability documented in Supplementary Fig.~\ref{fig:gate_confidence} (rising entropy, shrinking top-two margin) erodes calibration first and changes the decision only in the extreme. Per-operation calibration is therefore the sensitive, leading indicator of approaching the execution horizon, and operation accuracy the lagging, coarser one. This is consistent with the same probability-spread-not-argmax-flip mechanism seen in the soft-vs-hard gate agreement of Supplementary Fig.~\ref{fig:soft_hard_gap}b.

Supplementary Fig.~\ref{fig:trace_divergence} resolves \emph{where within a program} the error appears. For the 16-wide \qat{} families the per-step trace \MAE{} rises within the first ${\sim}10\%$ of execution to the range from $2\times10^{-3}$ to $3\times10^{-3}$ and then stays flat for the remainder, while the non-\qat{} instance remains at the $10^{-7}$ floor (bit-exact at every step) (Supplementary Fig.~\ref{fig:trace_divergence}a). The \qat{} heatmap confirms that this early plateau is essentially length-independent (Supplementary Fig.~\ref{fig:trace_divergence}b). The 32-wide extrapolation profile behaves qualitatively differently. Its trace error is at the floor for the first half of long programs and then rises sharply (Supplementary Fig.~\ref{fig:trace_divergence}a,c). Across the matched low-length grid ($L\le320$) it stays pinned at the trace-\MAE{} floor at every step (the uniformly dark band at the bottom of Supplementary Fig.~\ref{fig:trace_divergence}c), confirming that the wider-state breakdown is a horizon effect. It appears only past the trained step budget rather than being a generic width penalty that is merely diluted at short lengths. The heatmap boundary is \emph{diagonal} (Supplementary Fig.~\ref{fig:trace_divergence}c) because the onset is at a fixed absolute step. Across $L\in\{500,\dots,1000\}$ the first step exceeding a $10^{-3}$ trace error is step ${\approx}494$ in every case, which maps to normalized progress $0.99$ at $L{=}500$ but $0.49$ at $L{=}1000$. The wider-state breakdown is thus a training-horizon limit (${\sim}500$ executed steps for this instance trained on programs of at most 40 steps, and removed by longer-horizon training, Supplementary Table~\ref{tab:t400_mechanism}), whereas the 16-wide target shows soft, bounded, non-accumulating error within the same horizon.

\subsection{Robustness tests, cost-aware execution and architectural limits}
\label{sec:stress_cost_limits}

The length benchmark and matched replay analysis establish the main positive result. With visible opcodes and an appropriate low-precision reference, the executor preserves the symbolic operation path over long programs. We next change one condition at a time to identify which part of the implemented system carries each result. \Cref{fig:stress_cost_limits} therefore brings together four contrasts involving corrupted inputs, equivalent lower-cost programs, architecture ablations and hidden-opcode memory tasks. The hidden-opcode memory-pressure condition is the negative control for general program induction. When the operation cue is withheld, the model must infer latent operation and memory-use structure rather than execute an observed instruction stream. Several control measurements are saturated and are more informative as text than as separate panels. Matched fixed-point replay is exact on all hard device-reference splits, min-energy and any-valid operation paths are fully gate-aligned in the energy-choice evaluation, and most structured comparison architectures solve the baseline visible-opcode distribution. These saturated controls anchor the interpretation of the non-saturated cases, where a cue is removed, the reference semantics change or the task becomes memory-mediated.

Input perturbations affect the continuous datapath rather than the opcode cue. The robustness tests add bit flips or least-significant-bit (LSB)-scaled perturbations to the model inputs during evaluation, where LSB denotes the simulated input quantization scale rather than a physical circuit bit. They do not simulate radiation, thermal noise, circuit faults or measured device noise. The opcode field and the supervised operation gate remain present, so these perturbations primarily disturb the continuous register values that feed the fixed operation bank and destination writeback. Accordingly, the strongest reported perturbations raise final error to the range from $7\times10^{-3}$ to $8\times10^{-3}$ and reduce tolerance faithfulness to the range from $40\%$ to $60\%$, while the symbolic operation path remains essentially intact (\cref{fig:stress_cost_limits}a and Supplementary Fig.~\ref{fig:supp_robustness_stress}). The mechanism is therefore not a sudden collapse of the controller. It is value drift in the differentiable register datapath under corrupted inputs, with the categorical operation selector still anchored by the visible instruction stream.

Equivalent-program tasks change the correct reference path. The energy-choice benchmark is not a test of canonical-trace imitation. Each sample stores a canonical program and a finite set of admissible lower-cost alternatives generated by the symbolic data generator. Under the min-energy policy, the correct behaviour is to leave the canonical path whenever a lower-proxy-cost trace reaches the same final state. Consequently, canonical-path agreement can be low or zero without indicating an error. The model is being scored against a different, valid symbolic execution. In the energy-choice evaluation, min-energy \qat{} execution has final \MAE{} $5.84\times10^{-4}$, any-valid trace \MAE{} $4.57\times10^{-4}$, zero mean energy gap relative to the minimum-energy reference and a 38.6\% mean proxy-energy reduction relative to canonical execution (Table~\ref{tab:energy_choice}). \Cref{fig:stress_cost_limits}b therefore plots proxy-energy saving against any-valid trace error, while the saturated min-energy and any-valid gate agreement is stated in the caption rather than used as a separate heatmap.

Architecture ablations identify what the trace supervision contributes. The architecture comparison benchmark compares \gru-\ntm, no-gate-loss, no-memory, transformer-\ntm, transformer-scratchpad, temporal convolution, graph, temporal graph and state-space variants on the same visible-opcode task family (Table~\ref{tab:model_zoo} and Supplementary Fig.~\ref{fig:supp_model_zoo_detail}). These comparisons are architectural diagnostics rather than capacity-matched pairwise comparisons. Parameter counts and memory interfaces differ across variants, so the supported inference is mechanistic rather than a claim of capacity-controlled superiority. Because several structured controllers reach the numerical display floor on this baseline task distribution, the informative evidence is not another near-floor error measurement. It is the contrast introduced by the diagnostic controls. Removing the gate loss leaves the state and trace losses to shape the router indirectly. The network can reduce register error while allowing the named operation logits to drift away from the reference operation sequence. This produces final \MAE{} 0.0412, trace \MAE{} 0.0489 and gate agreement 0.745 at $L=200$. The transformer scratchpad baseline is limited by a different implemented interface. It predicts traces without exposing the \alu{} gate and writeback pathway, so even when it learns part of the state trajectory it cannot provide the auditable operation path required by the symbolic executor. Thus the ablation result is not merely that one model has higher MAE. It shows that gate supervision and an explicit symbolic interface are the mechanisms that make the computation inspectable.

Hidden-opcode memory pressure removes the cue that visible-opcode execution relies on. The memory-pressure benchmark masks opcode hints and evaluates delayed copy, associative lookup, pointer chasing, stack-style reversal and long hidden-memory reversal. This changes the problem from executing a visible program to inferring a latent operation and memory-use pattern from register transitions and delayed state dependencies. On held-out memory-pressure tasks and lengths, the main \gru-\ntm{} model reaches final \MAE{} 0.343 and the no-memory ablation reaches 0.364. ValueMemory improves the held-out task and length split to 0.301, and \qat{} ValueMemory further improves it to 0.243, but this remains far from the visible-opcode regime. The detailed task split explains why (\cref{fig:stress_cost_limits}c,d and Supplementary Figs.~\ref{fig:supp_memory_pressure_aggregate},\ref{fig:supp_memory_pressure_task_breakdown}). Stack push and pop cases benefit from an addressable value store because they require repeated local retrieval and update, whereas long hidden-memory reversal requires sustained temporal binding, ordered delayed writes and operation inference over a much longer latent sequence. The low gate agreement in these panels is therefore not a contradiction of the near-exact visible-opcode gate result. It is the direct measurement of the harder condition in which the operation cue is withheld.

\subsection{The cross-benchmark signature mechanisms}

\begin{figure}[!tbp]
    \centering
    \includegraphics[width=\textwidth,height=0.65\textheight,keepaspectratio]{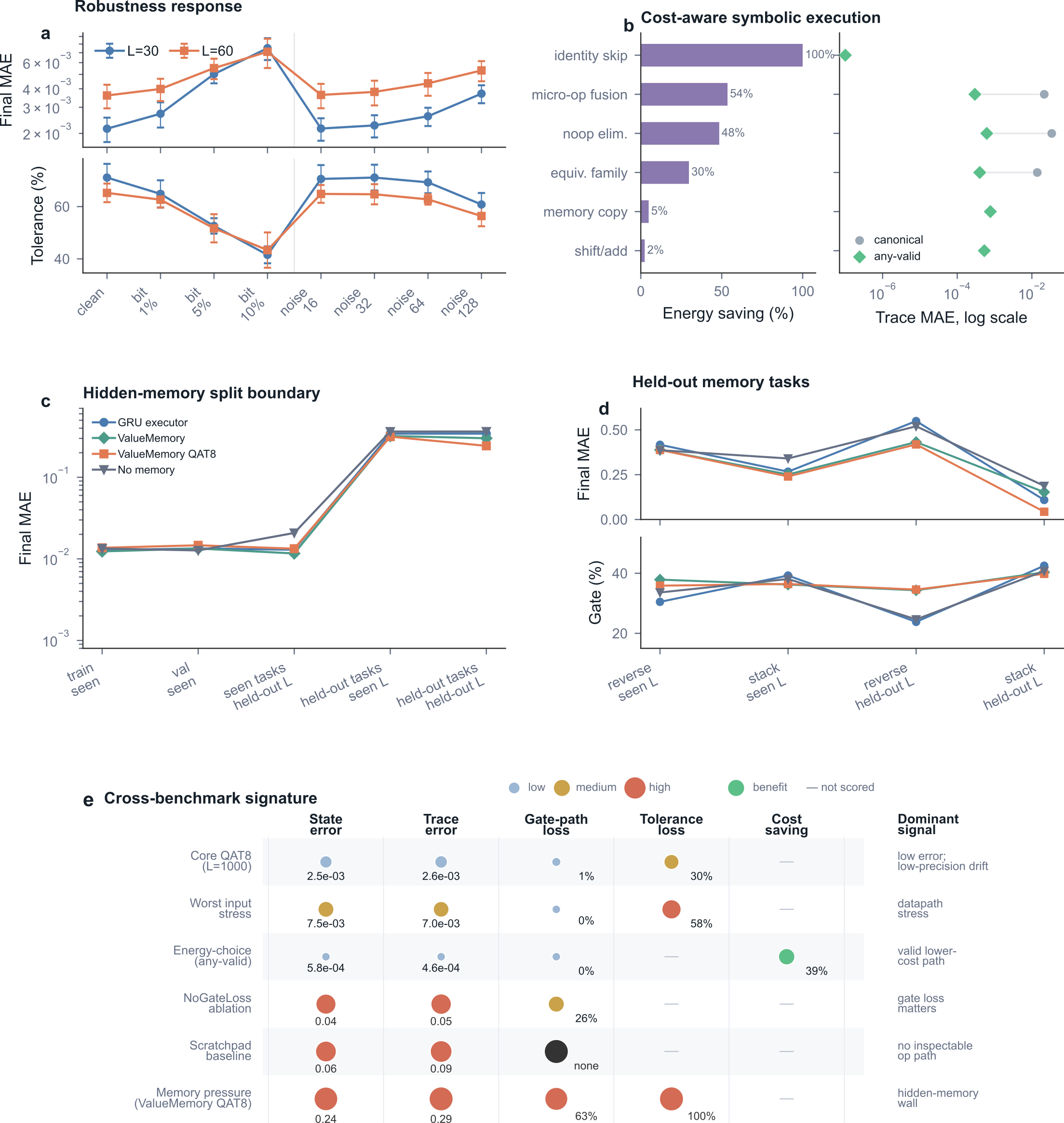}
    \caption[Robustness tests, cost-aware execution and hidden-memory limits.]
{\textbf{Robustness tests, cost-aware execution and hidden-memory limits.}
Metric formulas are defined in \cref{sec:methods_metrics}, and the energy-choice,
architecture-comparison and memory-pressure protocols are defined in
\cref{sec:methods_energy_choice}.
\textbf{(a)} Simulated model-input perturbations test whether bit-flip and LSB-noise
corruptions mainly affect continuous register values or the symbolic execution path.
\textbf{(b)} Equivalent-program cost choice tests whether the executor can select a
lower-proxy-cost valid trace rather than merely imitate the canonical trace.
\textbf{(c)} Hidden-opcode memory-pressure splits test how the main model, ValueMemory,
\qat{} ValueMemory and no-memory ablation behave when opcode cues are withheld.
\textbf{(d)} Held-out memory-task boundaries identify which delayed-memory tasks benefit
from ValueMemory and which remain difficult under hidden-opcode conditions.
\textbf{(e)} The cross-benchmark signature summarizes the dominant mechanism in each
stress condition by comparing state error, trace error, gate-path loss, tolerance loss
and cost saving. Detailed visual-scaling conventions and saturated-metric interpretations
are given in the surrounding text and Supplementary note~6.}
    \label{fig:stress_cost_limits}
\end{figure}

\Cref{fig:stress_cost_limits}e provides a mechanism map. State and trace entries use the measured \MAE{} values. Gate-path loss is $1$ minus gate agreement. Tolerance loss is $1$ minus tolerance faithfulness. Cost saving is shown as a benefit. Grey dashes indicate quantities outside a protocol, such as tolerance scoring for the energy-choice row, and the scratchpad row is marked as having no explicit operation path because the baseline does not expose the named \alu{} gate trace. Read from left to right, the row mechanisms are distinct. Core \qat{} execution retains the symbolic path but shows continuous-reference drift under low precision. Input perturbations affect values while preserving the opcode cue. Energy-choice succeeds under valid-path semantics and adds cost reduction. NoGateLoss removes the categorical constraint that pins operation identity. Scratchpad prediction lacks an inspectable gate and writeback interface. Hidden-opcode memory pressure combines state, trace, gate and tolerance degradation because it asks the model to infer latent operations and delayed memory use rather than execute a visible instruction stream.

\Cref{fig:stress_cost_limits} summarizes the robustness, cost-aware execution and
hidden-memory-limit analyses. Metric formulas are defined in
\cref{sec:methods_metrics}. The energy-choice, architecture-comparison and
memory-pressure protocols are defined in
\cref{sec:methods_energy_choice}.
Saturated and contrasting metrics are interpreted in Supplementary note~6.

\Cref{fig:stress_cost_limits}a reports robustness under simulated model-input
perturbations. The upper curve reports final \MAE{}, and the lower curve reports
tolerance faithfulness for clean execution, bit-flip perturbation and LSB-noise
perturbation at two program lengths. Error bars are benchmark-batch standard deviations.

\Cref{fig:stress_cost_limits}b reports the equivalent-program cost-choice benchmark.
The left subpanel shows proxy-energy reduction relative to canonical execution for each
task family. The right subpanel compares canonical-trace error and any-valid-trace error
on the same rows. Min-energy gate agreement and any-valid gate agreement are 100\% in
this run. Canonical-path agreement is low or zero when the selected lower-cost path
differs from the canonical trace.

\Cref{fig:stress_cost_limits}c reports hidden-opcode memory-pressure splits. Curves show
final \MAE{} for the main model, ValueMemory, \qat{} ValueMemory and no-memory ablation
across seen, length-held-out and task-held-out splits.

\Cref{fig:stress_cost_limits}d reports the held-out memory-task boundary. The upper
subpanel reports final \MAE{}, and the lower subpanel reports gate agreement for long
hidden-memory reversal and stack push and pop under seen and held-out lengths. ValueMemory
improves stack-style held-out cases, but long hidden-memory reversal remains difficult.

\Cref{fig:stress_cost_limits}e reports the cross-benchmark signature. Rows summarize the
principal 16-wide \qat{} family at $L{=}1000$, the worst simulated input-perturbation
condition, the energy-choice benchmark under any-valid scoring, the two informative
architecture negative controls, NoGateLoss and scratchpad, and the strongest \qat{}
ValueMemory model on the hardest memory-pressure split. Dot size and colour encode the
magnitude of state error, trace error, gate-path loss and tolerance loss. Green dots
encode cost saving as a positive benefit. For visual scaling, state and trace \MAE{} are
log-normalized between $10^{-3}$ and $3\times10^{-1}$. Gate-path loss is $1$ minus gate
agreement. Tolerance loss is $1$ minus tolerance faithfulness. Additionally, the cost saving is the
fractional proxy-energy reduction, grey dashes indicate metrics that are outside that
benchmark protocol rather than missing evaluations. ``none'' marks the scratchpad
baseline's absence of an explicit operation path, and the right-hand annotation states the
dominant interpretation of each row.

\begin{table*}[!t]
\centering
\captionsetup{justification=centering}
\caption{\textbf{Energy-choice execution summary.}
Energy is a dimensionless simulation proxy, not measured device energy.}
\label{tab:energy_choice}

\begin{minipage}{0.90\textwidth}
\centering
\small
\setlength{\tabcolsep}{5pt}
\renewcommand{\arraystretch}{1.15}

\begin{tabular*}{\textwidth}{
@{\extracolsep{\fill}}
l
c
c
c
c
c
@{}
}
\toprule
Task &
\shortstack{Final\\MAE} &
\shortstack{Any-valid\\trace MAE} &
\shortstack{Canonical\\energy} &
\shortstack{Model\\energy} &
Saving \\
\midrule
all                    & 0.000584 & 0.000457 & 1412.99 & 868.27  & 38.6\% \\
micro-op fusion        & 0.000175 & 0.000297 & 1504.87 & 698.50  & 53.6\% \\
redundant noop         & 0.000818 & 0.000618 & 1324.64 & 682.71  & 48.5\% \\
skip identity ops      & 0.000000 & 0.000000 & 1489.14 & 0.00    & 100.0\% \\
memory copy min-energy & 0.000870 & 0.000772 & 1410.57 & 1341.76 & 4.9\% \\
\bottomrule
\end{tabular*}

\end{minipage}
\end{table*}

\begin{table*}[!t]
\centering
\captionsetup{justification=centering}
\caption{\textbf{Representative architecture metrics at $L=200$.}}
\label{tab:model_zoo}

\begin{minipage}{0.90\textwidth}
\centering
\small
\setlength{\tabcolsep}{6pt}
\renewcommand{\arraystretch}{1.15}

\begin{tabular*}{\textwidth}{
@{\extracolsep{\fill}}
l
r
c
c
c
@{}
}
\toprule
Model &
Parameters &
\shortstack{Final\\MAE} &
\shortstack{Trace\\MAE} &
\shortstack{Gate\\agreement} \\
\midrule
\gru-\ntm main          & 160860 & 0.0000 & 0.0000 & 1.000 \\
\gru-\ntm \qat{}        & 160860 & 0.0038 & 0.0045 & 1.000 \\
No gate loss            & 160860 & 0.0412 & 0.0489 & 0.745 \\
No memory               & 160860 & 0.0000 & 0.0000 & 1.000 \\
Transformer-\ntm        & 482524 & 0.0000 & 0.0000 & 1.000 \\
Transformer scratchpad  & 438784 & 0.0587 & 0.0910 & no explicit gates \\
TCN-\ntm                & 284124 & 0.0000 & 0.0000 & 1.000 \\
Graph-\dcpu             & 147164 & 0.0000 & 0.0000 & 1.000 \\
TemporalGraph-\dcpu     & 246236 & 0.0000 & 0.0000 & 1.000 \\
SSM-\dcpu               & 111708 & 0.0000 & 0.0000 & 1.000 \\
\bottomrule
\end{tabular*}

\end{minipage}
\end{table*}

\subsection{Symbolic rollout diagnostics for the selected stack}
\label{sec:selected_stack_diagnostics}

\begin{figure*}[t]
    \centering
    \includegraphics[width=\textwidth]{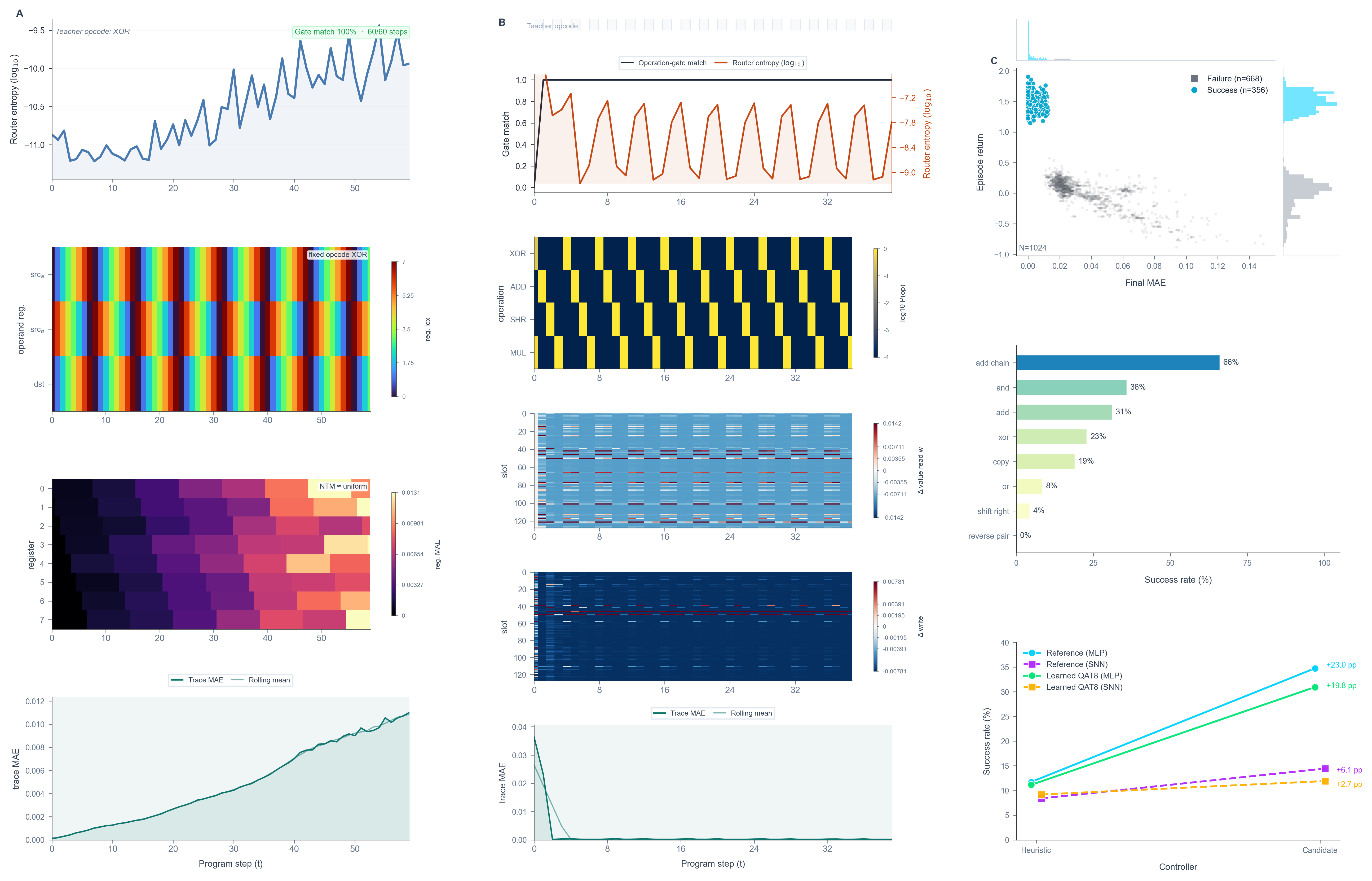}
    \caption{\textbf{Symbolic rollout diagnostics for the selected non-spiking stack.}
    Rollout variables, trace extraction and candidate-control protocol are defined in \cref{sec:selected_stack_diagnostics,sec:methods_closed_loop_control}.
    \textbf{(A)} Open-loop \gru-\ntm{} \qat{} rollout on \emph{reverse} ($L=60$, seed~7): constant reference opcode XOR (in-panel label); $\log_{10}$ router entropy with a 100\% gate-match badge (area fill for readability); operand routing for the two source registers and destination register (src$_a$, src$_b$, dst) in the constant-XOR program; NTM addressing is uniform so per-register trace MAE is shown instead; per-step trace MAE with rolling mean.
    \textbf{(B)} Open-loop \qat{} ValueMemory rollout on masked-opcode \emph{pointer-chase} ($L=40$, seed~21): segmented reference opcode strip; dual-axis gate match and $\log_{10}$ router entropy (with area fill); $\log_{10}$ operation probabilities; $\Delta$ value-memory read/write weights, where $\Delta$ is deviation from a uniform memory-slot distribution; per-step trace MAE with legend.
    \textbf{(C)} Closed-loop candidate control over $N=1024$ episodes (return versus final MAE): non-success density with marginal histograms and success overlay, per-task success bars, and paired heuristic$\to$candidate lifts for both factorized multilayer perceptron (MLP) and factorized \alif{} spiking neural network (SNN) policies on reference and learned-\qat{} transition suites. Exemplar repair traces are described in the main text. Columns~\textbf{A--B} are single-trace reference-conditioned execution; column~\textbf{C} is the candidate-control evaluation suite.}
    \label{fig:selected_stack}
\end{figure*}

Aggregate figures (\cref{fig:main_benchmark,fig:stress_cost_limits,fig:closed_loop_control} and Supplementary Fig.~\ref{fig:device_ref}) establish the benchmark-level outcome of the non-spiking executor, the memory-pressure variant and the closed-loop controller. A separate question is whether those outcomes correspond to interpretable executions on individual programs. \Cref{fig:selected_stack} therefore reports machine-visible observables including operation-gate alignment, router entropy, operation probabilities, memory addressing weights and per-step register trace error without relying on hidden-unit activations.

For clarity, we briefly define the plotted variables. At each program step $t$, the reference opcode is $o_t^{\mathrm{ref}}$ and the model produces gate logits $\ell_t$ and soft gate probabilities $p_t=\softmax(\ell_t/\tau)$. Operation-gate match is $m_t=\mathbf{1}[\arg\max_o \ell_{t,o}=o_t^{\mathrm{ref}}]$. Router entropy is $H_t=-\sum_o p_{t,o}\log(\max(p_{t,o},10^{-12}))$, displayed as $\log_{10}(H_t+\varepsilon)$. Operation heatmaps plot $\log_{10}P(\mathrm{op})$ from the same soft probabilities with a $10^{-4}$ floor. Memory-addressing heatmaps plot deviation from uniform, $\Delta w = w - 1/K$, for $K$ memory slots. Per-step trace error is $\mathrm{MAE}_t=\mathrm{mean}_{r,w}\lvert r^{\mathrm{pred}}_{t,r,w}-r^{\mathrm{ref}}_{t,r,w}\rvert$ over registers $r$ and scalar components $w$. In column~\textbf{C}, the joint plot shows episode return versus final MAE $\mathrm{MAE}^{\mathrm{final}}$, and success is $\mathbf{1}[\mathrm{MAE}^{\mathrm{final}}\leq 10^{-2}]$.

Column~\textbf{A} is an open-loop, reference-conditioned rollout of the principal \gru-\ntm{} \qat{} executor on a \emph{reverse} program ($L=60$, seed~7, selected from three candidate seeds by highest gate agreement and lowest mean trace MAE). Every reference step uses XOR, so the opcode strip is collapsed to an in-panel label. The gate/router row reports two complementary views of symbolic alignment, namely hard-gate agreement summarized as a badge and soft routing uncertainty via $\log_{10}(H_t+\varepsilon)$ for router entropy $H_t=-\sum_o p_{t,o}\log p_{t,o}$ with a subtle area fill for readability. On this exemplar, operation-gate match is exact ($m_t\equiv 1$, gate agreement $=1.0$) while entropy begins near a numerical floor and then rises episodically late in the program, indicating that the temperature-scaled softmax distribution can temporarily broaden even when the hard decision remains reference-aligned. Because $P(\mathrm{op})$ is degenerate for constant XOR, the operation panel instead shows reference register-operand routing (src$_a$, src$_b$, dst) over time, making explicit that the executor is still performing structured register selection even without opcode variation. NTM read/write weights are indistinguishable from uniform on this trace, showing that this example is solved through register routing and local writeback rather than selective slot addressing. Because the operation probabilities and NTM addressing are uniform on this trace, the displayed diagnostic is a per-register trace-\MAE{} map. The bottom trace plot shows that register drift remains modest (mean trace MAE $\approx 4.9\times 10^{-3}$), with a light rolling-mean overlay to distinguish local spikes from longer-term accumulation.

The \emph{reverse} exemplar is useful because its reference control flow is simple (constant opcode) while its data movement is non-trivial. The executor must repeatedly route operands and update destinations in the correct order. A case with uniform NTM addressing makes the explicit interpretability point that \emph{external memory is not always the mechanism}, even in an NTM-augmented architecture. When the task can be solved from registers and local instruction structure, the auditable objects are gates, operand routing and register-level error rather than slot selection. This prepares the comparison to memory pressure, where addressing becomes selective and therefore scientifically meaningful rather than an always-on visualization.

Column~\textbf{B} applies the same layout to the strongest memory-pressure model, \qat{} ValueMemory, on a masked-opcode \emph{pointer-chase} program ($L=40$, seed~21). Opcode one-hot hints are masked while register operands remain visible, and a slim strip marks reference opcode segments without overprinting. Unlike column~\textbf{A}, gate agreement is high but not exact ($97.5\%$), so the gate/router row uses dual axes with operation-gate match $m_t$ on the left and $\log_{10}(H_t+\varepsilon)$ on the right, again with a light entropy fill to emphasize oscillatory structure. The operation-probability heatmap (log$_{10}\,P(\mathrm{op})$) shows that the model maintains a structured distribution over plausible operations despite hidden opcodes, and that uncertainty varies systematically with reference opcode-transition boundaries. Crucially, the memory rows become informative because value-memory read/write weights deviate from uniform in sparse, repeating bands, consistent with pointer-style retrieval and periodic updates to a small subset of slots plotted as $\Delta w = w - 1/K$. The trace-MAE curve decays rapidly and remains near zero for most steps (mean $\approx 1.7\times 10^{-3}$), connecting \emph{interpretable memory use} to low register error on this exemplar while remaining consistent with aggregate degradation on harder held-out memory families (\cref{fig:stress_cost_limits}d).
Column~\textbf{C} moves from single-trace diagnostics to population-level control outcomes. The joint plot shows $N=1024$ closed-loop evaluation episodes from the reference-transition factorized multilayer-perceptron (MLP) candidate controller. It presents final MAE versus episode return with non-success density shown by a hexbin and successful episodes overlaid as points, while marginal histograms summarize densities along both axes and the legend reports the success and non-success counts. Two clusters emerge, with successes concentrating at $\mathrm{MAE}^{\mathrm{final}}\leq 10^{-2}$ and high return while non-success episodes spread along a diagonal band at larger MAE and low or negative return. This indicates that return is largely explained by proximity to the target state rather than by diffuse exploration. The per-task bar chart reports success rates for the most frequent tasks in the evaluation log and highlights harder families such as shift-right and low-level boolean chains even under candidate masking. Finally, the paired lift panel compares heuristic-only control with candidate-masked fine-tunes using the control summary table. On the \emph{reference} suite, the factorized MLP improves from $11.7\%$ to $34.8\%$ success ($+23.0$ pp) while the factorized adaptive leaky integrate-and-fire (\alif{}) spiking neural network (SNN) improves from $8.4\%$ to $14.5\%$ ($+6.1$ pp). On the \emph{learned-\qat{}} backend, the MLP improves from $11.1\%$ to $31.0\%$ ($+19.8$ pp) while the SNN improves from $9.2\%$ to $11.9\%$ ($+2.7$ pp). This column is \emph{not} task-matched to columns~\textbf{A} and~\textbf{B}. It summarizes controllability under candidate-constrained search across a multi-task suite rather than replaying the same reference programs.

Two representative repair traces illustrate the closed-loop policy (beam width~4, depth~2, up to 512 candidates per step). A \emph{successful} episode on the \texttt{shift\_left} task (episode~2, one repair step) applies a single symbolic action,
\begin{quote}
\small
\texttt{SHL R4, R3 -> R6}
\end{quote}
and reaches $\mathrm{MAE}^{\mathrm{final}}=0$. A \emph{non-successful} episode on shift to right (episode~1, four repair steps) instead proposes a short multi-instruction sequence,
\begin{quote}
\small
\texttt{MUL R0, R3 -> R3 | MUL R6, R5 -> R3 | MUL R5, R3 -> R3 | MUL R6, R5 -> R3}
\end{quote}
which does not recover the target state ($\mathrm{MAE}^{\mathrm{final}}>10^{-2}$). These examples illustrate that success is determined by whether the candidate-constrained search finds a short symbolic correction, not by a single scalar summary alone.

\subsection{Hybrid spiking-controller extension}

Spiking controllers test whether the same symbolic execution surface can be paired with event-driven temporal state. They are not presented as replacements for the non-spiking \qat{} executor. The implemented design restricts spikes to the controller dynamics. Instruction and register information are encoded as deterministic current input, the \alif{} module supplies temporal controller memory, and the downstream operation router, continuous \alu{} bank, destination write and ValueMemory or NTM pathways remain the same inspectable symbolic interface used by the non-spiking models. This hybrid construction preserves the named operation path while exposing membrane and spike variables that are relevant to future neuromorphic implementations. Supplementary Fig.~\ref{fig:snn_pilot} summarizes feasibility, and \cref{fig:snn_signals} provides rollout-level spike and membrane diagnostics.

For one symbolic program step, the implemented controller expands the encoded instruction and memory input into $K$ micro-steps. The state variables are current $c$, membrane voltage $u$, binary spike state $z$ and adaptive threshold state $a$. For \alif{} neurons, each micro-step follows.
\begin{align}
    c_{k+1} &= \alpha_c c_k+d+W_{\mathrm{rec}}z_k,\\
    u_{k+1}^{\mathrm{raw}} &= \alpha_u u_k+c_{k+1},\\
    z_{k+1} &= \mathrm{surrogate\_spike}(u_{k+1}^{\mathrm{raw}}-(\theta_0+\beta a_k)),\\
    u_{k+1} &= u_{k+1}^{\mathrm{raw}}(1-z_{k+1}),\\
    a_{k+1} &= \gamma a_k+z_{k+1},
\end{align}
matching the implemented spiking-controller update. The forward spike is a hard threshold, while the backward pass uses the implemented fast-sigmoid surrogate gradient. The returned controller feature is not a raw event list. It is the mean spike state over the $K$ micro-steps, concatenated with the final membrane state and projected through the controller readout. The spiking ValueMemory model then computes read weights from the previous spike state, blends the read value into the register operands, routes through the same \alu{} operation bank, clamps the writeback to $[0,1]$, and updates the value memory with a soft write address.

In the memory-pressure setting, the \alif{} ValueMemory substep-4 instance reaches held-out final MAE 0.447, trace MAE 0.419, gate agreement 39.2\% and spike rate 23.7\%. It therefore does not outperform the non-spiking \qat{} ValueMemory baseline. It serves instead to make the neuromorphic route concrete and inspectable. Supplementary Fig.~\ref{fig:snn_pilot} reports an 8-wide feasibility evaluation, the sensitivity to micro-step count and spike regularization, and the memory-pressure extension. \Cref{fig:snn_signals} then shows how operation gates, membrane voltage, averaged spikes and trace error evolve on individual programs.

\subsection{Interpretable spiking rollout diagnostics}
\label{sec:snn_rollout_diagnostics}
\begin{figure*}[t]
    \centering
    \includegraphics[width=\textwidth]{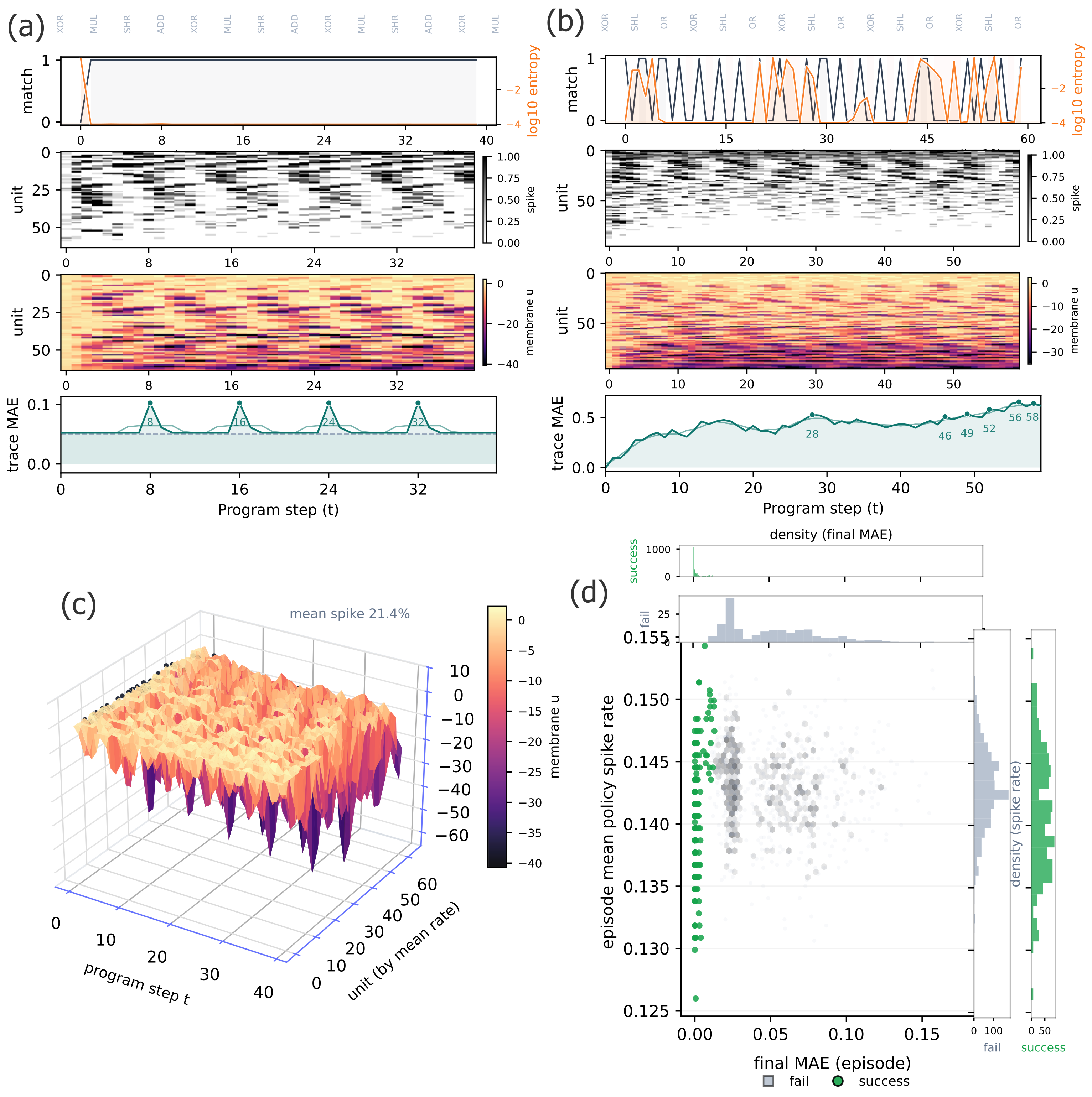}
    \caption{\textbf{Spiking controller rollout diagnostics and closed-loop episode outcomes.}
    Spiking variables, rollout extraction and episode-level control definitions are given in \cref{sec:snn_rollout_diagnostics,sec:methods_snn_signals,sec:methods_closed_loop_control}.
    \textbf{(a)} Open-loop rollout on a held-out pointer-chase memory-pressure program ($L=40$, seed~7, masked opcode hints). Top row: reference opcode labels. Second row: operation-gate match $m_t$ (left axis) and $\log_{10}$ router entropy (right axis). Third row: hidden-unit spike raster ($\bar{z}_{t,i}$, units sorted by mean rate). Fourth row: membrane potential $u_{t,i}$ (display clipped at 2nd--98th percentiles). Bottom row: per-step trace MAE with optional reference band.
    \textbf{(b)} Same layout for long reverse with hidden memory ($L=60$, seed~7).
    \textbf{(c)} Three-dimensional membrane landscape for the panel~\textbf{(a)} rollout with spike markers above the surface (mean spike rate $21.4\%$).
    \textbf{(d)} Closed-loop evaluation over $N=1024$ episodes: final register MAE versus mean policy spike rate for a factorized \alif{} controller with candidate masking and beam search (success if $\mathrm{MAE}^{\mathrm{final}}\leq 10^{-2}$; green, success; grey, non-success; marginal densities shown). Panels \textbf{(a--c)} are single-trace reference-conditioned execution; panel \textbf{(d)} is the multi-task control evaluation suite.}
    \label{fig:snn_signals}
\end{figure*}
The aggregate \alif{} ValueMemory metrics set the boundary for the spiking extension on held-out memory pressure. The substep-4 instance reaches final MAE 0.447, trace MAE 0.419 and gate agreement 39.2\%, and therefore does not replace the non-spiking \qat{} ValueMemory baseline. Those scalar metrics do not show how a trained spiking controller behaves on individual programs. \Cref{fig:snn_signals} links symbolic execution, hidden-unit event dynamics and closed-loop population statistics in one layout. The analysis uses the selected \alif{} ValueMemory trained instance with four micro-steps per program step, deterministic hard gates, gate temperature $\tau=0.5$ and masked opcode hints on memory-pressure programs. All register errors are computed in normalized $[0,1]$ space, consistent with the reference and benchmark implementation.

Panels \textbf{(a)} and \textbf{(b)} report \emph{open-loop}, trace-conditioned rollouts on fixed reference programs. This is not autonomous program synthesis because at each step $t$ the model receives the reference instruction stream, with opcode one-hot masked when required, and the evolving register state, while the neural executor predicts operation gates, register traces, hidden spikes and membrane voltages. Panel \textbf{(a)} shows a held-out \emph{pointer-chase} program of length $L=40$ (seed~7). After an initial transient, operation-gate match $m_t=\mathbf{1}[\arg\max_o \ell_{t,o}=o_t^{\mathrm{ref}}]$ remains high, while router entropy $H_t=-\sum_o p_{t,o}\log p_{t,o}$ with $p_{t,o}=\softmax(\ell_t/\tau)$ collapses toward a peaked distribution, plotted as $\log_{10}(H_t+\varepsilon)$. The spike raster displays micro-step-averaged events $\bar{z}_{t,i}$ with hidden units sorted by decreasing mean firing rate, while the membrane heatmap shows the post-substep voltage $u_{t,i}$ clipped for display between the 2nd and 98th percentiles. Trace mean absolute error,
\begin{equation}
    \mathrm{MAE}_t=\frac{1}{|\regset|W}\sum_{r,w}\left|r^{\mathrm{pred}}_{t,r,w}-r^{\mathrm{ref}}_{t,r,w}\right|,
\end{equation}
stays small but exhibits periodic peaks aligned with opcode-transition boundaries, shown by vertical guides on the heatmaps. Panel \textbf{(b)} applies the same visualization to a harder held-out \emph{long reverse with hidden memory} program ($L=60$, seed~7). Gate match oscillates, spike structure becomes less regular and $\mathrm{MAE}_t$ drifts upward with late-episode spikes. The juxtaposition supports a concrete error-accumulation interpretation. On the easier pointer structure, the controller remains symbolically aligned with low register drift, whereas under longer hidden-opcode memory pressure the dominant error is temporal accumulation rather than an isolated catastrophic step.

Panel \textbf{(c)} is a three-dimensional membrane landscape for the \textbf{same} rollout as panel \textbf{(a)}. The surface plots $u_{t,i}$ over program step and sorted unit index, while markers indicate steps with $\bar{z}_{t,i}>0.5$. Mean spike activity over the episode is $21.4\%$ in this example. This view places spike events and membrane voltages in a single geometry, making structured temporal locking visible without relying on separate rasters. Its interpretation is tied to panel \textbf{(a)} rather than to an independent benchmark score.

Panel \textbf{(d)} summarizes a \emph{different} experimental protocol involving closed-loop symbolic control on $N=1024$ evaluation episodes drawn from the multi-task control suite. A factorized \alif{} policy, fine-tuned with candidate-masked imitation and reinforcement learning, selects actions under a reference CPU transition model with at most $512$ heuristic candidates per step and beam search with width~4 and depth~2. Each point shows episode final register MAE to the target state versus the mean policy spike rate accumulated over the episode, and success is defined by $\mathrm{MAE}^{\mathrm{final}}\leq 10^{-2}$. Successful episodes cluster near zero final MAE, while non-success episodes spread to larger MAE at similar spike rates. Marginal densities separate success and non-success along both axes. This panel therefore addresses population-level controllability and the decoupling of mean spiking from target achievement. It must not be interpreted as a task-matched replay of panels \textbf{(a)} to \textbf{(c)}.

\Cref{fig:snn_signals} makes the spiking extension inspectable at the level of operations, registers and spikes rather than only scalar MAE. It also separates the two regimes visible in the aggregate results. On an easy masked-opcode program, \alif{} dynamics are structured and the gate path is stable, while on a hard program, trace error grows with the temporal-binding burden. The population panel then connects these rollout diagnostics to the closed-loop controller pathway, showing that spike rate alone does not explain autonomous-control success. In the present system, the dominant lever for control is symbolic action search over a constrained candidate set.

\subsection{Closed-loop candidate control}
\label{sec:closed_loop_control}

Having established supplied-program execution, we next test whether the same transition substrate can be used to choose instructions. A controller observes a current register state and a target register state and selects the next symbolic instruction. This turns execution into a short-horizon repair problem. Success means reaching the target state within a final-state tolerance, whereas the earlier figures measure whether a supplied program is executed with the correct trace and gate path. This setting tests whether trace-verified execution can support controllable action selection.

Formally, the controller defines an MDP
\begin{equation}
    \mathcal{M}=(\mathcal{S},\mathcal{A},P,R,\gamma),
\end{equation}
where the transition $P$ is either the reference CPU transition or the learned 8-bit \qat{} Neural CPU transition. The action space contains symbolic instruction tuples. Candidate-constrained control defines a state-dependent candidate set
\begin{equation}
    \mathcal{C}(s)\subset\mathcal{A},
\end{equation}
constructed by symbolic heuristics and local state and target comparisons. The policy distribution is masked as follows.
\begin{equation}
    \pi_{\theta}^{\mathcal{C}}(a\mid s)
    =
    \frac{\pi_{\theta}(a\mid s)\mathbf{1}[a\in\mathcal{C}(s)]}
    {\sum_{a'\in\mathcal{C}(s)}\pi_{\theta}(a'\mid s)}.
\end{equation}
Evaluation optionally uses beam search. For a beam element $b=(s_{1:t},a_{1:t-1})$, a candidate action receives score
\begin{equation}
    J(b,a)=
    \log\pi_{\theta}^{\mathcal{C}}(a\mid s_t)
    -10\,\MAE(F(s_t,a),s^\star)
    -w_E E(a)
    +\mathbf{1}[\MAE(F(s_t,a),s^\star)\leq\epsilon],
\end{equation}
where $s^\star$ is the target, $w_E$ is the environment energy weight and $\epsilon$ is the success tolerance. The coefficient 10 and the unit success bonus rank immediate target-error reduction above log-probability and proxy energy during beam evaluation. The candidate generator also uses fixed offsets for primitive actions and operation-tie priors. These constants are reported as algorithmic search choices rather than as physically derived quantities.

\begin{figure}[!tbp]
    \centering
\includegraphics[width=\textwidth,height=0.65\textheight,keepaspectratio]{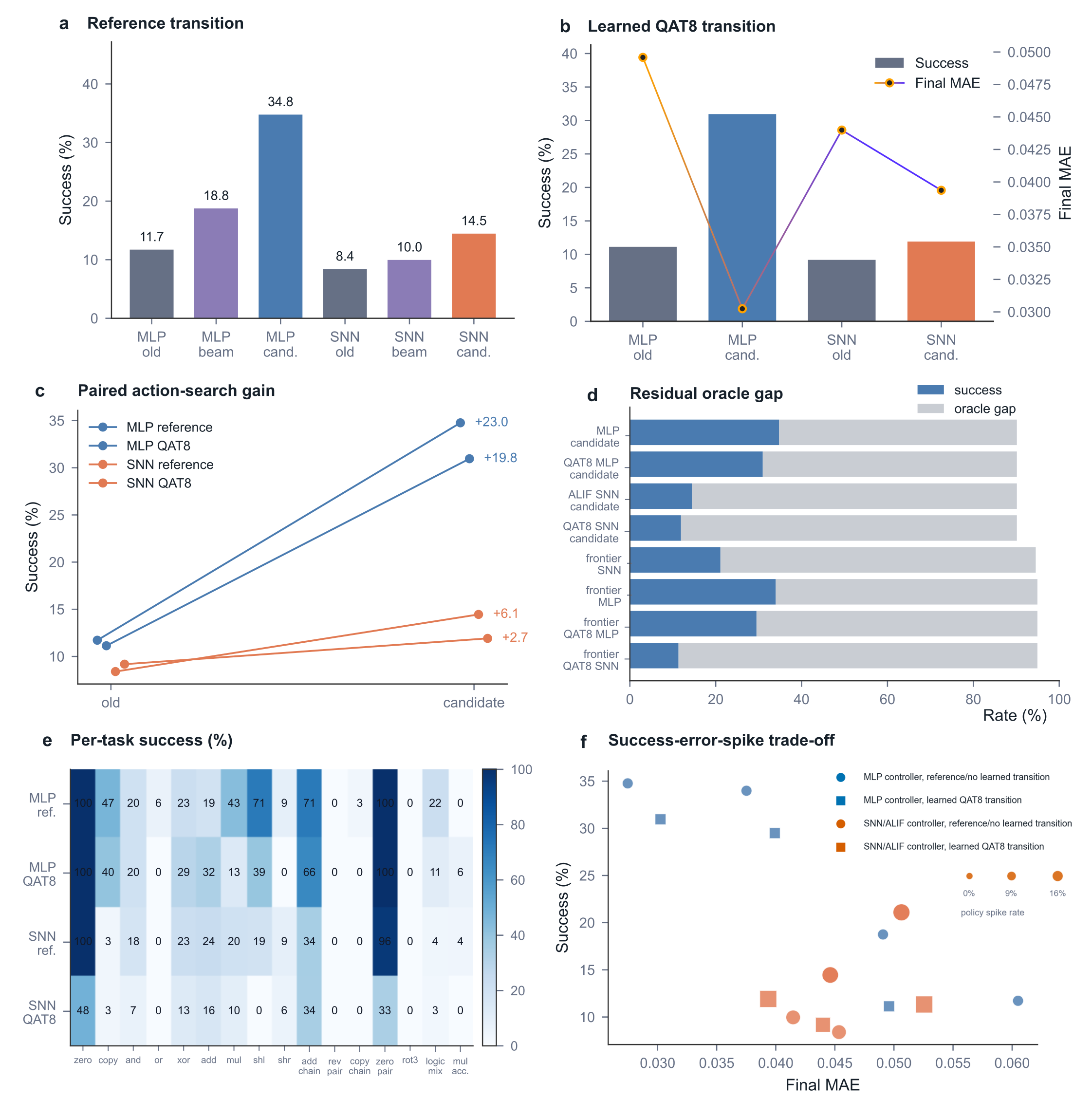}
    \caption[Candidate-constrained symbolic control.]
    {\textbf{Candidate-constrained symbolic control.}
    \textbf{(a)} Reference-transition control measures target-reaching success for MLP
    and \alif{} SNN policies under direct, beam and candidate-constrained evaluation.
    \textbf{(b)} Learned-\qat{} transition control repeats the same closed-loop protocol
    through the learned 8-bit transition model and relates success to final-state error.
    \textbf{(c)} Paired action-search gain shows the success-rate improvement from direct
    evaluation to candidate-constrained evaluation for each controller/back-end pair.
    \textbf{(d)} Residual oracle gap compares achieved success with the remaining gap to
    an oracle selector under the same task distribution.
    \textbf{(e)} The per-task success matrix identifies which symbolic control tasks are
    solved by each reference and learned-\qat{} MLP/SNN candidate policy.
    \textbf{(f)} The success-error-spike relation compares success rate, final-state
    \MAE{} and recorded spike rate across controller/back-end settings.}
    \label{fig:closed_loop_control}
\end{figure}

\Cref{fig:closed_loop_control} summarizes three evaluation modes, including direct policy evaluation, beam-style action selection without candidate fine-tuning, and candidate-constrained fine-tuning followed by candidate-constrained evaluation. In the figure labels, ``direct'' denotes the non-candidate policy baseline and ``cand.'' denotes the candidate-constrained policy. On the reference transition, candidate constraints increase factorized MLP success from 11.7\% to 34.8\%. Beam evaluation alone reaches 18.8\%. The factorized \alif{} SNN improves from 8.4\% to 14.5\% under the same candidate protocol. On the learned \qat{} transition, the MLP improves from 11.1\% to 31.0\%, while the SNN improves from 9.2\% to 11.9\%. The paired view reports the same effects as absolute gains. The gains are +23.0 percentage points for the reference MLP, +19.8 percentage points for the learned-\qat{} MLP, +6.1 percentage points for the reference SNN and +2.7 percentage points for the learned-\qat{} SNN.

These results define a control extension rather than a replacement for the open-loop execution benchmark. The per-task matrix separates simple state-setting and copy-like tasks from more compositional control families, and the residual oracle-gap panel quantifies the remaining distance to an oracle action selector under the same task suite, including the harder controller settings from the action-search analysis. The remaining gap has two concrete sources in the implemented protocol. First, an executable action is a tuple of operation, source registers and destination register. Even after candidate masking, many candidates are locally plausible but do not form the ordered multi-step repair needed by tasks such as reverse-pair, copy-chain or accumulated arithmetic. Beam scoring uses local, short-horizon rewards, so it can prefer an immediate reduction in final-state MAE that blocks a later correction. Second, in the learned-\qat{} setting, the controller scores actions through an approximate transition model. Small transition errors can change the ranking of candidate actions even when the underlying open-loop executor remains trace-faithful under replay. The success, error and spike panel shows that mean spike activity alone does not explain controllability. Policies can have comparable recorded spike activity while producing different final errors and success rates. The main conclusion is structural. When the symbolic action space grows, unconstrained neural policies allocate probability mass over many irrelevant instructions. Candidate generation reintroduces a symbolic prior at decision time and lets neural scoring operate over an inspectable action set. This supports the central thesis that the strongest systems in this setting are structured hybrids rather than pure black-box controllers.

\Cref{fig:closed_loop_control} summarizes the candidate-constrained symbolic-control
extension. Candidate sets, masked policies, beam scoring, success criteria and oracle-gap
definitions are given in \cref{sec:closed_loop_control,sec:methods_closed_loop_control}.
The figure compares direct policy evaluation, beam-style action selection and
candidate-constrained fine-tuning/evaluation for factorized multilayer perceptron (MLP)
and factorized \alif{} spiking neural network (SNN) policies. In the in-panel labels,
``direct'' denotes the non-candidate baseline and ``cand.'' denotes candidate-constrained
evaluation.

In our investigation, \Cref{fig:closed_loop_control}a reports reference-transition control. Bars show the
percentage of episodes reaching the target register state for factorized multilayer
perceptron (MLP) and factorized \alif{} spiking neural network (SNN) policies. The three
bars correspond to direct policy evaluation, beam-style action selection and
candidate-constrained fine-tuning/evaluation.

\Cref{fig:closed_loop_control}b shows the learned-\qat{} transition control. The same
closed-loop protocol is run through the learned 8-bit transition model. The bars show success
and the overlaid line reports the final-state \MAE{}.

\Cref{fig:closed_loop_control}c reports paired action-search gain. Each line connects
direct policy evaluation to the candidate-constrained policy for the same controller and
transition back-end pair. Annotations give absolute success-rate gains in percentage
points.

\Cref{fig:closed_loop_control}d reports the residual oracle gap. Blue segments show
achieved success, and grey segments show the remaining gap to the oracle selector under
the same task distribution, including the harder controller settings.

\Cref{fig:closed_loop_control}e reports the per-task success matrix for reference and
learned-\qat{} MLP/SNN candidate policies. Rows are controller/back-end families and
columns are symbolic control tasks. Entries are success rates in percent.

\Cref{fig:closed_loop_control}f reports the success, error and spike relation. Points compare
success rate with final-state \MAE{} across controller/back-end settings. Marker size
follows the recorded spike-rate field where available and is used as a visual cue rather
than a hardware-energy measurement.

Together, \cref{fig:closed_loop_control} shows that candidate-constrained symbolic action
search improves short-horizon repair, with the clearest gains for MLP policies, while the
learned transition and spiking controllers define the next control boundary.

\subsection{An RV32I base-integer bridge tests the architecture against a standardized ISA}
\label{sec:rv32i_bridge_results}

The preceding experiments use a controlled symbolic register-machine distribution because it exposes the complete execution trajectory, including operation labels, destination masks, register traces, memory signals and low-precision replay. As an external-validity test, we implemented an isolated RV32I-facing extension. The question is deliberately narrower than official \riscv{} architectural compliance. Can the symbolic neural CPU execution ingredients, including structured instruction representation, recurrent control, value-addressed memory and learned \alu{} routing, be connected to an implemented RV32I base-integer semantic bridge with explicit PC, memory, writeback and trap semantics?

The bridge decodes 32-bit RV32I instruction words into opcode, register, function and immediate fields, then applies deterministic state transitions over 32 architectural registers, byte-addressed memory, a program counter and an explicit trap state. The audited inventory covers the RV32I base-integer entries LUI, AUIPC, JAL, JALR, conditional branches, loads, stores, immediate \alu{} operations, register-register \alu{} operations, FENCE, ECALL and EBREAK. Register x0 is forced to zero after every step. The transition implements signed immediate extension, branch and jump target updates, JAL/JALR link behaviour, JALR low-bit clearing, load/store width and sign-extension behaviour, alignment checks, out-of-bounds memory traps and illegal-instruction traps. In this implementation, FENCE is treated as a single-hart semantic no-op, and ECALL/EBREAK enter explicit trap states.

The deterministic RV32I base-integer bridge passed 40/40 implemented inventory checks, 28/28 edge-case tests, 55,524/55,524 randomized property checks and 7/7 signature-memory harness tests. This deterministic audit validates the reference bridge and harness rather than learned execution. The harness executes raw instruction words, records a signature-memory region and supports expected signature, trap and register checks, but it is not yet the full \texttt{riscv-arch-test} compilation, linking and reporting flow.

The same RV32I teacher traces then diagnose learned semantic control. The pre-rescue RV32I learned-control diagnostic exposed the failure modes that mattered for inclusion, including trap state, JALR, load/store mode selection, held-out program-exact sequences and length extrapolation. The rescue controller exposes decoded RV32I bitfields and symbolic predicates to the learned heads, including instruction-family/name indicators, immediate lanes, branch comparison predicates, effective addresses, memory alignment and bounds predicates, jump-target alignment and destination-register-zero predicates. We repeated this symbolic-predicate controller over seeds 42, 43 and 44. Across seeds, hard semantic modes were recovered at $100.0\pm0.0\%$ accuracy for trap, JALR, load, store and memory-trap decisions. Program-level step agreement was $100.0\pm0.01\%$ on held-out programs, $100.0\pm0.02\%$ on branch/memory challenge programs and $99.2\pm0.26\%$ on length extrapolation. Program-exact sequence accuracy was $92.2\pm2.7\%$, $89.9\pm3.5\%$ and $69.2\pm9.9\%$ on the same three splits, respectively. For reference, the corresponding pre-rescue values were 30.4\% held-out exact, 24.1\% stress exact, 1.8\% length exact, 79.8\% length step agreement, 37.5\% trap, 60.9\% JALR, 81.4\% load and 80.1\% store. Program-exact sequence scoring is stricter than step-level agreement because a single wrong PC, memory or writeback decision can invalidate the whole sequence. The RV32I result is therefore architecture-aligned semantic-control evidence, not reliable execution of arbitrary \riscv{} binaries.

Together, the deterministic bridge and learned-control diagnostic support a precise extension of the main claim. The symbolic neural CPU architecture is not confined to synthetic opcodes. Its structured execution interface can be attached to an RV32I base-integer semantic substrate when binary decode, PC update, memory behaviour and traps are represented explicitly, and when learned control is trained over architecture-aligned predicates. The result is an RV32I base-integer bridge and learned-control diagnostic, not a claim that the main trained executor runs arbitrary \riscv{} binaries end to end.

\section{Discussion}

The results establish a trace-verifiable form of neural symbolic execution in
which a learned controller updates register-machine state while exposing the
operation path, register trajectory and writeback events required for
step-by-step inspection. The object of interest is therefore not merely the final register file, but the complete trajectory of register states, operation gates and writeback events. On visible-opcode programs, the selected \gru-\ntm{} family preserves this execution object with near-exact hard operation paths, while the primary trained runs maintain exact hard gate agreement across the reported horizon. The matched fixed-point reference is central to this interpretation. When the independent replay uses the same low-precision writeback grid as the \qat{} executor, final-state and trace disagreement vanish on the hard splits. The residual drift against the continuous interpreter is therefore a mismatch between continuous writeback semantics and 8-bit writeback semantics, rather than evidence that the learned executor has lost the symbolic path it was trained to follow.

The model comparisons show why this result is more than sequence prediction. The operation-gate loss is the categorical term that ties each router decision to a named operation. Removing it leaves the state and trace losses active, so the model can still approximate useful register values, but the operation logits become weakly constrained. Mixtures in the continuous operation bank can then compensate for endpoint error while erasing the inspectable opcode sequence. The NoGateLoss ablation therefore degrades trace fidelity and gate agreement for a mechanistic reason, not merely because one loss weight changed. The scratchpad baseline exposes the complementary limitation. It predicts register trajectories without a named operation gate or destination-writeback interface, so its output is a trajectory estimate rather than a replayable symbolic execution trace. Together, these controls separate endpoint approximation from the processor-like property tested in this article, namely whether a reader can inspect which operation was selected and which state variable was written at every step.

The robustness, cost-aware and memory-pressure experiments reveal which information supports each behaviour. Input bit flips and LSB-scaled perturbations corrupt continuous register values but leave the visible operation cue intact. The resulting degradation is therefore value drift in the datapath. Tolerance faithfulness declines and final \MAE{} rises, while the categorical operation path remains stable. Equivalent-program energy-choice tasks change the reference problem instead. A lower-proxy-cost valid program can differ from the canonical trace, so canonical agreement is not the appropriate success criterion. The relevant evidence is any-valid trace error together with proxy-energy reduction. Hidden-opcode memory pressure removes the operation cue itself. The model must then infer operation identity, address use and delayed writeback structure from state transitions. ValueMemory improves stack-like retrieval and update tasks because it gives the controller an addressable value pathway. Long hidden-memory reversal remains difficult because latent operation inference, temporal ordering and delayed destination writes must be solved together over long sequences. This is the clearest unresolved boundary of the current executor.

The rollout panels translate aggregate metrics into observable execution mechanisms. In the selected non-spiking trace, hard gates are exact while soft router entropy broadens late in the program. Hard-gate alignment therefore does not require the soft distribution to be fully degenerate. The same trace shows that NTM slots can remain uniform when a program is solved through visible instructions, operand routing and local writeback. In that regime, the operative audit variables are the gates, source and destination registers, and register-error profiles rather than selective memory slots. The memory-pressure rollout shows the opposite regime. With opcode hints masked, ValueMemory read and write weights become sparse and structured, while operation probabilities vary with the reference opcode segments. The scalar aggregate is therefore linked to an interpretable mechanism rather than remaining opaque. The model can discover useful addressable value structure, but that mechanism is not yet sufficient for the hardest held-out temporal-binding families.

The spiking-controller and closed-loop control extensions test how far the same interface can be carried beyond supplied-program execution. The implemented spiking design confines spikes to the temporal controller state while keeping instruction encoding, the symbolic \alu{} and writeback non-spiking and inspectable. This hybrid choice is important. \alif{} dynamics can represent event-driven temporal memory, while arithmetic and destination writes remain named program operations. The evidence supports feasibility, instrumentation and a device-facing research route rather than superiority over the strongest non-spiking executor. On harder hidden-memory programs, trace error grows with the temporal-binding burden. At the population level, mean spike rate alone does not explain control success. The larger closed-loop gains instead come from candidate-constrained symbolic action search. Candidate generation reduces the combinatorial operation, source and destination action space before policy scoring, which explains why the factorized MLP policies benefit most. The residual oracle gap remains because the controller must still rank symbolic tuples under finite-horizon rewards, while learned-\qat{} transition evaluation introduces approximation error into that ranking.

The RV32I bridge sharpens the same architectural principle at a standardized instruction-set boundary. The difficult modes in that experiment, including trap state, JALR, load and store modes, held-out sequences and length extrapolation, were not resolved by making the model less structured. They were resolved by making the interface more architectural. Decode, immediate reconstruction, PC update, memory access, x0 suppression and trap predicates are represented explicitly, while learned heads select among semantic-control modes above that substrate. Repeating the symbolic-predicate controller across three seeds resolved the targeted weak modes, including 100\% mode-level accuracy for trap, JALR, load, store and memory-trap decisions. Program-exact length extrapolation nevertheless remains below exact sequence execution. This strengthens the central claim because it shows that the architecture can be coupled to a standardized base-integer ISA substrate rather than only to synthetic symbolic opcodes. It also defines a precise boundary for future work. The reported RV32I result is a base-integer semantic bridge with an official-style signature harness. It is not official \riscv{} compliance, privileged execution, ABI/runtime support, M/A/C/F/D/V extension support or execution of arbitrary \riscv{} binaries by the main trained executor.

Translation to physical implementation remains staged and testable. The current device reference is an independent fixed-point replay of destination writes. It is not RTL, FPGA firmware, microcontroller code, neuromorphic hardware or \riscv{} execution. Energy and delay remain dimensionless accounting proxies, while spike rates are averages of returned spike traces rather than measured event energy. The next physical step is to export the implemented operation bank and low-precision writeback semantics to a fixed-point kernel, verify value-level agreement against the matched replay, and then measure latency and energy on FPGA, microcontroller or neuromorphic hardware. The most credible first demonstrator is a quantized symbolic coprocessor or controller module that proposes and verifies register-state transitions, rather than a full general-purpose CPU replacement.

A broader industrial motivation is edge and embedded anomaly detection on constrained devices, where multivariate sensor streams must be monitored under tight latency, energy and interpretability constraints. The present work does not evaluate a specific industrial deployment. Its contribution is the inspectable, quantization-simulated execution substrate on which a later anomaly-detection controller could be built, tested under matched replay and validated through hardware-in-the-loop experiments.

Several limitations define the scope of these claims. The teacher is a continuous register-vector interpreter with named CPU-like operations, not a bit-exact integer instruction-set emulator, and all training and evaluation programs are synthetic. Quantization is simulated at the writeback boundary and verified against a matched fixed-point replay. We report no measured silicon energy, physical delay, physical fault injection or circuit-noise validation, and the energy and delay figures remain dimensionless proxies. The spiking-controller and closed-loop control experiments establish feasibility of the shared execution interface rather than superiority over the strongest non-spiking executor, while autonomous control remains unsolved. The RV32I result is a base-integer semantic bridge and learned-control diagnostic. It is not official \riscv{} compliance, privileged execution, ABI/runtime support or execution of arbitrary \riscv{} binaries. More broadly, the contribution is not a general differentiable neural computer, neural algorithmic reasoner or program synthesizer. It is a controlled register-machine substrate in which learned execution can be checked step by step against reference semantics.

In summary, a trace-supervised symbolic neural CPU can execute controlled register-machine programs while keeping the operation path, register trajectory and low-precision writeback semantics open to inspection. Under eight-bit quantization-simulated writeback, the symbolic operation path remains effectively exact through programs of 1{,}000 instructions. The residual numerical drift arises from the continuous reference rather than from failure of the learned execution, and it vanishes under a matched fixed-point replay. The remaining boundaries involving hidden-opcode memory pressure, the training-horizon limit of the wider profile and the oracle gap in closed-loop control define concrete next steps toward learned executors that are inspectable enough to be trusted.

\section{Data and Code Availability}

The source code, experiment configurations, plotting scripts and source-data
tables required to reproduce the reported figures and benchmark summaries are
available from the corresponding author upon reasonable request. The author
will also deposit materials in the public GitHub repository
\href{https://github.com/jseluis/symbolic-neural-cpu}
{\texttt{jseluis/symbolic-neural-cpu}}. The repository will
include the implementations, experiments configurations, evaluation scripts,
figure-generation scripts and derived source-data tables used in this study. No industrial field dataset is used in the present experiments.
Device-facing deployment remains a future application direction.

\section{Acknowledgements}

This work was prepared as part of a broader research program on neuromorphic and quantized controllers for industrial anomaly detection. The present article establishes the symbolic execution substrate. Future work will address device-facing anomaly detection and hardware-in-the-loop deployment. The research program from which this work emerged was initiated under the
SNIC Small Compute project \textit{Artificial Intelligence for Physics and
Engineering: Modeling and Simulation}, project No.~SNIC 2022/22-843, conducted
at Link\"oping University under the principal investigatorship of the author
\cite{silva2022aiphysicsengineering}. That project established the broader
research direction connecting artificial intelligence, graph neural networks,
deep reinforcement learning, neuromorphic computing, and the modeling and
simulation of physics and engineering systems.

This work was supported by the Brazilian National Council for Scientific and
Technological Development, CNPq, under grant No.~445344/2024-5. The author
acknowledges financial support through the Talentos Brasil Fellowship as Project Coordinator. The author also acknowledges the
Group of Artificial Intelligence in Applied Geophysics, GAIA, at the Federal
University of Bahia, UFBA, for computational resources, research infrastructure
and institutional support.

\section{Author Contributions}

J.L.L.J.S. conceived the research direction, developed the \sncpu{} implementation and benchmark suite, designed and executed the experiments, interpreted the results and prepared the manuscript.

\section{Competing Interests}

The author declares no competing interests.

\clearpage

\section*{Supplementary Information}
\subsection*{Supplementary figures}
\renewcommand{\thefigure}{S\arabic{figure}}
\renewcommand{\figurename}{Supplementary Figure}
\setcounter{figure}{0}

\begin{figure}[t]
    \centering
    \includegraphics[width=0.62\textwidth]{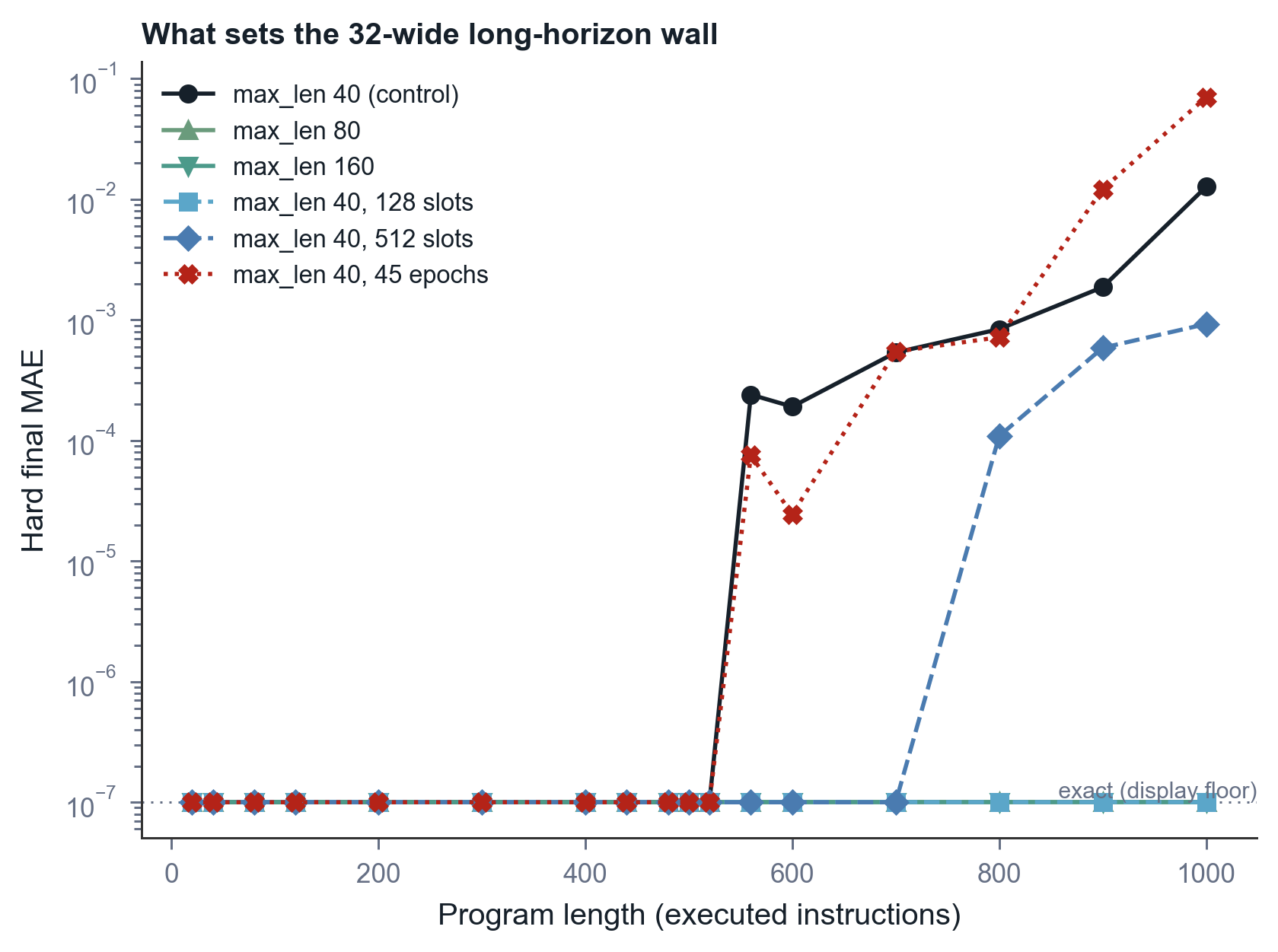}
    \caption{\textbf{What sets the 32-wide long-horizon limit: a controlled training ablation.} Hard final \MAE{} versus program length (log axis; exact/zero values drawn at a $10^{-7}$ display floor) for the identical 32-wide non-\qat{} architecture retrained while varying one factor at a time (numbers in Supplementary Table~\ref{tab:t400_mechanism}). The three short-horizon configurations trained on programs of at most 40 steps---the baseline configuration, the $512$-memory-slot variant and the $45$-epoch variant---peel off the floor beyond $L{\approx}500$--$700$, whereas training on longer programs (up to 80 or 160 steps) keeps execution bit-exact through $L{=}1000$. Memory-slot count is non-monotonic (the $128$-slot run stays exact) and more epochs \emph{worsen} the far-horizon error, so the limit is governed by the training horizon, not state width, memory capacity or training duration.}
    \label{fig:t400_mechanism}
\end{figure}

The supplementary figures provide the detailed diagnostics underlying the main figures. Robustness, calibration, proxy-cost, energy-choice, architecture comparison and memory-pressure analyses support \cref{fig:stress_cost_limits}; their metric definitions follow \cref{sec:methods_metrics} and Table~\ref{tab:notation}. The figures are organized around four checks: (i) how low-precision replay changes reference semantics, (ii) whether the discrete operation path remains stable when soft confidence changes, (iii) which operations and timesteps carry residual error, (iv) and which harder task families require latent operation or memory-use inference beyond visible-opcode execution. The controlled training-horizon ablation (Supplementary Fig.~\ref{fig:t400_mechanism}) is the only supplementary comparison that required additional model training.

\begin{figure*}[b]
    \centering
    \includegraphics[width=\textwidth]{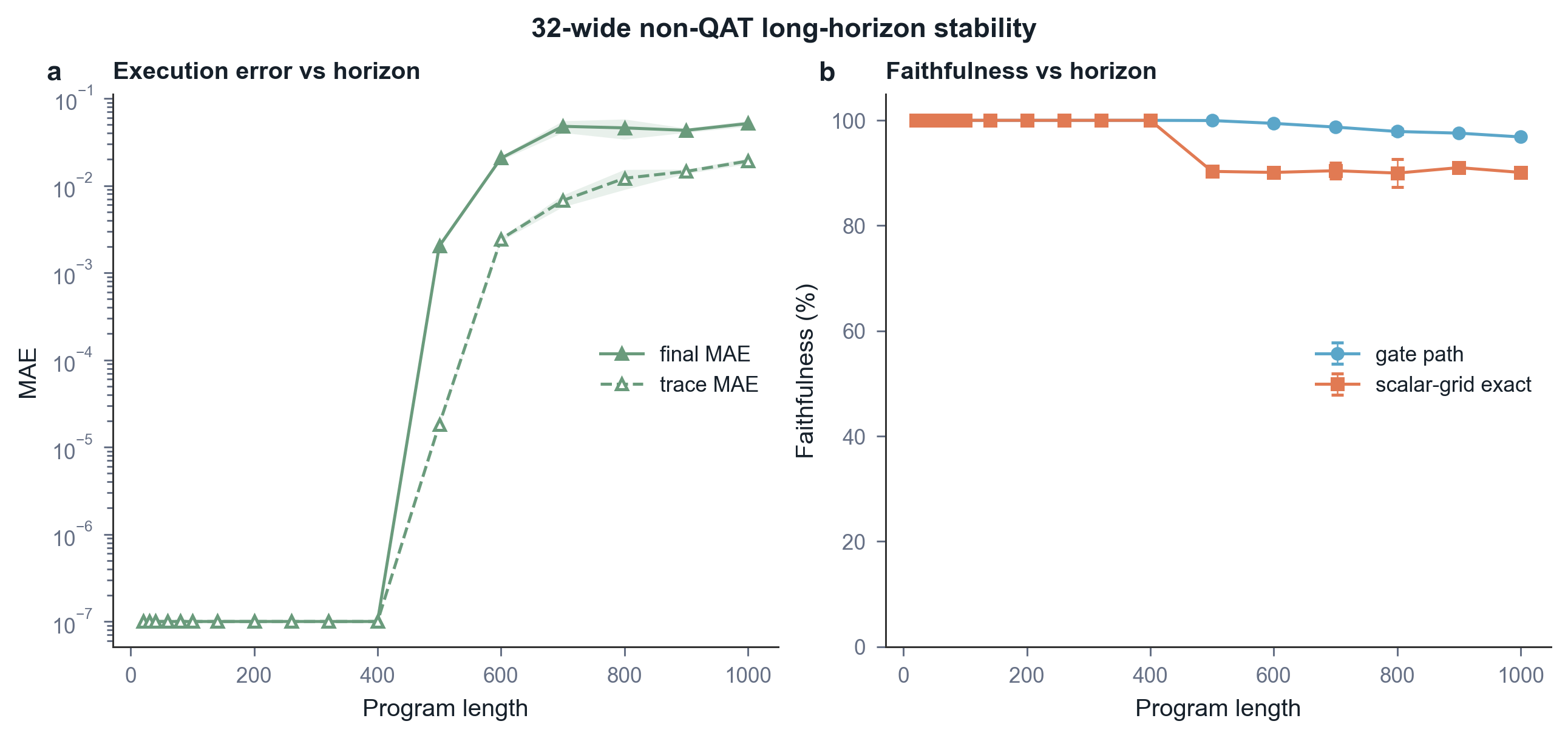}
    \caption{\textbf{Long-horizon stability of the 32-wide non-\qat{} extrapolation profile.} The 32-wide continuous executor evaluated across program lengths $L\in\{20,\dots,1000\}$. Lengths $L<400$ (i.e.\ $\le320$) reuse the single-seed short/medium-length benchmark grid (numerically exact there, so they carry no across-seed band); lengths $L\ge400$ are the mean over three evaluation seeds $\times$ eight benchmark batches per length, with bands and bars showing across-seed standard deviation. \textbf{(a)} Final \MAE{} (solid) and trace \MAE{} (dashed) on a logarithmic axis; exact (zero) values are drawn at a $10^{-7}$ display floor and the across-seed band is shown only where the metric exceeds that floor. Execution is exact through $L{=}400$; an endpoint-only deviation appears at $L{=}500$, after which final \MAE{} saturates near $4$--$5\times10^{-2}$ while trace \MAE{} keeps growing, indicating progressive---no longer endpoint-confined---trajectory drift beyond the training horizon. \textbf{(b)} Faithfulness: gate-path agreement drifts down from $1.0$ to ${\sim}0.97$, whereas scalar-grid-exactness stays near $90\%$ throughout, showing that the degradation concentrates in a minority of off-grid scalars rather than a global collapse of the opcode path. This figure characterizes a training-horizon extrapolation limit of this short-horizon-trained instance (programs of at most 40 steps)---removed by retraining the same architecture at a longer horizon (Supplementary Fig.~\ref{fig:t400_mechanism})---rather than a limit of the wider state itself; the main-text benchmark (\cref{fig:main_benchmark}) uses the robust 16-wide \qat{} profile.}
    \label{fig:supp_32wide_longhorizon}
\end{figure*}

\begin{figure*}[t]
    \centering
    \includegraphics[width=\textwidth]{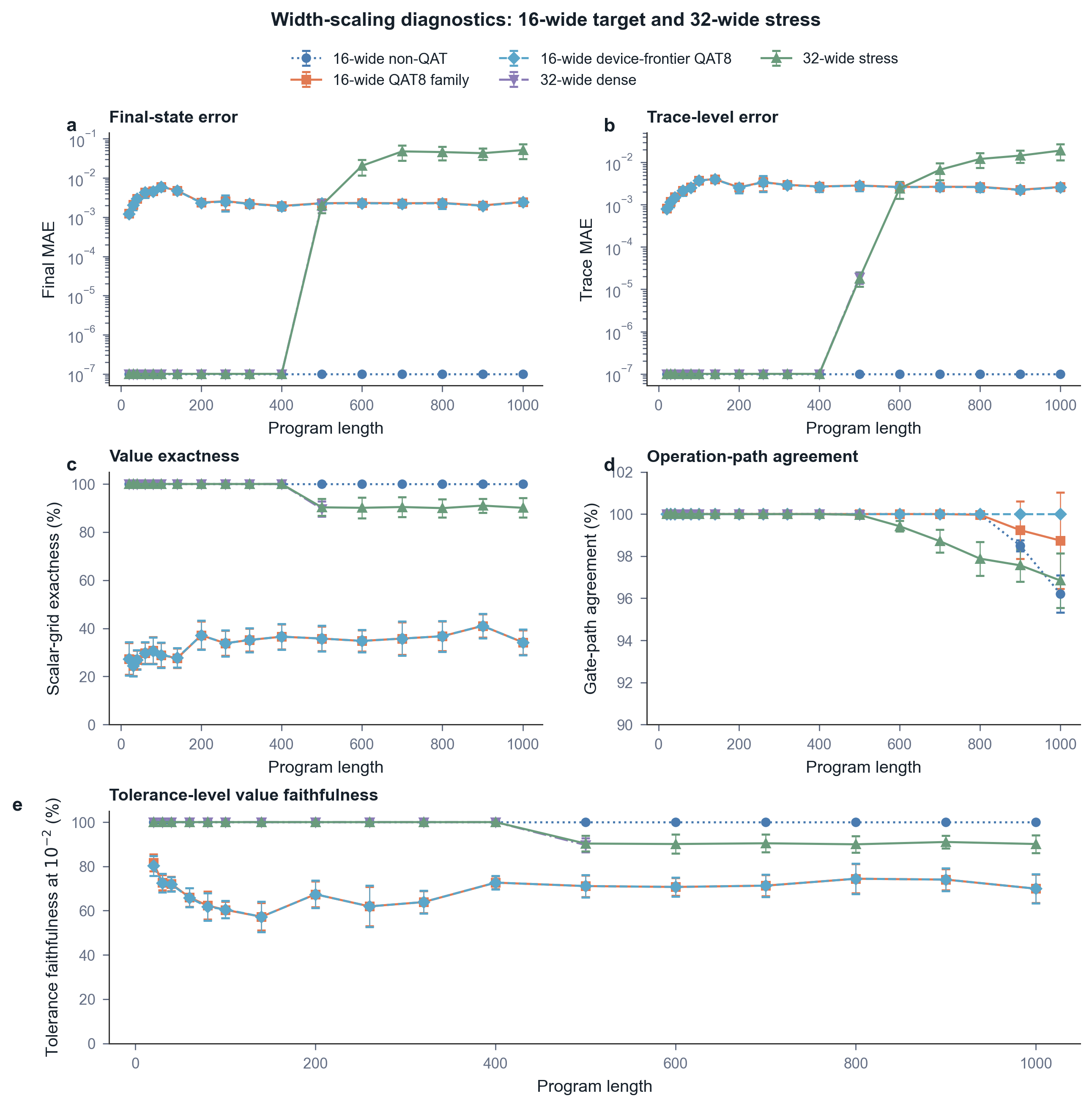}
    \caption{\textbf{Width-scaling diagnostics: 16-wide target versus 32-wide extrapolation.} The 16-wide families of \cref{fig:main_benchmark} together with the 32-wide non-\qat{} profile, all on a matched $L\in\{20,\dots,1000\}$ horizon. ``32-wide dense'' is the single-seed dense run; ``32-wide extrapolation'' is the multi-seed sweep---two seeds at $L\le320$ and three seeds at $L\ge400$, so the across-seed band (standard deviation) tightens at the $L{=}320\to400$ boundary as the seed count increases. \textbf{(a,b)}~Final-state and trace \MAE{} (log axis; exact values at the $10^{-7}$ floor): 16-wide non-\qat{} stays on the floor at every length, the 16-wide \qat{} families stay in the $10^{-3}$ band, and only the 32-wide profile rises beyond $L{=}500$. \textbf{(c)}~Scalar-grid exactness: $100\%$ for 16-wide non-\qat{} throughout and ${\sim}90\%$ for 32-wide beyond $L{=}500$, while the \qat{} families sit at $23$--$41\%$ against the \emph{continuous} reference (matched replay is exact, Supplementary Fig.~\ref{fig:device_ref}). \textbf{(d)}~Operation-path (gate) agreement, axis zoomed to $90$--$102\%$: the long-program ValueMemory \qat{} run holds $100\%$, the \qat{} family and the 32-wide extrapolation run drift to ${\sim}97$--$99\%$ by $L{=}1000$, and even 16-wide non-\qat{} dips to ${\sim}96\%$ at $L\geq900$ despite bit-exact outputs (output-equivalent routing). \textbf{(e)}~Tolerance faithfulness at $10^{-2}$: $100\%$ for 16-wide non-\qat{}, ${\sim}56$--$74\%$ for the \qat{} families, ${\sim}90\%$ for 32-wide beyond $L{=}500$. Means aggregate benchmark batches and available seeds.}
    \label{fig:width_comparison}
\end{figure*}

\begin{figure*}[t]
    \centering
    \includegraphics[width=\textwidth]{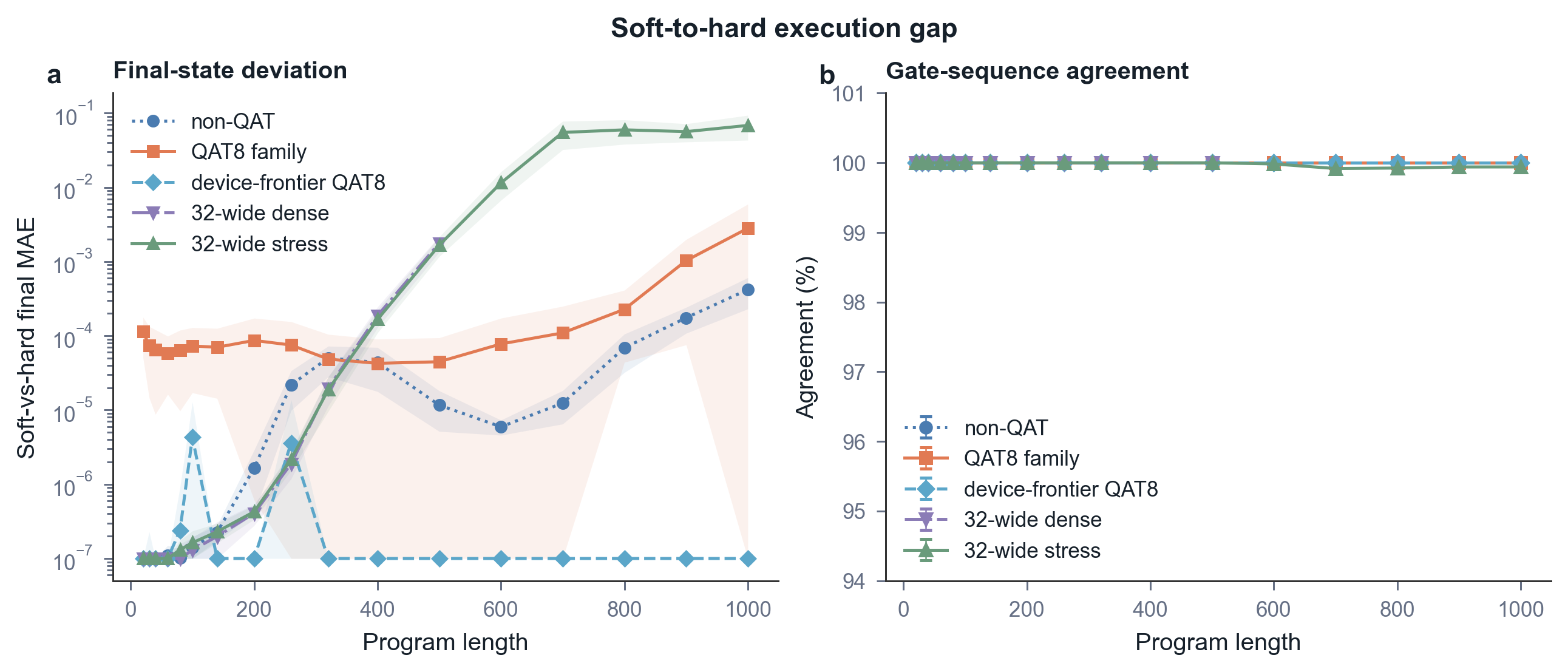}
    \caption{\textbf{Soft-to-hard execution gap.} \textbf{(a)}~Soft-vs-hard final \MAE{} (log axis): the 16-wide non-\qat{} and \qat{} runs agree to ${\sim}10^{-4}$ through $L{\approx}600$ and diverge to $4.2\times10^{-4}$ and $2.8\times10^{-3}$ by $L{=}1000$; the 32-wide extrapolation run diverges sharply beyond $L{=}400$; the long-program ValueMemory \qat{} run is identical in both modes (${\sim}10^{-16}$, at the floating-point floor). \textbf{(b)}~Soft-vs-hard gate-sequence agreement stays $\geq99.9\%$ for every profile and length, so the gap in (a) is entirely a continuous-datapath effect, not a difference in opcode selection.}
    \label{fig:soft_hard_gap}
\end{figure*}

\begin{figure*}[t]
    \centering
    \includegraphics[width=\textwidth]{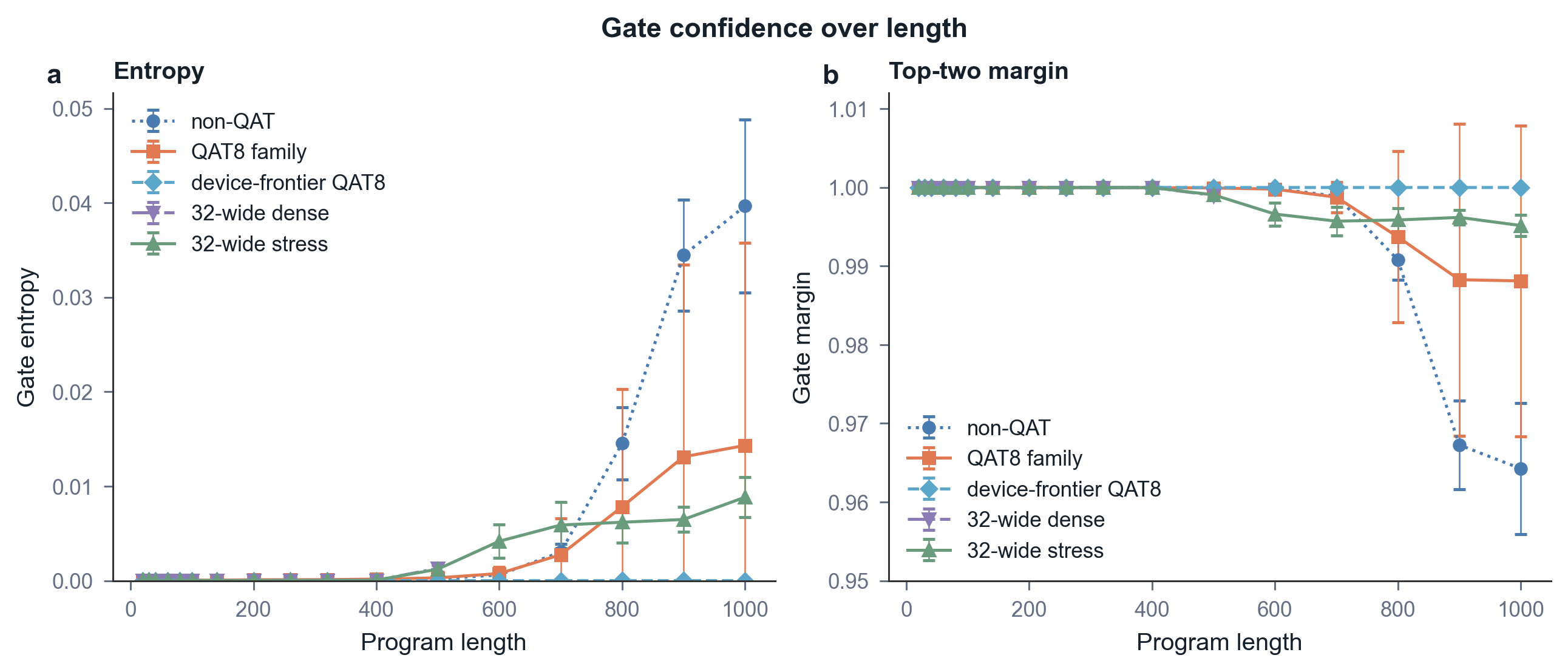}
    \caption{\textbf{Gate confidence over length.} \textbf{(a)}~Soft gate entropy and \textbf{(b)}~top-two gate margin versus program length. Both are flat (near-zero entropy, unit margin) through $L{\approx}400$, after which probability mass spreads. The bit-exact non-\qat{} run becomes the \emph{least} soft-confident at $L{=}1000$ (entropy $0.040$, margin $0.964$), the \qat{} family is intermediate (entropy $0.014$, margin $0.988$), and the long-program ValueMemory run is essentially unaffected (entropy ${\sim}2\times10^{-6}$, margin $1.000$). The argmax decision, meaning the highest-probability operation selected by the hard gate, is unchanged throughout (cf.\ Supplementary Fig.~\ref{fig:soft_hard_gap}b); only the probability \emph{spread} grows.}
    \label{fig:gate_confidence}
\end{figure*}

\begin{figure*}[t]
    \centering
    \includegraphics[width=\textwidth]{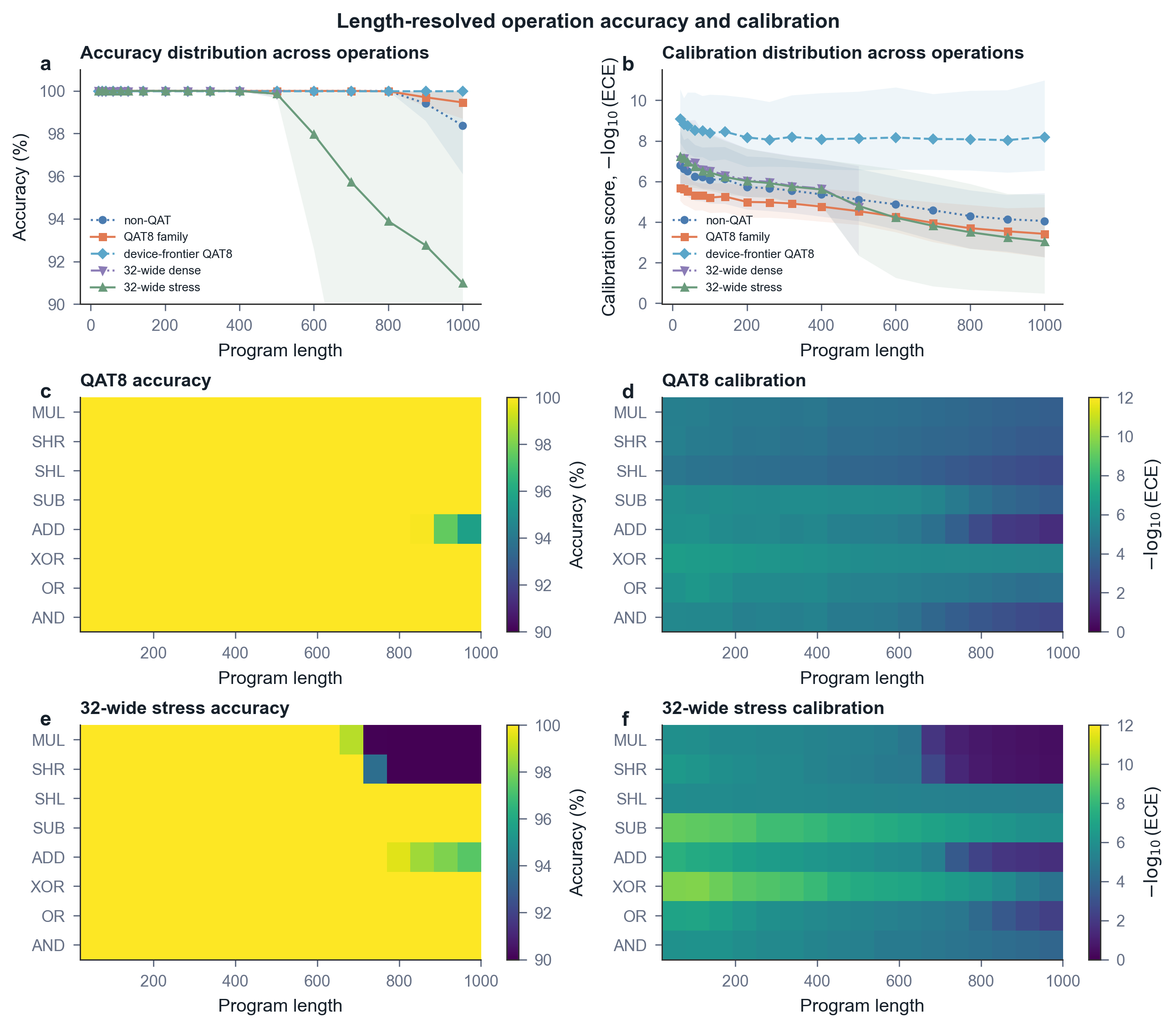}
    \caption{\textbf{Length-resolved per-operation accuracy and calibration.} \textbf{(a)}~Operation accuracy (soft-gate decoding) and \textbf{(b)}~calibration ($-\log_{10}\ECE$, where \ECE{} is expected calibration error) summarized across the eight opcodes (line: mean; band: 10--90th percentile across opcodes) versus length, per profile. \textbf{(c,d)}~Per-opcode heatmaps for the 16-wide \qat{} family: only \texttt{ADD} loses accuracy at long lengths ($95.7\%$ at $L{=}1000$) and is the weakest-calibrated row. \textbf{(e,f)}~Per-opcode heatmaps for the 32-wide extrapolation profile: the loss concentrates in \texttt{MUL} ($64.0\%$) and \texttt{SHR} ($66.5\%$) at $L{=}1000$, with the other opcodes near-exact. The long-program ValueMemory track retains the highest calibration ($-\log_{10}\ECE\approx8$) at all lengths.}
    \label{fig:per_op_calibration}
\end{figure*}

\begin{figure*}[t]
    \centering
    \includegraphics[width=\textwidth]{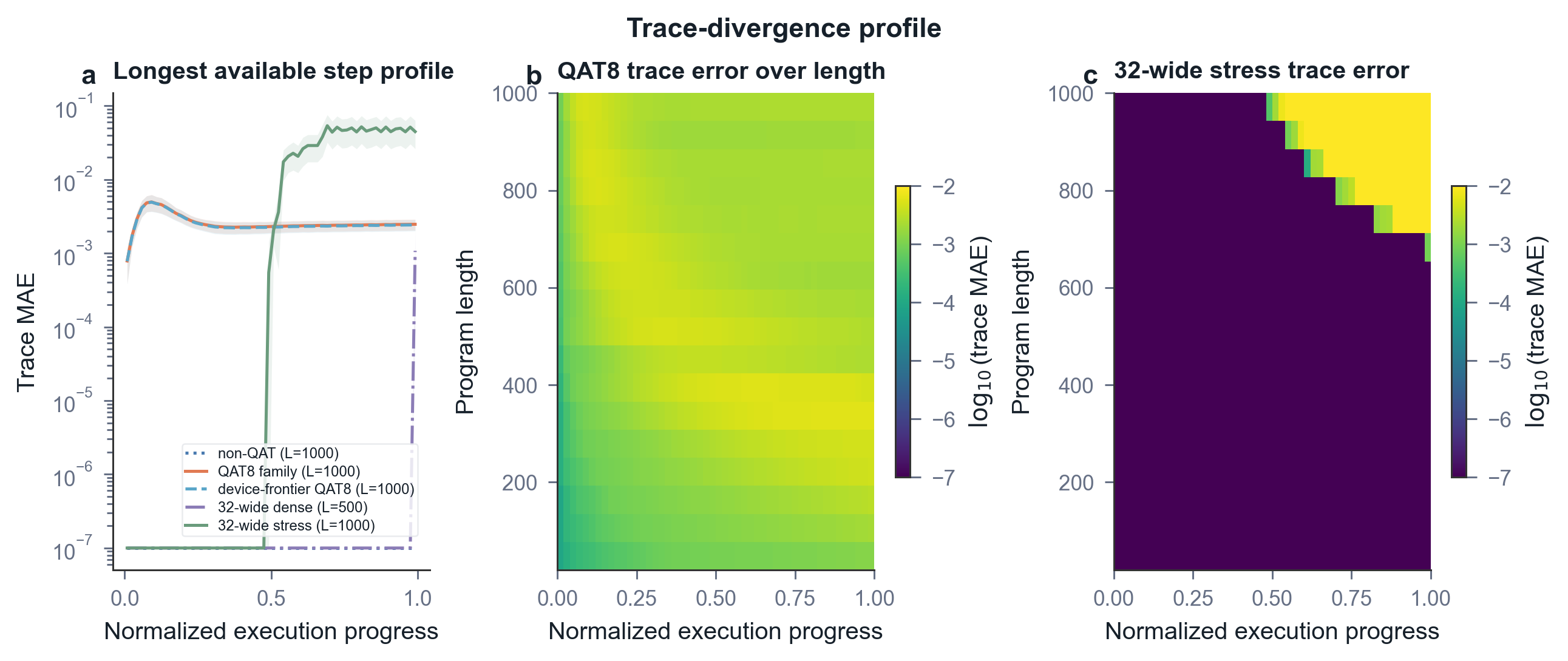}
    \caption{\textbf{Trace-divergence profile: where execution error accumulates within a program.} Per-step hard trace \MAE{} versus normalized execution progress (step$/(L{-}1)$). \textbf{(a)}~Profile at each run's longest available length (16-wide families and 32-wide extrapolation at $L{=}1000$; 32-wide dense at $L{=}500$): the 16-wide \qat{} families rise to a $2$--$3\times10^{-3}$ plateau within the first ${\sim}10\%$ of execution and stay flat, the non-\qat{} run stays at the $10^{-7}$ floor (exact at every step), and the 32-wide extrapolation run is at the floor for the first half before climbing to ${\sim}5\times10^{-2}$. \textbf{(b)}~16-wide \qat{} family heatmap ($\log_{10}$ trace \MAE{} over progress $\times$ length): the early-rising plateau is essentially length-independent. \textbf{(c)}~32-wide extrapolation heatmap: the high-error boundary is diagonal because error onset is at a fixed absolute step (${\approx}494$), i.e.\ at normalized progress $494/L$.}
    \label{fig:trace_divergence}
\end{figure*}

\begin{figure*}[t]
    \centering
    \includegraphics[width=\textwidth]{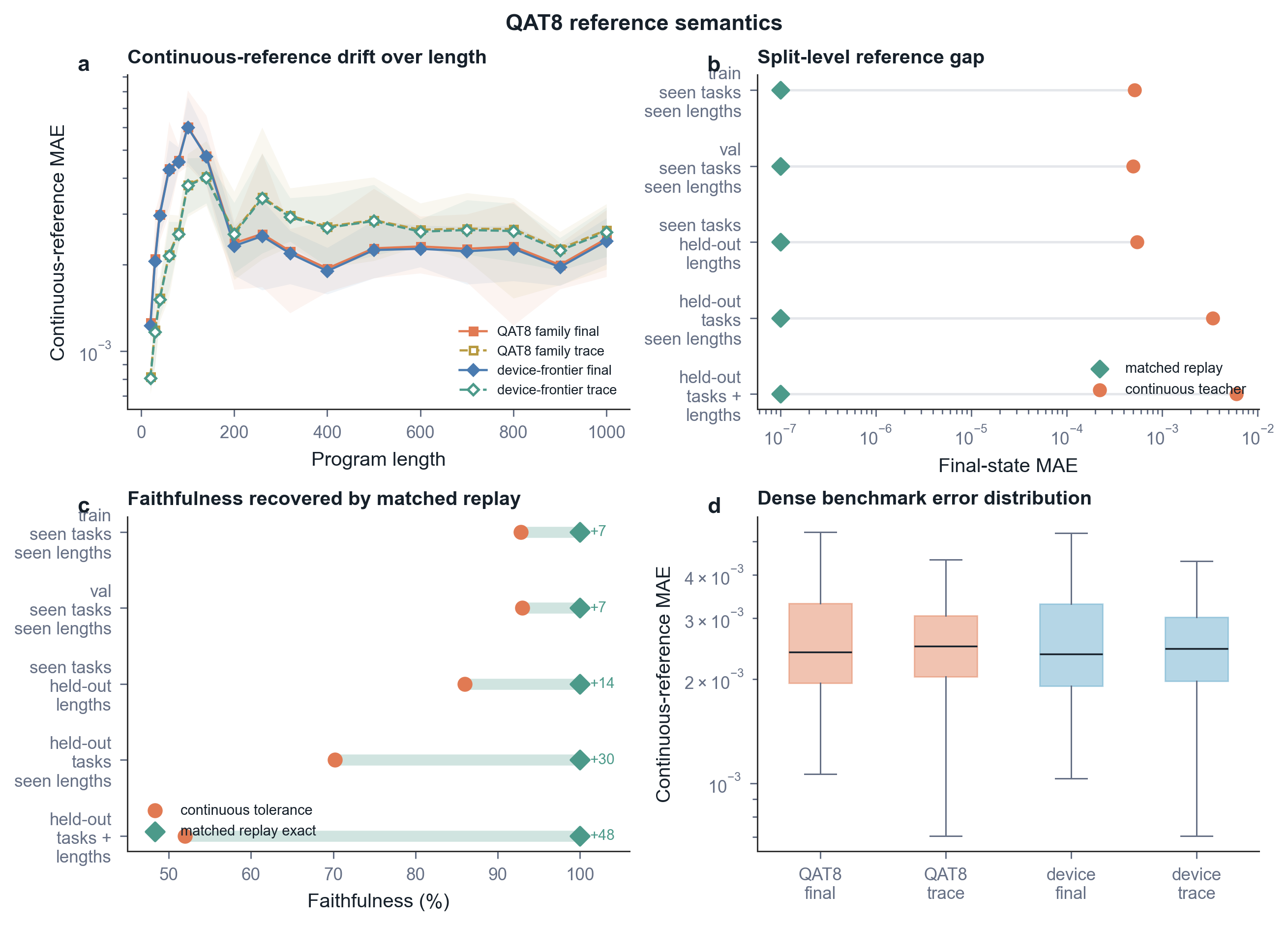}
    \caption{\textbf{Reference semantics for \qat{} writeback (full breakdown).} The matched-replay recovery summarized in \cref{fig:main_benchmark}d, reported here in full across all evaluation splits.
    \textbf{(a)}~Continuous-reference final and trace \MAE{} over the extended length benchmark for the \qat{} family and long-program ValueMemory \qat{} run; shaded bands summarize available batches/runs.
    \textbf{(b)}~Split-level error \emph{magnitude}: the final-state gap between continuous-reference scoring and matched fixed-point replay (replay errors plotted at the numerical floor when zero).
    \textbf{(c)}~Split-level program \emph{pass-rate}: faithfulness recovered by matched replay---continuous-reference tolerance faithfulness declines on held-out task/length splits, whereas matched replay exactness returns to 100\% (annotations show the recovered percentage points); this panel is the source for \cref{fig:main_benchmark}d. Panels~(b) and (c) are complementary views of the same five splits: (b) how large the continuous-reference error is, (c) how many programs that error pushes outside tolerance.
    \textbf{(d)}~Dense-benchmark distribution of continuous-reference final and trace errors.
    Matched replay gives zero final and trace \MAE{} and 100\% scalar-grid, tolerance and gate-path faithfulness on all evaluated hard splits; the continuous drift is therefore a reference-semantics mismatch rather than a breakdown of the quantization-simulated replay.}
    \label{fig:device_ref}
\end{figure*}

\begin{figure*}[t]
    \centering
    \includegraphics[width=0.95\textwidth]{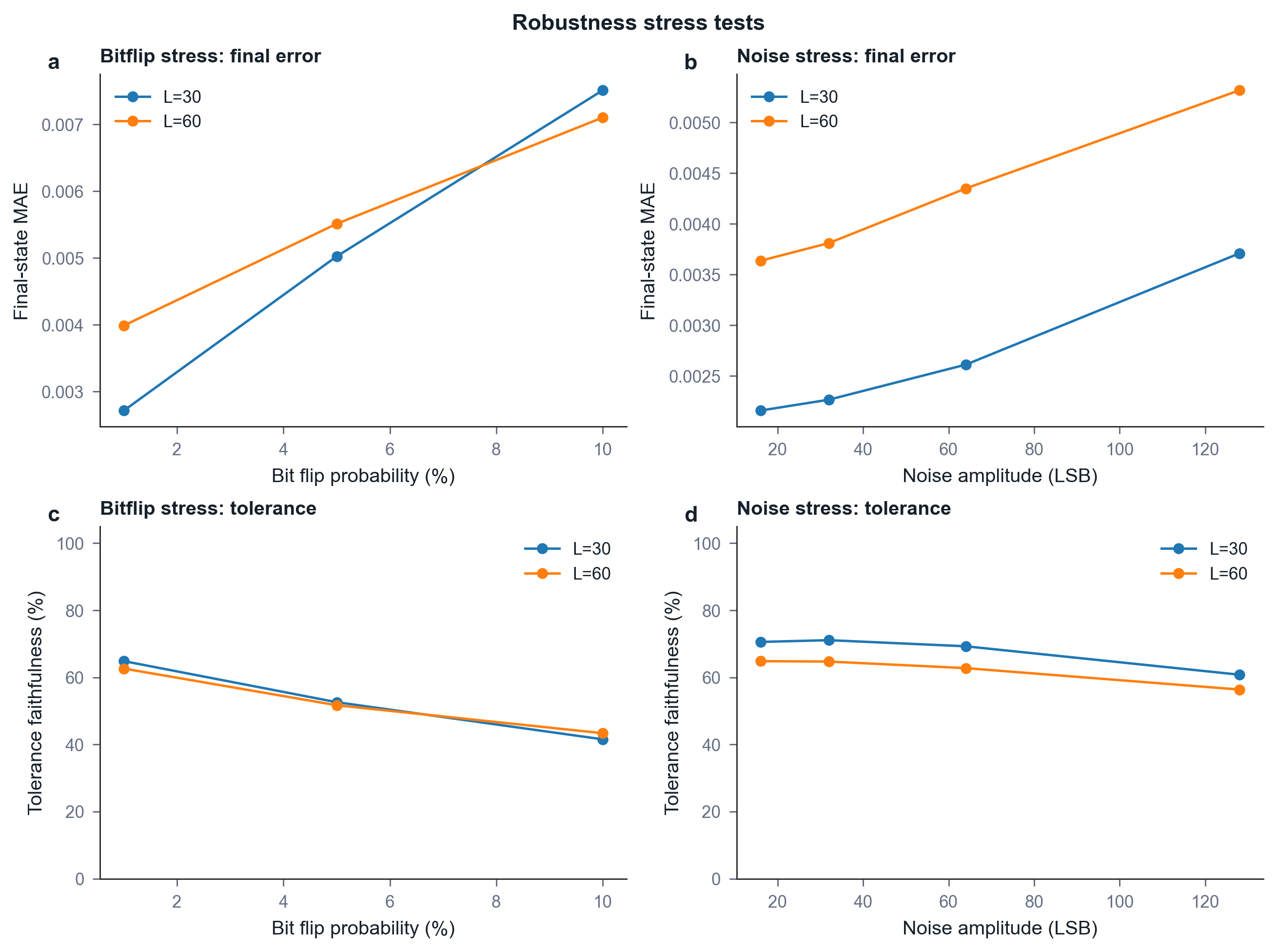}
    \caption{\textbf{Detailed robustness stress tests.} The perturbation curves underlying the robustness summary in \cref{fig:stress_cost_limits}a. \textbf{(a,b)}~Final-state \MAE{} under input bit-flip perturbation and least-significant-bit (LSB)-scaled input noise, where LSB refers to the simulated input quantization scale rather than a physical circuit bit. \textbf{(c,d)}~Tolerance faithfulness under the same perturbations. These are simulated model-input perturbations applied during evaluation, not physical fault injection, radiation modelling or circuit-level noise. The corresponding main-text panel reports the envelope: error rises and tolerance falls gradually rather than catastrophically.}
    \label{fig:supp_robustness_stress}
\end{figure*}

\begin{figure*}[t]
    \centering
    \includegraphics[width=0.95\textwidth]{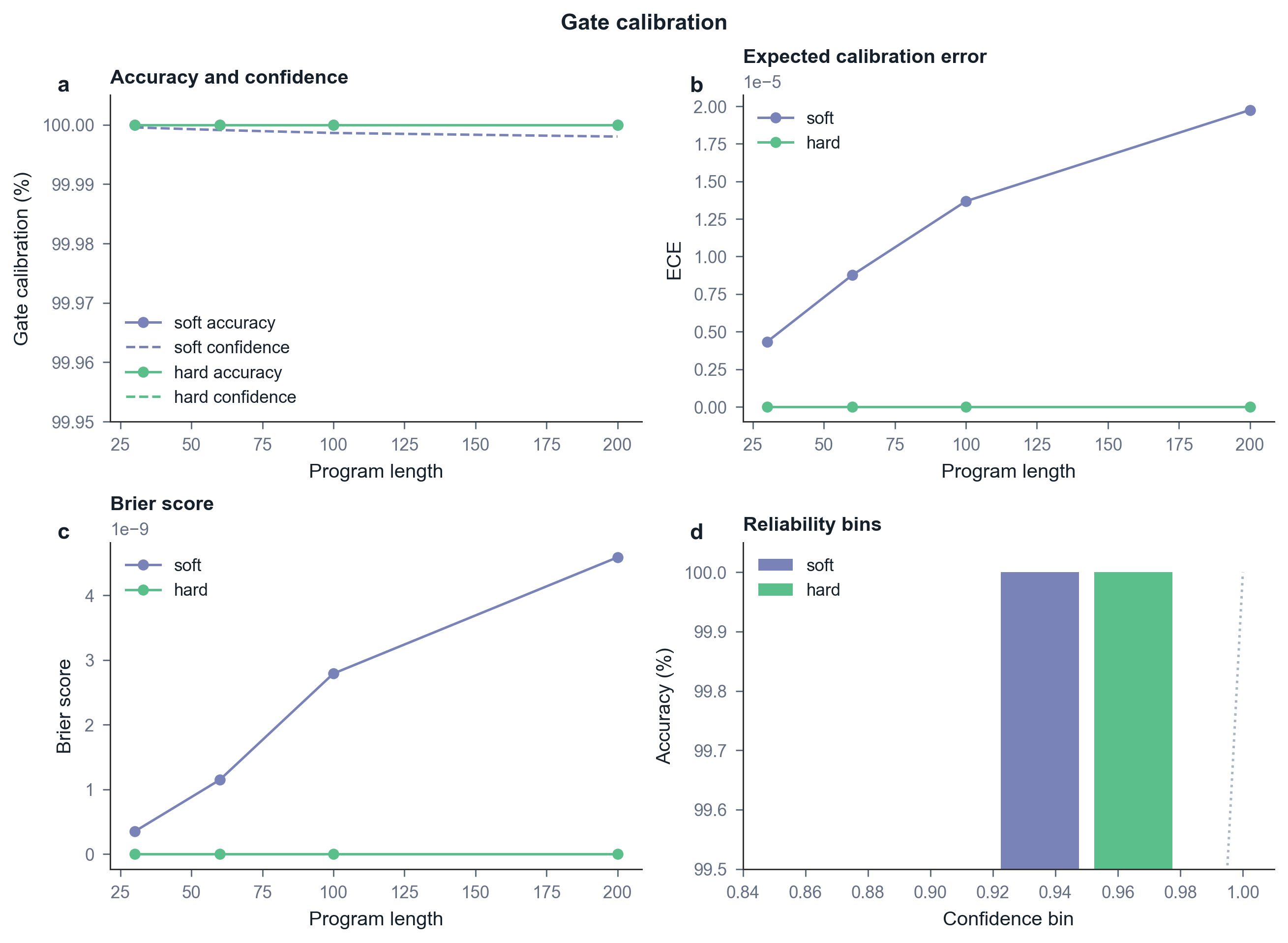}
    \caption{\textbf{Aggregate gate-calibration diagnostic.} Calibration summary for the gate router on the ordinary benchmark. The hard argmax gate path is saturated, so this aggregate figure serves as a verification diagnostic rather than as a main-text contrast. The more informative length-, width- and opcode-resolved calibration analyses are Supplementary Figs.~\ref{fig:soft_hard_gap}--\ref{fig:per_op_calibration}.}
    \label{fig:supp_gate_calibration_aggregate}
\end{figure*}

\begin{figure*}[t]
    \centering
    \includegraphics[width=0.95\textwidth]{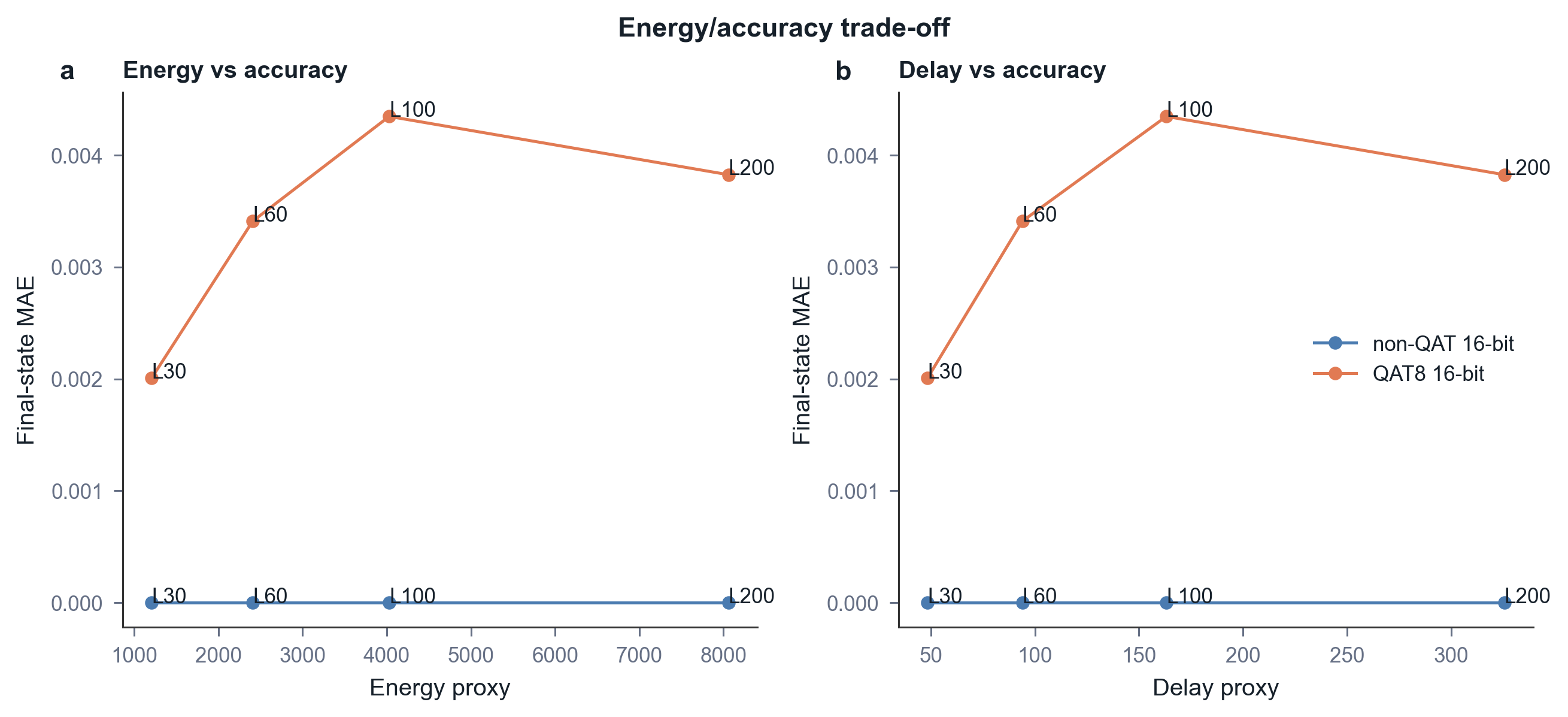}
    \caption{\textbf{Proxy cost and accuracy diagnostic.} Energy/delay proxy relation for the ordinary 16-wide benchmark. This panel is not a true Pareto frontier and is not interpreted as measured device energy or delay. It documents the shared dimensionless proxy-cost accounting; the main-text cost result is instead the equivalent-program choice experiment in \cref{fig:stress_cost_limits}b and Table~\ref{tab:energy_choice}.}
    \label{fig:supp_proxy_cost_accuracy}
\end{figure*}

\begin{figure*}[t]
    \centering
    \includegraphics[width=0.95\textwidth]{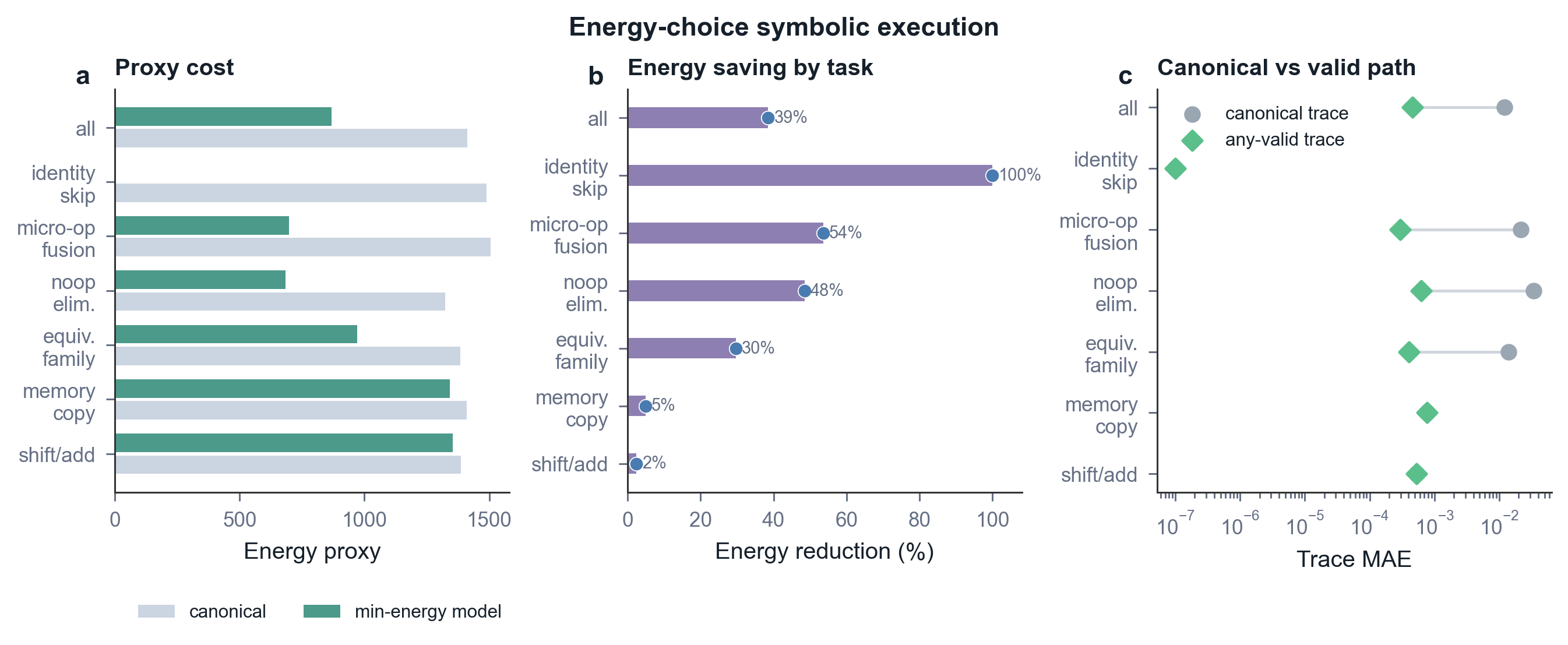}
    \caption{\textbf{Detailed energy-choice symbolic execution.} Full breakdown of the equivalent-program benchmark summarized in \cref{fig:stress_cost_limits}b. The key reading is any-valid, not canonical-only, trace agreement: for task families where the lower-cost program differs from the canonical trace, canonical-path agreement can be low while any-valid and min-energy gate agreement remain 100\%. This detailed view reports task-specific proxy-energy savings and trace errors; saturated agreement cells are stated in the main text rather than used as main evidence.}
    \label{fig:supp_energy_choice_detail}
\end{figure*}

\begin{figure*}[t]
    \centering
    \includegraphics[width=0.90\textwidth]{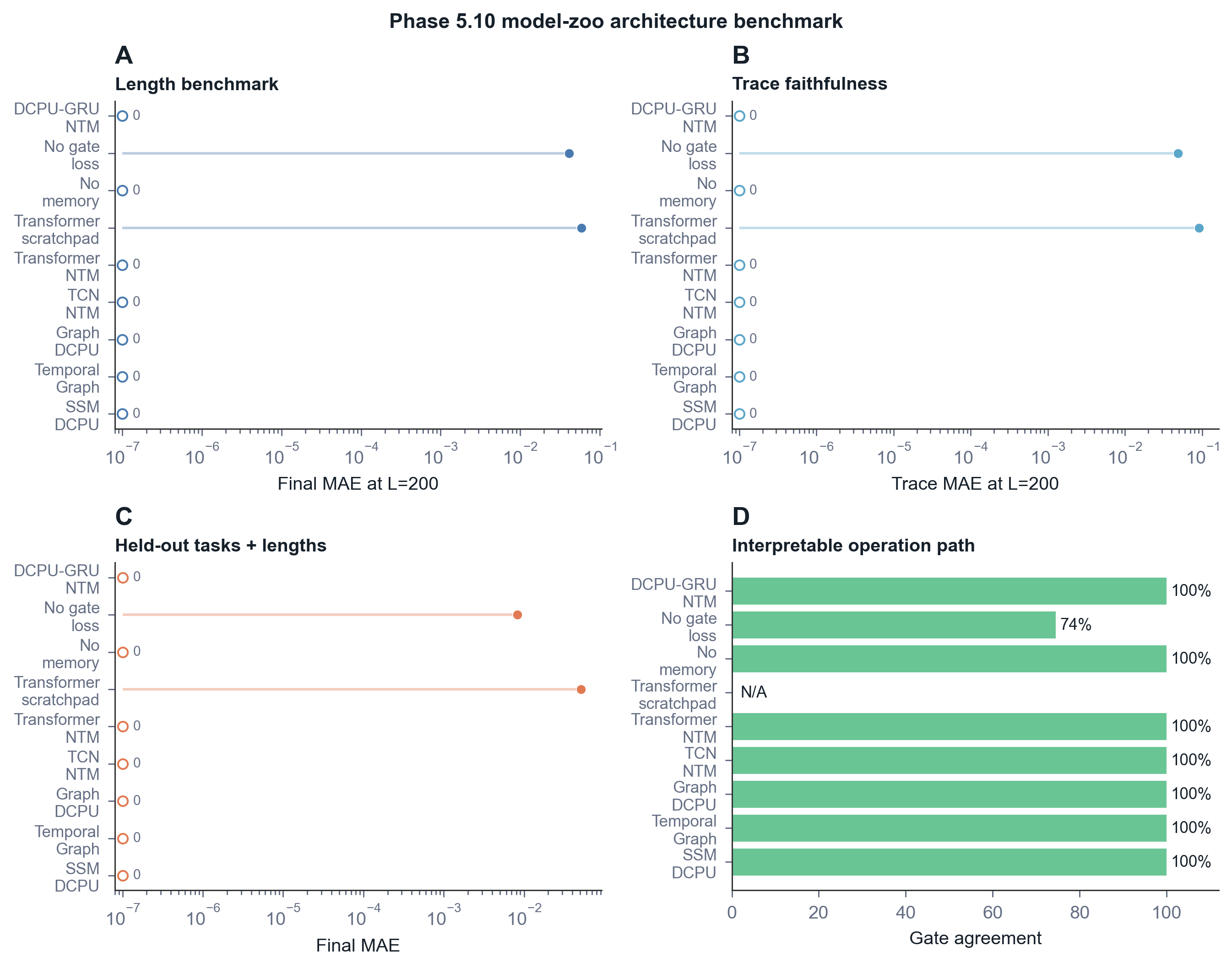}
    \caption{\textbf{Detailed architecture comparison benchmark.} Full architecture comparison on the standard shared synthetic-CPU distribution. Zero-length bars indicate errors at the display floor, not missing data. The transformer scratchpad has no operation-gate trace by design, so gate agreement is not computed for that baseline. Because most structured variants solve this baseline task distribution exactly, the main text uses this figure primarily to support two diagnostic controls: removing gate supervision degrades operation-path interpretability, and the ungated scratchpad baseline loses trace faithfulness.}
    \label{fig:supp_model_zoo_detail}
\end{figure*}

\begin{figure*}[t]
    \centering
    \includegraphics[width=0.92\textwidth]{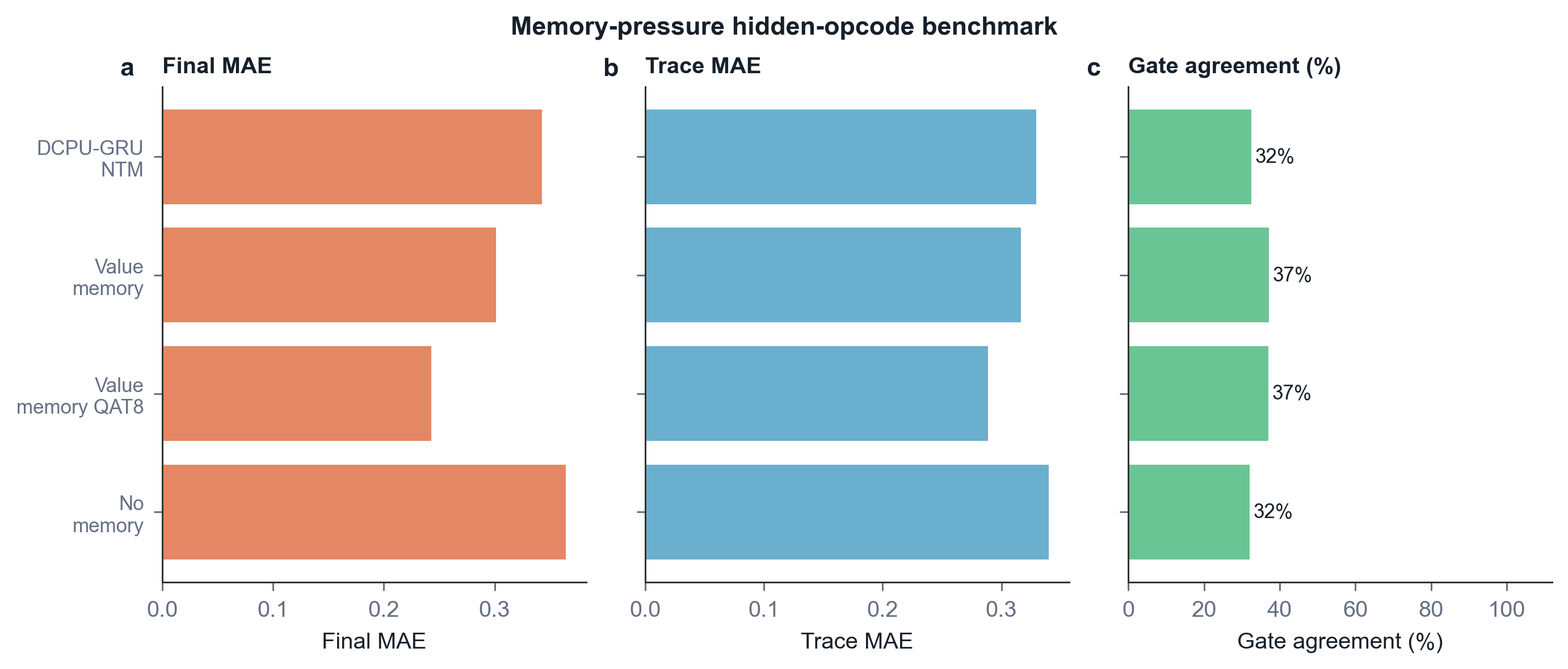}
    \caption{\textbf{Aggregate hidden-opcode memory-pressure benchmark.} Full split-level memory-pressure diagnostic behind \cref{fig:stress_cost_limits}c. The benchmark masks opcode hints and tests memory-mediated task families, so its low gate agreement is not in conflict with the near-exact visible-opcode gate-path result in \cref{fig:main_benchmark}. It measures a different mechanism: operation identity and delayed memory use must be inferred from register-state transitions rather than read directly from the instruction vector. ValueMemory improves state and trace errors in some splits, but the aggregate remains far from the visible-opcode regime because latent memory induction is still underconstrained for held-out task families.}
    \label{fig:supp_memory_pressure_aggregate}
\end{figure*}

\begin{figure*}[t]
    \centering
    \includegraphics[width=0.92\textwidth]{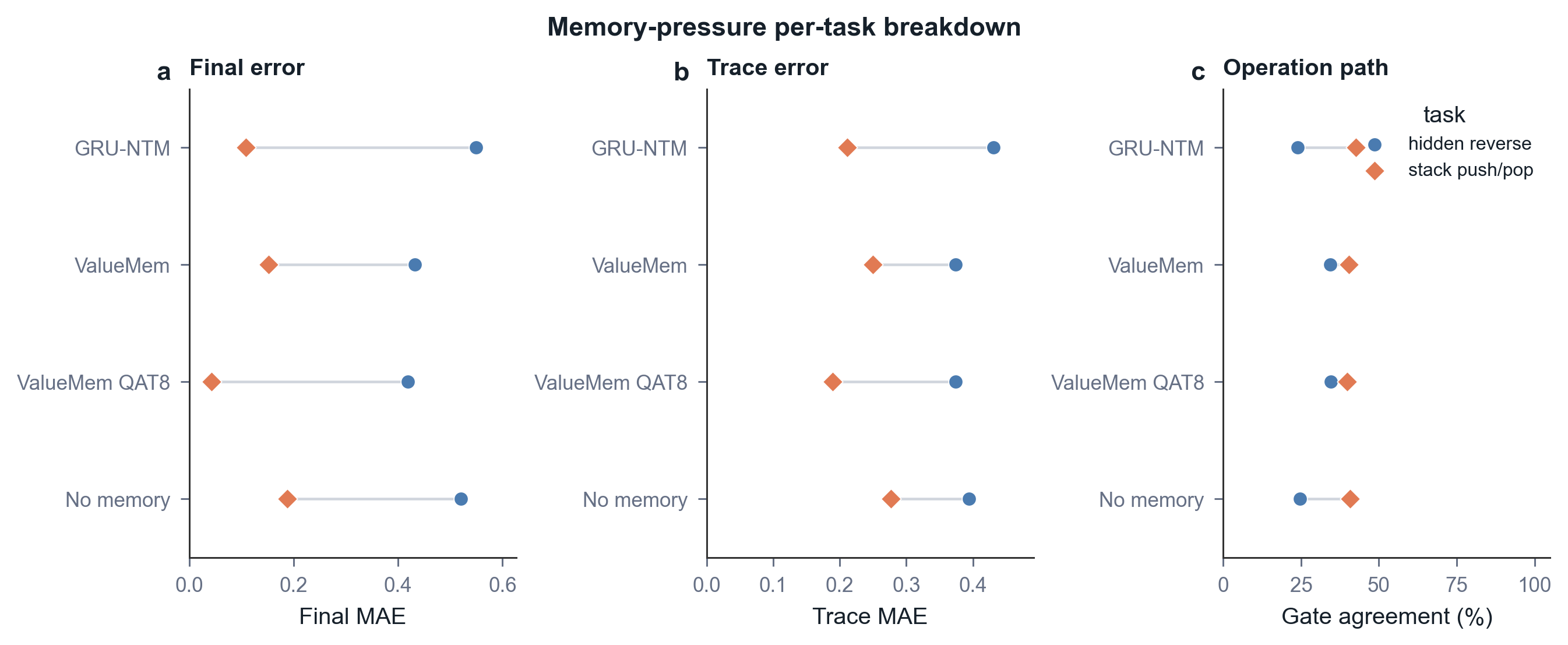}
    \caption{\textbf{Task-level hidden-memory breakdown.} Per-task decomposition of the memory-pressure benchmark summarized in \cref{fig:stress_cost_limits}d. ValueMemory and \qat{} ValueMemory improve stack-style held-out cases, where repeated local retrieval and update are useful, whereas long hidden-memory reversal remains difficult across variants because it requires sustained temporal binding, latent operation inference and ordered delayed writes. The task-level contrast explains why the aggregate degradation is concentrated in specific temporal-memory families rather than distributed uniformly across all memory tasks.}
    \label{fig:supp_memory_pressure_task_breakdown}
\end{figure*}

\begin{figure*}[t]
    \centering
    \includegraphics[width=0.92\textwidth]{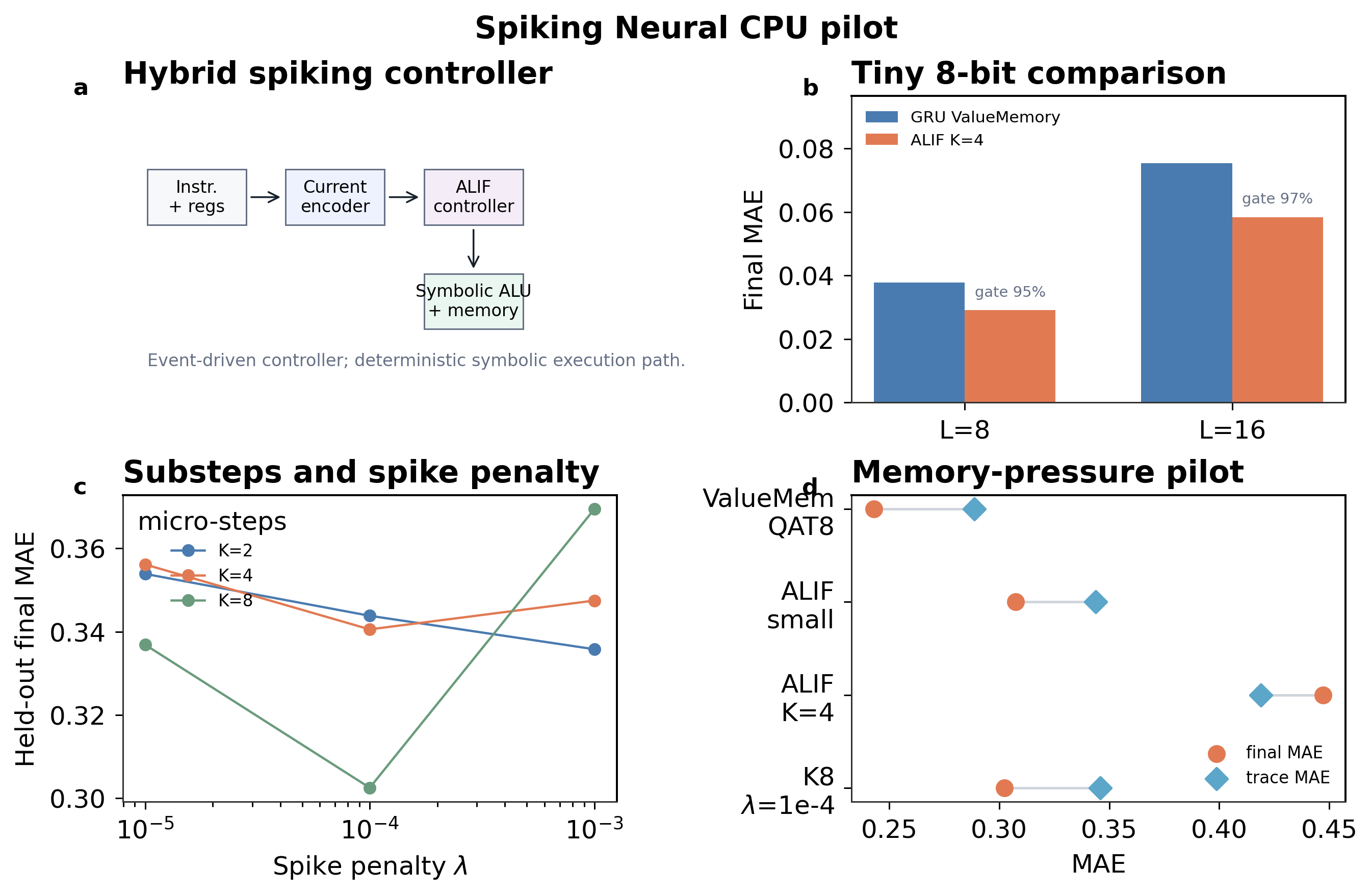}
    \caption{\textbf{Spiking-controller extension.}
    \textbf{(a)} Hybrid controller pathway used by the implemented spiking variants: instruction and register information are encoded as deterministic current, the adaptive leaky integrate-and-fire (\alif{}) controller supplies temporal hidden state, and the symbolic \alu{} and memory/writeback path remain non-spiking and auditable.
    \textbf{(b)} Compact 8-wide comparison between the non-spiking GRU ValueMemory baseline and an \alif{} controller with $K{=}4$ micro-steps; labels report gate agreement for the \alif{} run.
    \textbf{(c)} Held-out memory-pressure final \MAE{} as a function of spike-regularization weight $\lambda$ and micro-step count $K$.
    \textbf{(d)} Memory-pressure comparison of final and trace \MAE{} for ValueMemory-\qat{} and selected \alif{} variants. These panels characterize a neuromorphic-controller extension; they are not a claim of measured event energy or superiority over the non-spiking \qat{} baseline. Spike rates are averaged returned spike traces, not raw hardware event counts.}
    \label{fig:snn_pilot}
\end{figure*}

\renewcommand{\figurename}{Figure}
\clearpage
\subsection*{Supplementary tables}
\renewcommand{\thetable}{S\arabic{table}}
\renewcommand{\tablename}{Supplementary Table}
\setcounter{table}{0}

\begin{table}[t]
\centering
\caption{\textbf{What sets the 32-wide long-horizon limit: a controlled training ablation.} The identical 32-wide non-\qat{} GRU--NTM symbolic executor (width $32$, $16$ registers, hidden state $256$; fixed seed) was retrained while varying one factor at a time---maximum training length, memory slots, or epochs---and benchmarked over $L{=}20$--$1000$ ($8$ batches $\times\,64$). ``Exact through'' is the largest length with mean hard final \MAE{} exactly $0$. Training horizon---not state width, memory capacity, or training duration---governs the limit: maximum training length ${\ge}80$ removes it entirely, while more memory slots or more epochs at maximum training length $40$ do not.}
\label{tab:t400_mechanism}
\begin{tabular}{lrrrrr}
\toprule
Variant & Max train $L$ & epochs & mem.\ slots & Exact through & Final \MAE{} @ $L{=}1000$\\
\midrule
Baseline configuration & 40  & 15 & 256 & 520  & $1.27\times10^{-2}$\\
Longer horizon            & 80  & 15 & 256 & 1000 & $0$\\
Longer horizon            & 160 & 15 & 256 & 1000 & $0$\\
Fewer memory slots        & 40  & 15 & 128 & 1000 & $0$\\
More memory slots         & 40  & 15 & 512 & 700  & $9.24\times10^{-4}$\\
More epochs               & 40  & 45 & 256 & 520  & $7.02\times10^{-2}$\\
\bottomrule
\end{tabular}
\end{table}

\renewcommand{\thetable}{\arabic{table}}
\renewcommand{\tablename}{Table}
\clearpage
\subsection*{Supplementary note 1: full operation semantics}

For normalized vectors $x,y\in[0,1]^W$, the implemented reference operations are:
\begin{align}
    \mathrm{AND}(x,y) &= \clip(x\odot y,0,1),\\
    \mathrm{OR}(x,y) &= \max(x,y),\\
    \mathrm{XOR}(x,y) &= |x-y|,\\
    \mathrm{ADD}(x,y) &= \clip(x+y,0,1),\\
    \mathrm{SUB}(x,y) &= \clip(x-y,0,1),\\
    \mathrm{SHL}(x,y) &= \clip(2x,0,1),\\
    \mathrm{SHR}(x,y) &= 0.5x,\\
    \mathrm{MUL}(x,y) &= \clip(x\odot y,0,1).
\end{align}
The neural model uses the same fixed operation bank and learns to select the correct operation gate. The matched fixed-point replay uses the same operations and quantizes destination writeback values to the same simulated low-precision grid.
In this operation bank, \texttt{AND} and \texttt{MUL} are identical continuous maps. They remain different labels for gate-supervision analysis, but they do not yet implement distinct integer bitwise and arithmetic semantics.

\subsection*{Supplementary note 2: manuscript scope}

This scope map separates the evidence reported in the article from the follow-up directions it motivates. The central evidence is symbolic execution and quantization-simulated writeback under matched fixed-point replay. Architecture comparisons provide ablations, spiking controllers provide a device-facing extension, candidate-constrained action search provides the closed-loop control result, and the RV32I base-integer bridge provides a standardized-ISA extension. Official \riscv{} compliance, broader \riscv{} extensions, applied anomaly detection on constrained devices and hardware implementation are reserved for future studies.

\begin{longtable}{p{0.18\linewidth}p{0.33\linewidth}p{0.39\linewidth}}
\caption{\textbf{Manuscript scope.} Evidence blocks are assigned to the main execution claim, supplementary analyses or future research directions.}
\label{tab:research_program_map}\\
\toprule
Experiment block & Role in this manuscript & Interpretation\\
\midrule
\endfirsthead
\toprule
Experiment block & Role in this manuscript & Interpretation\\
\midrule
\endhead
Implementation and generated data & Symbolic neural CPU implementation & Reproducible basis for the reported experiments\\
Trace learning & Gate-supervised execution & Primary evidence for interpretable execution\\
Length generalization & Long-program evaluation & Primary evidence for extrapolation within controlled tasks\\
Low-precision replay & \qat{} and simulated robustness & Primary quantization-simulated writeback evidence\\
Cost-aware execution & Dimensionless energy/delay proxies and energy choice & Primary evidence for cost-aware symbolic execution\\
Architecture benchmark & Architecture comparison & Supplementary architecture comparison and ablations\\
Spiking controllers & Event-driven controller variants & Device-facing extension, not main claim\\
Closed-loop controller & Actor-critic symbolic control & Supplementary controller extension\\
RV32I base-integer bridge & Standardized-ISA extension & Deterministic RV32I semantic audit and learned-control diagnostic; not official \riscv{} compliance or arbitrary \riscv{} binary execution\\
Broader \riscv{} extensions and runtime & Follow-up processor/toolchain project & Privileged modes, CSRs, interrupts, ABI/runtime behaviour and M/A/C/F/D/V extensions are outside the present article\\
Candidate action search & Candidate masking and beam evaluation & Control extension built on the trace-audited execution substrate\\
\bottomrule
\end{longtable}

\subsection*{Supplementary note 3: derivation of the gate-supervised objective}

The gate-supervised objective can be viewed as a constrained maximum-likelihood problem. Let $r_{1:T}$ be the reference trace and $o_{1:T}$ the operation sequence. A fully probabilistic model factorizes as
\begin{equation}
    p_{\theta}(r_{1:T},o_{1:T}\mid x_{1:T},r_0)
    =
    \prod_{t=1}^{T}
    p_{\theta}(o_t\mid h_t)
    p_{\theta}(r_t\mid r_{t-1},o_t,h_t).
\end{equation}
Assuming a Laplace observation model for registers,
\begin{equation}
    p_{\theta}(r_t\mid r_{t-1},o_t,h_t)
    \propto
    \exp\left(
    -\frac{1}{\sigma}
    \|\hat{r}_t-r_t\|_1
    \right),
\end{equation}
and a categorical model for operations,
\begin{equation}
    p_{\theta}(o_t=k\mid h_t)=\pi_{t,k},
\end{equation}
the negative log-likelihood is proportional to
\begin{equation}
    \sum_t \|\hat{r}_t-r_t\|_1
    -
    \sigma \sum_t \log \pi_{t,o_t}.
\end{equation}
The implemented loss is a weighted and regularized version of this objective, with an additional final-state term to emphasize endpoint correctness and entropy/smoothness terms to improve optimization. Thus gate supervision is not an arbitrary auxiliary loss. It is the categorical part of the likelihood for the symbolic execution path.

\subsection*{Supplementary note 4: why energy-choice tasks are necessary}

Passive energy reporting is insufficient for a physics-aware execution claim. If a benchmark contains only one valid program for each target state, then energy is a label attached to the path, not a decision variable. Equivalent-program tasks create at least two paths:
\begin{equation}
    x^{(1)}\neq x^{(2)},
    \qquad
    \mathcal{T}(r_0,x^{(1)})=\mathcal{T}(r_0,x^{(2)}),
    \qquad
    E(x^{(1)})\neq E(x^{(2)}).
\end{equation}
Only then can the model be tested for cost-aware symbolic execution. The min-energy policy solves
\begin{equation}
    x^\star =
    \argmin_{x\in\mathcal{X}(r_0,r_T)}
    E(x)
    \quad
    \mathrm{subject\ to}\quad
    \mathcal{T}(r_0,x)=r_T.
\end{equation}
The implementation uses constructed equivalent families rather than a general compiler optimizer. This demonstrates the principle under controlled conditions while leaving general program optimization for future work.

\subsection*{Supplementary note 5: real-device pathway}

The simulation-to-device pathway is staged. First, export the trained \qat{} executor as fixed-point kernels with explicit scales and clipping ranges matching the $Q_B$ projection. Second, validate agreement between the matched fixed-point replay and the exported implementation. Third, deploy the kernel to an FPGA, microcontroller-class accelerator or neuromorphic co-processor interface. Fourth, replace proxy energy with measured switching, latency and power. Fifth, close the loop with hardware-in-the-loop tests.

The most plausible first physical demonstrator is not a general-purpose CPU. It is a symbolic neural co-processor with a compact instruction vocabulary, fixed-point register state and auditable operation gates. Such a demonstrator can be useful for edge controllers, sensor-state machines and anomaly-detection front-ends before it attempts richer instruction-set compatibility.

\subsection*{Supplementary note 6: interpretation of saturated and contrasting metrics}

\paragraph{Controlled execution distribution.}
The main distribution is controlled so that exact trace supervision, named operation labels and matched fixed-point replay can be computed without ambiguity. This design is necessary for a strict audit of execution: each active timestep has a reference register state, a reference opcode, a destination register and a low-precision replay reference. The harder task-family and memory-pressure benchmarks are therefore not replacements for the main distribution; they are contrastive tests that remove or perturb specific cues after the visible-opcode executor has been established.

\paragraph{Visible-opcode and hidden-opcode regimes.}
The main length benchmark evaluates visible-opcode symbolic execution, whereas \cref{fig:stress_cost_limits} combines input perturbations, equivalent-program choices, architecture ablations and hidden-opcode memory tasks. Exact operation paths on visible-opcode programs are compatible with low gate agreement on hidden-opcode memory-pressure tasks because the latter removes the opcode one-hot cue and requires the model to infer the operation family from register-state transitions, delayed memory use and address-dependent writes. The mechanism is therefore not a contradiction between figures: it is a change in available information. In visible-opcode execution the router can condition directly on the encoded operation hint; in hidden-opcode memory pressure the same router must infer the operation identity and memory-access pattern from context, which is underconstrained for long hidden reversals and stack-like updates.

\paragraph{Cross-benchmark signature.}
The cross-benchmark signature is a compact visual index of benchmark axes, not a universal scalar ranking. State and trace entries are measured \MAE{} values; gate-path loss is $1$ minus gate agreement; tolerance loss is $1$ minus tolerance faithfulness; and cost saving is a positive energy-proxy reduction. Blank cells indicate metrics that are not defined by the corresponding benchmark protocol. The scratchpad row has no gate-path value because the baseline does not expose the named operation-gate trace used by the structured executor. This distinction matters: scratchpad models can imitate endpoints, but they do not provide the operation-level audit trail that defines the symbolic Neural CPU evaluation.

\paragraph{Trace supervision and output imitation.}
The executor is not evaluated only as an input--output approximator. Training includes final-state, trace and categorical operation-gate objectives, and evaluation reports full register trajectories and operation-path agreement. Removing gate supervision leaves the state and trace losses active, so the model can still learn some value updates, but the operation logits become weakly constrained by endpoint imitation alone. This explains the NoGateLoss pattern: the benchmark does not collapse into random output, yet operation-path agreement drops because the categorical likelihood term for the symbolic path has been removed.

\paragraph{Low-precision reference semantics.}
Continuous-reference drift appears under \qat{} because the executor writes back on an 8-bit scalar grid while the reference trajectory is continuous. Matched fixed-point replay changes the reference to the same writeback semantics and recovers exact final and trace agreement on the relevant splits. This is the appropriate simulation comparison before hardware export: it verifies that the learned executor follows its own low-precision symbolic semantics, while not claiming integer-only deployment, measured power or cycle-accurate hardware timing.

\paragraph{Proxy energy and spiking controllers.}
Energy and delay values are dimensionless simulation proxies derived from operation tables and program length, not physical measurements. They are used to test whether equivalent-program tasks can induce lower-proxy-cost valid execution, not to claim measured silicon savings. Likewise, the spiking controllers are a device-facing architectural extension: they replace the recurrent controller state with LIF/ALIF spike dynamics while keeping the symbolic ALU and memory path non-spiking. The SNN results support an inspectable neuromorphic research pathway, but they do not supersede the non-spiking \qat{} executor on memory-pressure benchmarks.

\paragraph{Closed-loop control.}
Candidate-constrained control improves success by narrowing the combinatorial symbolic action space before reinforcement learning chooses executable instructions. The residual oracle gap remains because the learned controller must still rank candidate operations, source registers and destinations under finite-horizon rewards and, in the learned-transition setting, under approximation error from the \qat{} transition model. The control results are therefore an extension of the learned execution substrate rather than the core evidence for trace-faithful symbolic execution.

\subsection*{Supplementary note 7: future device and industrial application pathway}

Edge and embedded anomaly detection on constrained devices is a motivating application direction. No industrial field dataset is included in this manuscript, and no anomaly-detection result is claimed. The present article is focused on implemented trace-supervised symbolic execution, quantization-simulated writeback under matched fixed-point replay, architecture comparison ablations, energy-choice execution, memory-pressure limits, spiking-controller extensions and candidate-constrained control. A later applied study can test whether this substrate improves anomaly detection on real multivariate sensor streams under device latency, energy and interpretability constraints.

The future device route is similarly staged. The present work provides fixed-width register states, auditable operation gates, quantization-simulated \alu{} or writeback execution and proxy costs. A hardware study would need exported fixed-point kernels, agreement against the matched replay, hardware-in-the-loop validation and measured latency/energy. These steps are outside the present results and are framed as the next research layer rather than as claims of this article.

\subsection*{Supplementary note 8: notation reference}

This notation table is included to keep the main text readable while preserving strict mathematical consistency for implementation-oriented readers.

\begin{longtable}{p{0.18\linewidth}p{0.72\linewidth}}
\caption{\textbf{Notation used in the manuscript.}}\label{tab:notation}\\
\toprule
Symbol & Meaning\\
\midrule
\endfirsthead
\toprule
Symbol & Meaning\\
\midrule
\endhead
$W$ & Register-vector width used by the profile; implemented as a continuous vector dimension, not a packed integer word.\\
$R$ & Number of registers.\\
$D_R$ & Guarded instruction-index denominator, $D_R=\max(1,R-1)$, matching the implementation.\\
$r_t,\hat{r}_t$ & Reference and predicted register files at program step $t$.\\
$x_t$ & Encoded instruction containing one-hot opcode and normalized source/destination register indices.\\
$\xi_t$ & Symbolic instruction tuple $(o_t,i_t,j_t,d_t)$ before numeric encoding.\\
$o_t$ & Symbolic operation label at step $t$.\\
$i_t,j_t,d_t$ & Zero-based source-register and destination-register indices.\\
$u_t,v_t$ & Operand vectors read from registers $i_t$ and $j_t$.\\
$A_k$ & Implemented continuous operation proxy for operation label $k$.\\
$\pi_t$ & Learned gate distribution over symbolic operations.\\
$m_t$ & Binary valid-step or padding mask.\\
$M_t$ & Differentiable NTM-style memory state.\\
$V_t$ & ValueMemory state used by the value-memory variant.\\
$Q_B$ & Uniform quantization-simulated \alu{} or writeback projection with $B$ bits.\\
$N_B$ & Mini-batch size; distinct from quantization bit depth $B$.\\
$B_{\mathrm{beam}}$ & Beam width in candidate-control evaluation.\\
$E,\Delta$ & Dimensionless simulation proxy energy and delay, not physical measurements.\\
$z_{t,k}$ & SNN spike state at micro-step $k$ inside program step $t$.\\
$\mathcal{C}(s)$ & Candidate action set for controller state $s$.\\
\bottomrule
\end{longtable}

\bibliographystyle{sn-nature}
\bibliography{references}

\end{document}